\pdfoutput=1

\documentclass[11pt]{article}

\usepackage[preprint]{acl}

\usepackage{times}
\usepackage{latexsym}

\usepackage[T1]{fontenc}

\usepackage[utf8]{inputenc}

\usepackage{microtype}

\usepackage{inconsolata}

\usepackage{graphicx}
\usepackage{multirow,multicol}
\usepackage{cleveref}
\usepackage{paralist}
\usepackage{xspace}
\usepackage[normalem]{ulem}
\usepackage{stfloats}
\usepackage{enumitem}
\usepackage{makecell}
\usepackage{paralist}
\usepackage{adjustbox}
\usepackage{xcolor}
\usepackage[edges]{forest}

\crefname{equation}{Eq.}{Eq.}
\crefname{section}{\S}{\S}

\newcommand{\ie}{\emph{i.e.,}\xspace}

\newcommand{\eg}{\emph{e.g.,}\xspace}

\newcommand{\refSubfigure}[2]{Figure~\hyperref[#1]{\ref{#1}(#2)}}

\definecolor{darkgreen}{rgb}{0.0, 0.5, 0.0}
\definecolor{darkpurple}{rgb}{0.4, 0.0, 0.4}
\definecolor{lightpurple}{rgb}{0.7, 0.3, 0.7}
\colorlet{training-fill}{blue!15}
\colorlet{training-border}{blue!50}
\colorlet{inference-fill}{darkgreen!15}
\colorlet{inference-border}{darkgreen!50}
\colorlet{post-inference-fill}{red!15}
\colorlet{post-inference-border}{red!50}
\colorlet{benchmark-fill}{lightpurple!15}
\colorlet{benchmark-border}{lightpurple!50}

\title{
    \adjustbox{valign=c}{\includegraphics[width=2em]{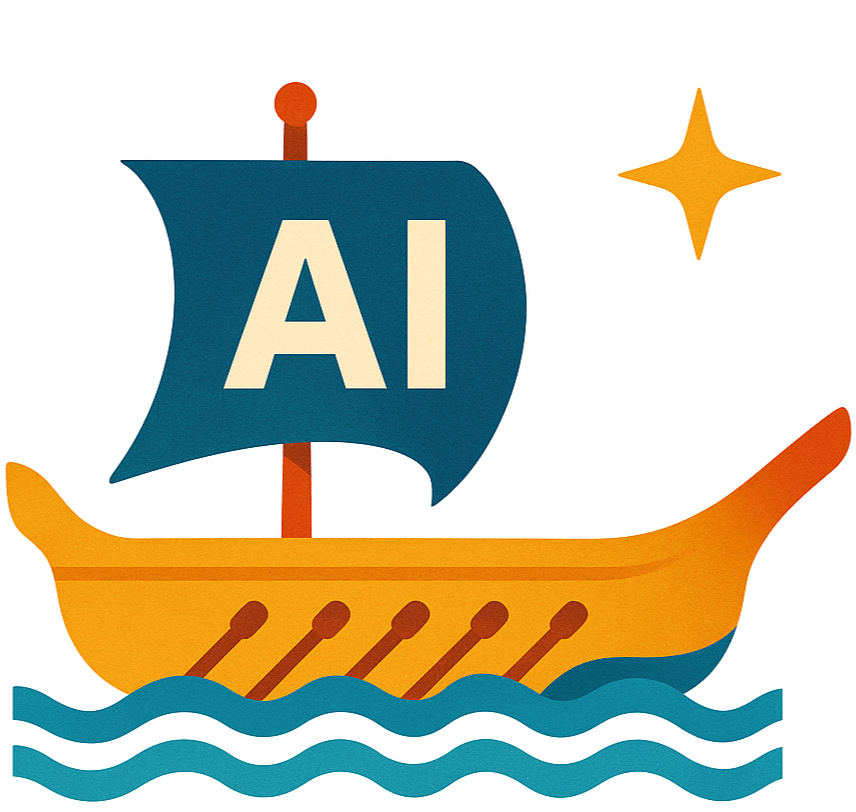}}
    \textit{Sailing by the Stars}: A Survey on Reward Models and Learning Strategies for Learning from Rewards
}

\author{
    Xiaobao Wu
    \\
    Nanyang Technological University \\
   \texttt{xiaobao002@e.ntu.edu.sg}
}

\begin{document}
\maketitle
\begin{abstract}
    Recent developments in Large Language Models (LLMs) have shifted from pre-training scaling to post-training and test-time scaling.
    Across these developments, a key unified paradigm has arisen: \textit{Learning from Rewards}, where reward signals act as the guiding stars to steer LLM behavior.
    It has underpinned a wide range of prevalent techniques, such as reinforcement learning (RLHF, RLAIF, DPO, and GRPO), reward-guided decoding, and post-hoc correction.
    Crucially, this paradigm enables the transition from passive learning from static data to active learning from dynamic feedback.
    This endows LLMs with aligned preferences and deep reasoning capabilities for diverse tasks.
    In this survey, we present a comprehensive overview of learning from rewards,
    from the perspective of reward models and learning strategies across training, inference, and post-inference stages.
    We further discuss the benchmarks for reward models and the primary applications.
    Finally we highlight the challenges and future directions.%
    ~\footnote{We maintain a paper collection at \url{https://github.com/bobxwu/learning-from-rewards-llm-papers}.}
\end{abstract}

\section{Introduction}
    Recent years have witnessed the rapid advancement of Large Language Models (LLMs), such as ChatGPT \cite{openai2023gpt4}, Claude \cite{claude3.7sonnet}, and Llama \cite{llama2-2023,llama3herdmodels2024}.
    These models are initially empowered by \textit{pre-training scaling} \cite{kaplan2020scaling}, which trains LLMs on massive corpora through next-token prediction.
    While this approach enables broad linguistic and knowledge representations, it suffers from several fundamental limitations:
    misalignment with human values \cite{bai2022constitutional,zhang2023hallucination-snowball,deshpande2023toxicity-chatgpt},
    difficulty in adapting to various task objectives \cite{lyu2023faithful-cot,wang2023decodingtrust},
    and deficiencies in deep reasoning \cite{mirzadeh2024gsm-symbolic,wu2024reasoning-or-reciting}.
    As a result, these limitations confine pre-trained models to surface-level tasks, falling short of the long-term goal of robust and general AI.
    To address these limitations, recent efforts have turned to \textit{post-training} and \textit{test-time scaling},
    which seek to further refine LLMs after pre-training.

\begin{figure}[!t]
    \centering
    \includegraphics[width=\linewidth]{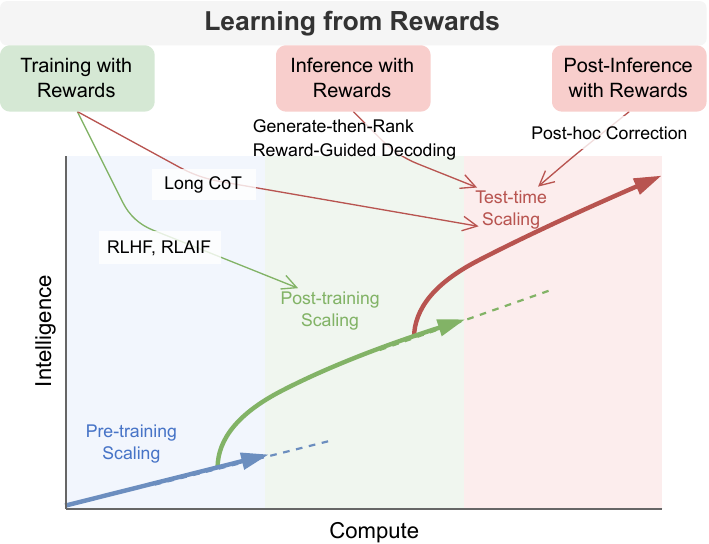}
    \caption{
        Illustration of the scaling phases of LLMs.
        The learning-from-rewards paradigm plays a pivotal role in the post-training and test-time scaling.
    }
    \label{fig_scaling}
\end{figure}

    Across the post-training and test-time scaling, a critical unified paradigm has emerged as illustrated in \Cref{fig_scaling}: \textbf{\textit{Learning from Rewards}}, which leverages reward signals to guide model behavior through diverse learning strategies.
    For post-training scaling,
    this paradigm has underpinned several key techniques, including preference alignment through Reinforcement Learning from Human Feedback \cite[RLHF, ][]{ouyang2022rlhf} or AI Feedback \cite[RLAIF, ][]{bai2022constitutional} with scalar rewards and PPO \cite{schulman2017proximal}, and Direct Preference Optimization \cite[DPO, ][]{rafailov2023dpo} with implicit rewards.
    For test-time scaling, this paradigm supports eliciting long Chain-of-Thoughts reasoning via GRPO \cite{shao2024deepseekmath} with rule-based rewards in DeepSeek-R1 \cite{deepseekai2025deepseek-r1},
    generate-then-rank (\textit{Best-of-N}) \cite{cobbe2021training-verifiers,lightman2023letsverify},
    reward-guided decoding \cite{deng2023rad,khanov2024args},
    and post-hoc correction \cite{akyurek2023rl4f,madaan2023self-refine}.
    By these techniques, this paradigm enables LLMs to learn actively from dynamic feedback, in contrast to learning passively from static data.
    As such, this endows LLMs with aligned preferences and deep reasoning and planning abilities, leading to more intelligent agents.
    In consequence, this paradigm has inspired many applications, such as mathematical reasoning \cite{wang2023math-shepherd,deepseekproverv2-2025}, code generation \cite{zhu2024deepseek-coder-v2,zhou2025refinecoder}, multimodality \cite{liu2025visualrft}, agents \cite{xia2025agentrm,openai-deepresearch}, and embodied AI \cite{zhang2025embodied-reasoner,zhao2025embodied-r}.

\begin{figure*}[!t]
    \centering
    \includegraphics[width=\linewidth]{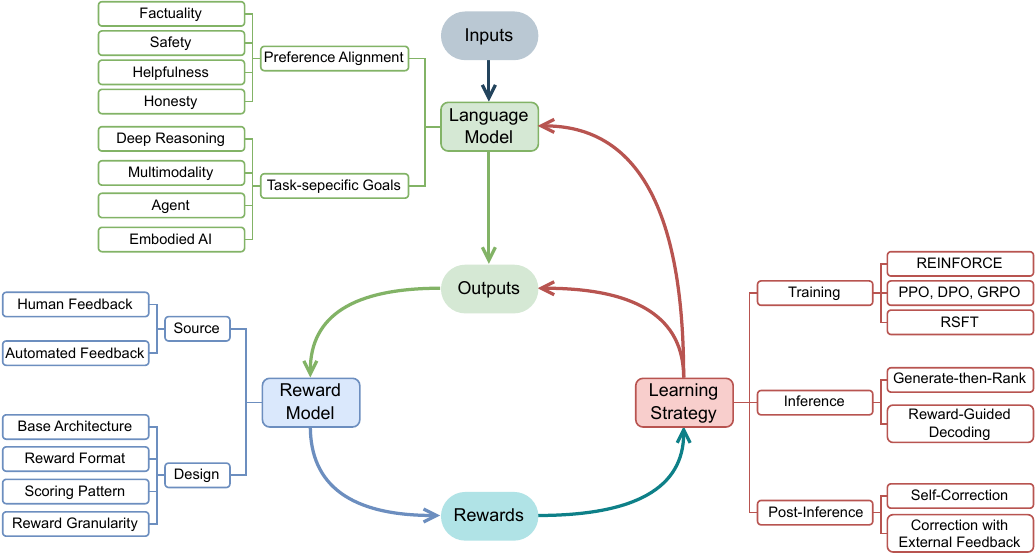}
    \caption{
        A unified framework of learning from rewards.
        The language model generates outputs conditioned on the inputs;
        the reward model evaluates the outputs and provides reward signals based on diverse feedback sources and design choices;
        the learning strategy leverages the rewards to either fine-tune the language model or refine the outputs.
        This learning-from-rewards paradigm aims to fulfill preference alignment and task-specific goals.
        The learning strategy can occur at the training, inference, or post-inference stages.
    }
    \label{fig_framework}
\end{figure*}

    Due to this growing prevalence,
    we in this paper comprehensively review the \textit{learning from rewards} for LLMs.
    We first introduce a taxonomy that categorizes existing works with a unified conceptual framework regarding reward models and learning strategies.
    Then we review representative techniques across three main stages of LLMs:
    \textit{training with rewards}, \textit{inference with rewards}, and \textit{post-inference with rewards}.
    We additionally summarize recent reward model benchmarks and finally conclude by outlining key challenges and promising directions for future research.

\section{A Taxonomy of Learning from Rewards for LLMs}
    We first introduce a unified conceptual framework that captures the key components and interactions to understand learning from rewards systemically.
    Building upon this framework, we categorize the primary dimensions along which existing methods vary:
    \begin{inparaenum}[(i)]
        \item \textbf{Reward Source};
        \item \textbf{Reward Model};
        \item \textbf{Learning Stage};
        \item \textbf{Learning Strategies}.
    \end{inparaenum}
    Each dimension reflects a distinct aspect of how reward signals are acquired, represented, and utilized in language models.

    \subsection{A Unified Conceptual Framework}
        We present a unified conceptual framework for learning from rewards in \Cref{fig_framework}.
        It abstracts the key components and interactions involved in learning from rewards for language models.
        In this framework, the \textit{language model} generates outputs conditioned on the inputs;
        the reward model then provides rewards to evaluate the output quality;
        the learning strategy leverages the reward signals to update the language model or adjusts the outputs.

        \paragraph{Language Model.}
            A language model $\mathcal{M}: \mathcal{X} \rightarrow \mathcal{Y}$
            generates an output $\hat{y} \in \mathcal{Y}$ given an input $x \in \mathcal{X}$.
            This formulation covers a wide range of tasks,
            such as question answering, summarization, and image captioning.

        \paragraph{Reward Model.}
            A reward model evaluates the quality of an output $\hat{y}$ given an input $x$ and produces a reward signal $r$
            that reflects desired properties, such as helpfulness, safety, or task-specific correctness.
            In different contexts, a reward model may be referred to as a verifier and an evaluator.
            We emphasize that here we adopt a broad definition of the reward model: it can be model-based or model-free.
            We will discuss these later.

        \paragraph{Learning Strategy.}
            A learning strategy uses reward signals to adjust the behavior of the language model.
            Here we consider both the training-based (updating model parameters) and training-free strategies (directly refining model outputs).

\begin{figure*}
    \centering
    \includegraphics[width=\linewidth]{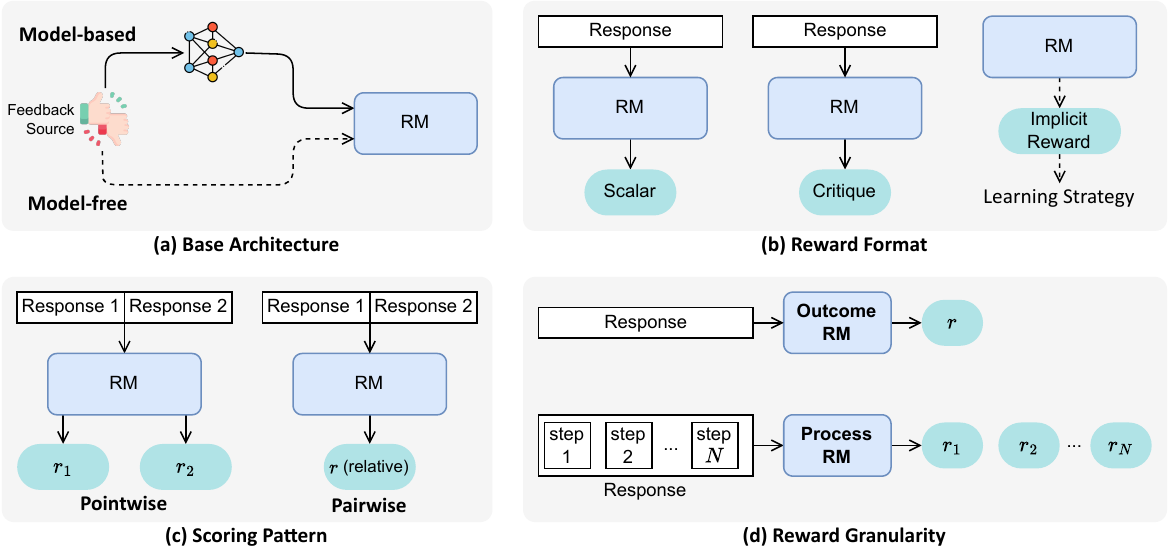}
    \caption{
        Reward Model (RM) design dimensions:
        (a) Base Architecture (Model-based and Model-free);
        (b) Reward Format (Scalar, Critique, and Implicit);
        (c) Scoring Pattern (Pointwise and Pairwise);
        (d) Reward Granularity (Outcome and Process).
    }
    \label{fig_reward_model_design}
\end{figure*}

    \subsection{Reward Source}
        Reward signals originate from two primary sources: \textbf{Human Feedback} and \textbf{Automated Feedback}.
        Each offers trade-offs in terms of reliability, scalability, and cost.
        We introduce them respectively as follows.

        \paragraph{Human Feedback.}
            Human feedback provides high-quality reward signals grounded in human judgment and intent.
            It typically collects human annotations through pairwise comparisons between alternative model outputs, \eg chosen and rejected responses.
            The collected preference data can be used to train explicit reward models like RLHF \cite{ouyang2022rlhf} or directly fine-tune the language model like DPO \cite{rafailov2023dpo}.
            While effective, this approach is resource-intensive and may not scale easily across domains or tasks.

        \paragraph{Automated Feedback.}
            To reduce the cost of human annotations and scale up the reward model training, automated feedback has been increasingly explored as an alternative.
            The automated feedback mainly includes
            \begin{inparaenum}[(\bgroup\bfseries i\egroup)]
                \item
                    \textbf{Self-rewarding}, where the language model critiques its own outputs \cite{yuan2024self-rewarding,wang2024self-taught};
                \item
                    \textbf{Trained Models}, such as powerful LLMs following the LLM-as-a-Judge design \cite{bai2022constitutional,lee2023rlaif-rlhf};
                \item
                    \textbf{Predefined Rules}, such as accuracy and format rules used in DeepSeek-R1 \cite{shao2024deepseekmath,deepseekai2025deepseek-r1}.
                \item
                    \textbf{Knowledge}, such as structured knowledge base or Wikipedia \cite{peng2023llmaugmenter,tian2023factuality}.
                \item
                    \textbf{Tools}, such as program compilers and interactive systems \cite{le2022coderl,liu2023rltf-unittest}.
            \end{inparaenum}
        The automated feedback enables scalable reward generation but may introduce limitations in interpretability, generality, and alignment quality.

    \subsection{Reward Model}
        Reward models are the central foundation of learning from rewards.
        As shown in \Cref{fig_reward_model_design},
        we organize the design space of reward model into four key dimensions:
        \begin{inparaenum}[(i)]
            \item \textbf{Base Architecture};
            \item \textbf{Reward Format};
            \item \textbf{Scoring Pattern};
            \item \textbf{Reward Granularity}.            
        \end{inparaenum}

        \paragraph{Base Architecture.}
            As shown in \refSubfigure{fig_reward_model_design}{a}, this refers to the base architecture of a reward model.
            Here we consider a broad view of reward models, including both model-based and model-free architectures.
            \begin{itemize}[leftmargin=*, itemsep=0pt]
                \item
                    \textbf{Model-based Architecture}.
                    A dedicated reward model is trained to evaluate outputs.
                    Common variants include
                    \begin{inparaenum}[(\bgroup\bfseries a\egroup)]
                        \\ \item
                            \textbf{Scalar Reward Models}.
                            These models assign a scalar score to a candidate response, indicating its quality.
                            Typically, they are built upon Transformer backbones (\eg GPT or BERT variants) with a value head that outputs scalars.
                            They are trained with preference data via pairwise ranking losses such as the Bradley-Terry loss \cite{nakano2021webgpt,ouyang2022rlhf,liu2024skywork}.
                        \\ \item
                            \textbf{Generative Reward Models}.
                            These models generate natural language critiques as reward signals.
                            They commonly follow LLM-as-a-Judge with general models \cite{zheng2023judging-llm-as-a-judge} or training specialized models \cite{li2023generative-judge,cao2024compassjudger,ye2024con-j,mcaleese2024criticgpt-catch-bugs}.
                            They have become more popular recently
                            because they can leverage the deep reasoning capabilities of large reasoning models and provide finer-grained supervision \cite{huang2025think-judge,guo2025reward-reasoning-model}.
                        \\ \item
                            \textbf{Semi-scalar Reward Models}.
                            These models combine scalars with critiques, offering both quantitative and qualitative assessment \cite{yu2024critic-rm,zhang2025mm-rlhf}.
                            Their architectures usually involve two heads, one for scalar rewards and another for critique rewards.
                    \end{inparaenum}
                \item
                    \textbf{Model-free Architecture}.
                    Instead of an explicit reward model, model-free approaches derive reward signals directly from diverse feedback sources, such as preference data, tools, or knowledge.
                    The resulting rewards can be scalar, critique, or implicit signals.
                    For example, DPO \cite{rafailov2023dpo} circumvents the need to train a reward model by directly aligning the language model with preference data through fine-tuning.
                    Similarly, GRPO \cite{deepseekai2025deepseek-r1} adopts rule-based rewards from handcrafted constraints and task-specific heuristics.
            \end{itemize}

                Model-based and model-free approaches each present distinct trade-offs in reward specification and practical applicability.
                Model-based approaches provide flexible and generalizable reward evaluation. Once trained, reward models can be reused across tasks and enable iterative optimization. However, they require costly preference data, are prone to overfitting, and may introduce bias or reward hacking issues.
                Model-free methods avoid training a separate reward model, offering a simpler, sample-efficient, and usually more stable pipeline.
                However, they are typically task-specific, lack generalization, and offer limited flexibility for downstream reuse.

                    In order to align with previous literature,
                    \textbf{we hereafter refer to the reward model as the model-based by default}.

        \paragraph{Reward Format.}
            As shown in \refSubfigure{fig_reward_model_design}{b}, this describes the specific format of reward signals:
            \begin{itemize}[leftmargin=*, itemsep=0pt]
                \item
                    \textbf{Scalar Rewards}, numerical scores that quantify the quality of model outputs.
                    They are the most commonly used format due to their simplicity and compatibility with learning strategies such as reinforcement learning.
                    Their limitation lies in the sparsity and interpretability.
                \item
                    \textbf{Critique Rewards}, natural language feedback that evaluates the quality of outputs \cite{saunders2022self-critiquing-assisting,kwon2023reward-design-language-models},
                    such as \textit{``The score of this response is 3 out of 5''}.
                    They are more expressive and interpretable than scalar rewards, enabling finer-grained guidance, but they may require additional processing to be used in certain learning algorithms.
                \item
                    \textbf{Implicit Rewards} are signals implicitly embedded in the source without explicit supervision, such as preference data in DPO \cite{rafailov2023dpo,meng2024simpo}.
                    This format simplifies the implementation but places more burden on the learning strategies to infer appropriate optimization signals.
            \end{itemize}

        \paragraph{Scoring Pattern.}
            As shown in \refSubfigure{fig_reward_model_design}{c}, 
            this dimension determines how responses are scored:
            \begin{itemize}[leftmargin=*, itemsep=0pt]
                \item
                    \textbf{Pointwise Scoring} assigns a score to each response independently.
                    It is the most widely used scoring pattern in reward models.
                \item
                    \textbf{Pairwise Scoring} compares response pairs and selecting the preferred one.
                    The pairwise scoring can be expressed as a scalar score indicating relative preference or a natural language critique such as \textit{``Response 1 is better than Response 2''}.
            \end{itemize}

        \paragraph{Reward Granularity.}
            As shown in \refSubfigure{fig_reward_model_design}{d}, 
            we identify two kinds of reward granularity:
            reward granularity reflects the level of resolution at which feedback is provided:
            \begin{itemize}[leftmargin=*, itemsep=0pt]
                \item
                    \textbf{Outcome Reward Models} evaluate the holistic quality of outputs, treating it as a single unit.
                \item
                    \textbf{Process Reward Models} evaluate intermediate steps within the reasoning process of outputs, enabling fine-grained supervision during generation \cite{lightman2023letsverify,wang2023math-shepherd}.
            \end{itemize}

    \subsection{Learning Stage}
        Learning from rewards can occur at different stages of the language model lifecycle,
        including \textbf{Training}, \textbf{Inference}, and \textbf{Post-Inference}.
        \begin{itemize}[leftmargin=*, itemsep=0pt]
            \item
                \textbf{Training with Rewards.}
                At the training stage,
                reward signals can be transformed into optimization signals by training algorithms to fine-tune the language model,
                which is the most extensively explored in the literature.
                It can support post-training alignment with human preference \cite{ouyang2022rlhf,bai2022constitutional}
                and test-time scaling by eliciting the language models' deep reasoning capabilities through long Chain-of-Thoughts (CoT) \cite{deepseekai2025deepseek-r1}.
            \item
                \textbf{Inference with Rewards.}
                During inference, reward signals can guide the decoding of model outputs without modifying model parameters.
                This enables test-time scaling by searching in a larger decoding space, such as \textit{Best-of-N} and tree search \cite{cobbe2021training-verifiers,snell2025scaling}.
            \item
                \textbf{Post-Inference with Rewards.}
                This stage uses rewards to refine model outputs after generation without modifying model parameters.
                Post-inference with rewards also supports test-time scaling by iteratively refining the outputs \cite{shinn2023reflexion}.
        \end{itemize}

    \subsection{Learning Strategy}
        Various learning strategies have been developed to incorporate reward signals to steer model behavior.
        These strategies are commonly divided into two types: \textbf{Training-based} and \textbf{Training-free}.

        \begin{itemize}[leftmargin=*, itemsep=0pt]
            \item
                \textbf{Training-based Strategies.}
                Training-based strategies optimize the language model by converting reward signals into gradient-based updates.
                The optimization mainly depends on Reinforcement Learning (RL) where language models act as policy models, or Supervised Fine-Tuning (SFT).
                Representative examples include
                Proximal Policy Optimization \cite[PPO, ][]{schulman2017proximal,ouyang2022rlhf},
                Direct Preference Optimization \cite[DPO, ][]{rafailov2023dpo,meng2024simpo},
                Group Relative Policy Optimization \cite[GRPO, ][]{shao2024deepseekmath},
                and Rejection-Sampling Fine-Tuning \cite[RSFT, ][]{nakano2021webgpt,yuan2023scaling-math-rsft,dong2023raft}
            \item
                \textbf{Training-free Strategies.}
                Training-free strategies leverage reward signals to guide or refine model outputs without updating the language model parameters.
                They include generate-then-rank, such as \textit{Best-of-N} \cite{cobbe2021training-verifiers,lightman2023letsverify},
                reward-guided decoding \cite{deng2023rad,khanov2024args}, and post-inference correction \cite{shinn2023reflexion,pan2023logic-lm}.
                These methods provide a relatively lightweight mechanism for improving model outputs, and some are highly compatible with various model architectures. They are particularly useful when model fine-tuning is infeasible or computationally expensive.
        \end{itemize}

        The above presents a detailed taxonomy of learning from rewards for LLMs.
        We will review the representative studies across the three learning stages: training, inference, and post-inference with rewards in the following \Cref{sec_training_rewards,sec_inference_rewards,sec_post-inference_rewards}.

\begin{figure*}[!ht]
    \centering
    \includegraphics[width=0.9\linewidth]{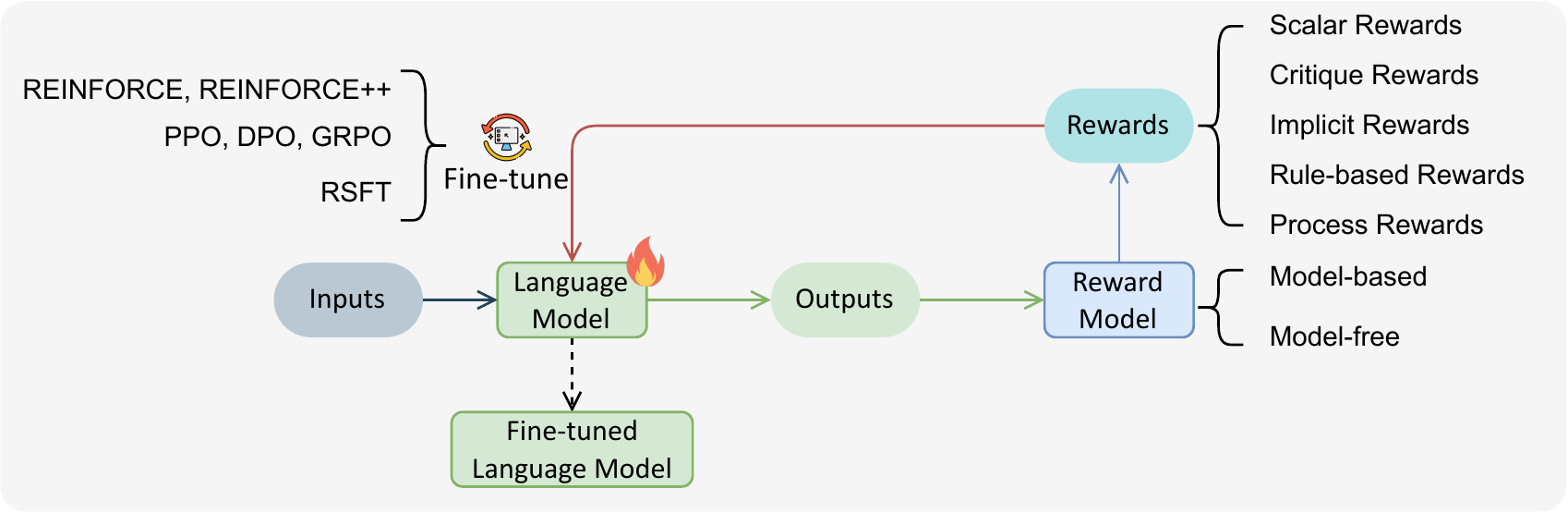}
    \caption{
        Illustration of \textbf{Training with Rewards}.
        Based on the reward model design,
        we mainly focus on scalar rewards, critique rewards, implicit rewards, rule-based rewards, and process rewards.
        These reward signals are used to fine-tune the language model through reinforcement learning algorithms or supervised fine-tuning.
    }
    \label{fig_training_rewards}
\end{figure*}

\section{Training with Rewards} \label{sec_training_rewards}
    In this section, we introduce the methods that incorporate rewards into the training phase of large language models (LLMs).
    These methods contribute to post-training scaling for preference alignment
    and test-time scaling by eliciting long Chain-of-Thoughts (CoT) abilities.
    We begin with a brief review of the primary training algorithms and 
    then categorize existing methods by reward design:
    \begin{inparaenum}[(\bgroup\bfseries i\egroup)]
        \item Training with Scalar Rewards;
        \item Training with Critique Rewards;
        \item Training with Implicit Rewards;
        \item Training with Rule-based Rewards;
        \item Training with Process Rewards;
    \end{inparaenum}
    The first three form the core categories, and the latter two, though conceptually overlapped with the former, are separately presented due to their emerging importance.

    The primary training algorithms depend on Reinforcement Learning (RL) or Supervised Fine-Tuning (SFT):
    \begin{itemize}[leftmargin=*,itemsep=0pt,topsep=2pt]
        \item \textbf{REINFORCE} \cite{sutton1999policy}. REINFORCE is a fundamental policy gradient algorithm that optimizes expected rewards by estimating gradients through sampled actions and their observed rewards.
        \item \textbf{PPO} \cite{schulman2017proximal}. PPO is a widely used reinforcement learning algorithm. It constrains policy updates through clipped objectives to balance learning speed and stability.
        \item \textbf{DPO} \cite{rafailov2023dpo}. DPO is a direct preference optimization method that learns from preference data without explicitly modeling rewards.
        \item \textbf{GRPO} \cite{shao2024deepseekmath}. GRPO directly uses the average reward of multiple sampled rollouts as the baseline, which avoids the reward and value model of PPO.
        \item \textbf{REINFORCE++} \cite{hu2025reinforce++}. REINFORCE++ is a variant of the REINFORCE algorithm that incorporates key techniques from PPO without a critic network.
        \item \textbf{RSFT} \cite[Rejection-Sampling Fine-Tuning, ][]{nakano2021webgpt,yuan2023scaling-math-rsft}. RSFT samples high-reward data offline to construct training datasets for fine-tuning.
    \end{itemize}

\begin{figure*}[!t]
    \centering
    \begin{forest}
        forked edges,
        for tree={
            grow=east,
            reversed=true,
            anchor=base west,
            parent anchor=east,
            child anchor=west,
            base=center,
            font=\small,
            rectangle,
            draw=black,
            rounded corners,
            align=center,
            text centered,
            minimum width=4em,
            edge+={darkgray, line width=0.5pt},
            s sep=8pt,
            line width=0.5pt,
            ver/.style={rotate=90, child anchor=north, parent anchor=south, anchor=center, minimum width=19em, fill=gray!10},
            leaf/.style={font=\scriptsize, align=left, draw=none, inner xsep=8pt}, %
            leaf2/.style={text width=10em, font=\scriptsize, draw=none, inner xsep=8pt} %
        },
        where level=1{text width=6.5em,align=center,font=\scriptsize}{},
        where level=2{text width=7.5em, align=center, font=\scriptsize}{},
        where level=3{text width=18em, align=left,font=\scriptsize}{},
        where level=4{align=left,font=\scriptsize}{},
        [
            \textbf{Training with Rewards}, ver
            [
                Training with \\Scalar Rewards \\ \Cref{sec_training_scalar_rewards}, fill=training-fill, draw=training-border
                [
                    Scalar Rewards \\from Human Feedback, fill=training-fill, draw=training-border
                    [
                     \textit{RLHF} \cite{ouyang2022rlhf};
                     \textit{Safe RLHF} \cite{dai2023safe-rlhf}; \\
                     \textit{Fine-Grained RLHF} \cite{wu2023fine-grained-rlhf};
                     \textit{Fact-RLHF} \cite{sun2023fact-rlhf}; \\
                     \textit{Skywork-Reward} \cite{liu2024skywork};
                     \textit{ImageReward} \cite{xu2023imagereward}; \\
                     \textit{RAHF} \cite{liang2024rich-texttoimage};
                     \textit{LiFT} \cite{wang2024lift}
                     ,leaf, fill=training-fill, draw=none
                    ]
                ]
                [
                    Scalar Rewards \\from Automated Feedback, fill=training-fill, draw=training-border
                    [
                     \textit{RLAIF} \cite{bai2022constitutional};
                     \textit{Self-Taught} \cite{wang2024self-taught};\\
                     \citet{dutta2024rlaif-codeapi};
                     \textit{VLM-RLAIF} \cite{ahn2024vlm-rlaif}; \\
                     \textit{RLTF} \cite{liu2023rltf-unittest};
                     \textit{RLEF} \cite{gehring2024rlef}; \\
                     \textit{StepCoder} \cite{dou2024stepcoder};
                     \textit{RLEF} \cite{gehring2024rlef}\\
                     , leaf, fill=training-fill, draw=none
                    ]
                ]
            ]
            [
                Training with \\Critique Rewards \\ \Cref{sec_training_critique_rewards}, fill=training-fill, draw=training-border
                [
                    \textit{Auto-J} \cite{li2023generative-judge};
                    \textit{CompassJudger-1} \cite{cao2024compassjudger}; \\
                    \textit{Con-J} \cite{ye2024con-j};
                    \textit{GemRM} \cite{mahan2024genrm}; \\
                    \textit{LLaVA-Critic} \cite{xiong2024llava-critic};
                    \textit{DeepSeek-GRM} \cite{liu2025deepseek-grm}; \\
                    \textit{RRM} \cite{guo2025reward-reasoning-model};
                    \textit{Think-J} \cite{huang2025think-judge}; \\
                    \textit{UnifiedReward-Think} \cite{wang2025unifiedreward-think}
                    , leaf, fill=training-fill, draw=none, text width=
                ]
            ]
            [
                Training with \\Hybrid Rewards \\ \Cref{sec_training_critique_rewards}, fill=training-fill, draw=training-border
                [
                    \textit{Critic-RM} \cite{yu2024critic-rm};
                    \textit{MM-RLHF} \cite{zhang2025mm-rlhf}; \\
                    \textit{Critique-GRPO} \cite{zhang2025critique-grpo}
                    , leaf, fill=training-fill, draw=none, text width=
                ]
            ]
            [
                Training with \\Implicit Rewards \\ \Cref{sec_training_implicit_rewards}, fill=training-fill, draw=training-border
                [
                    Implicit Rewards \\from Human Feedback, fill=training-fill, draw=training-border
                    [
                        DPO \cite{rafailov2023dpo};
                        \textit{SimPO} \cite{meng2024simpo}; \\
                        \textit{RLHF-V} \cite{yu2024rlhf-v};
                        \textit{UnifiedRM} \cite{wang2025unified}; \\
                        \textit{RAFT} \cite{dong2023raft};
                        \textit{ReST} \cite{gulcehre2023rest}; \\
                        \textit{RSO} \cite{liu2024rso};
                        \textit{RRHF} \cite{yuan2023rrhf-rank-responses-align}
                        , leaf, fill=training-fill, draw=none
                    ]
                ]
                [
                    Implicit Rewards \\from Automated Feedback, fill=training-fill, draw=training-border
                    [
                        \textit{Self-Rewarding} \cite{yuan2024self-rewarding}; \\
                        \textit{Meta-Rewarding} \cite{wu2024metarewarding}; \\
                        \textit{SCPO} \cite{prasad2024scpo-self-consistency};
                        \citet{zhang2025process-self-rewarding}; \\
                        \textit{PFPO} \cite{jiao2024pseudo-feedback-PFPO};
                        \textit{HA-DPO} \cite{zhao2023ha-dpo};\\
                        \citet{tian2023factuality};
                        \textit{FLAME} \cite{lin2024flame}; \\
                        \textit{TRICE} \cite{qiao2023making};
                        \textit{CodeLutra} \cite{tao2024codelutra}; \\
                        , leaf, fill=training-fill, draw=none
                    ]
                ]
            ]
            [
                Training with \\Rule-based Rewards \\ \Cref{sec_training_rule-based_rewards}, fill=training-fill, draw=training-border
                [
                    DeepSeek-Math \cite{shao2024deepseekmath};
                    DeepSeek-R1 \cite{deepseekai2025deepseek-r1}; \\
                    \textit{DAPO} \cite{yu2025dapo};
                    \textit{Open-R1} \cite{huggingface2025open-r1}; \\
                    \textit{Logic-RL} \cite{xie2025logicrl};
                    \textit{Visual-RFT} \cite{liu2025visualrft}; \\
                    \textit{CLS-RL} \cite{li2025clsrl};
                    \textit{R1-VL} \cite{zhang2025r1-vl}; \\
                    \textit{RefAlign} \cite{zhao2025refalign}
                    , leaf, fill=training-fill, draw=none, text width=
                ]
            ]
            [
                Training with \\Process Rewards \\ \Cref{sec_training_process_rewards}, fill=training-fill, draw=training-border
                [
                    Process Rewards \\from Human Feedback, fill=training-fill, draw=training-border
                    [
                     \citet{uesato2022solving}; \citet{lightman2023letsverify}
                     , leaf, fill=training-fill, draw=none
                    ]
                ]
                [
                    Process Rewards \\from Automated Feedback, fill=training-fill, draw=training-border
                    [
                        \textit{WizardMath} \cite{luo2023wizardmath};
                        \textit{ActPRM} \cite{duan2025actprm-efficient}; \\
                        \textit{Math-Shepherd} \cite{wang2023math-shepherd};
                        \textit{PQM} \cite{li2024pqm};\\
                        \textit{OmegaPRM} \cite{luo2024omegareward};
                        \textit{HRM} \cite{wang2025hierarchical-rm};\\
                        \textit{PRIME} \cite{cui2025prime-process-implicit-reward};
                        \textit{OREAL} \cite{lyu2025oreal};\\
                        \textit{GenPRM} \cite{zhao2025genprm};
                        \textit{R-PRM} \cite{she2025reasoning-prm};\\
                        \textit{ThinkPRM} \cite{khalifa2025thinkprm};
                        \textit{M-STAR} \cite{liu2024diving-multimodal}
                        , leaf, fill=training-fill, draw=none
                    ]
                ]
            ]
        ]
    \end{forest}
    \caption{
        Overview of \textbf{Training with Rewards}.
    }
    \label{forest_training_rewards}
\end{figure*}
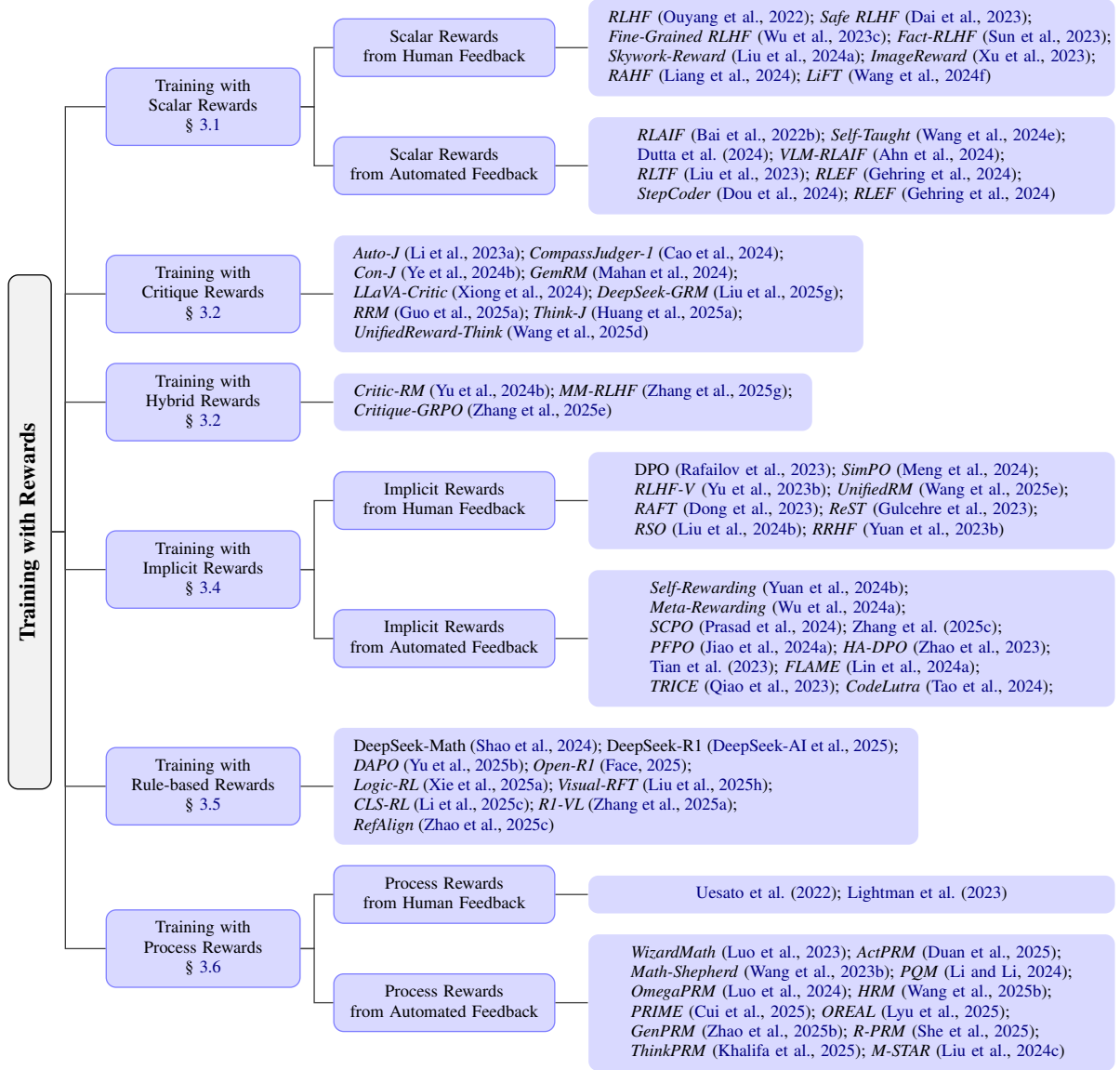

    \subsection{Training with Scalar Rewards} \label{sec_training_scalar_rewards}
        Training the language model with \textit{scalar rewards} is the most extensively studied strategy in the literature.
        Most methods derive scalar rewards by training a dedicated reward model, while some extract rewards directly from the source without training a model (\ie model-free).
        We introduce the methods based on the reward source human and automated feedback as follows.

        \paragraph{Scalar Rewards from Human Feedback.}
            Human feedback is a key source for constructing scalar rewards.
            The most prominent example is Reinforcement Learning from Human Feedback (RLHF) \cite{ziegler2019finetuning,ouyang2022rlhf,bai2022training,glase2022sparrow}.
            RLHF trains a scalar reward model on human preference data (pairwise comparisons with chosen and rejected responses).
            The reward models commonly adopt the Transformer architecture with a value head that outputs scalars,
            and their training objectives follow the Bradley-Terry loss \cite{bradley1952rank},
            which maximizes the reward differences between preferred and dispreferred outputs.
            The trained reward model assigns evaluative scalar scores to the model outputs, serving as a proxy for human judgment.
            For instance, \textit{Skywork-Reward} \cite{liu2024skywork} is a reward model trained on various high-quality human-labeled preference datasets.
            With the reward model, RLHF fine-tunes the language model through PPO to align it with human preferences, such as harmlessness and helpfulness.

            Various variants have been explored based on RLHF.
            \textit{Safe RLHF} \cite{dai2023safe-rlhf} emphasizes safety-centric alignment.
            \textit{Fine-Grained RLHF} \cite{wu2023fine-grained-rlhf} moves beyond holistic feedback to dense, segment-level fine-grained feedback regarding relevance, factual accuracy, and completeness.

            Training with scalar rewards has also been extended to multimodal tasks.
            \textit{Fact-RLHF} \cite{sun2023fact-rlhf} incorporates factuality-aware scalar rewards to reduce hallucinations of multimodal understanding.
            Others focus on multimodal generation tasks.
            \textit{HPS} and its variant \cite{wu2023hps,wu2023hpsv2} train a binary classifier as the reward model to evaluate generated image quality.
            \citet{lee2023aligning} trains a multimodal reward model based on human feedback and fine-tunes a diffusion model to improve
            image generation.
            \textit{ImageReward} \cite{xu2023imagereward} introduces a general-purpose reward model for text-to-image generation.
            It is trained on image preferences from human experts and then fine-tunes a diffusion model via a direct tuning algorithm.
            \textit{RAHF} \cite{liang2024rich-texttoimage} further enriches the reward signals with scalar scores, heatmaps, and keyword omissions to guide sample filtering for fine-tuning.
            \textit{LiFT} \cite{wang2024lift} applies a similar idea to the text-to-video task.

        \paragraph{Scalar Rewards from Automated Feedback.}
            While human feedback offers high-quality supervision, it is expensive and difficult to scale.
            To overcome this challenge, a growing body of work explores automated feedback as a substitute to provide scalar rewards.
            A prominent example is Reinforcement Learning from AI Feedback (RLAIF) \cite{bai2022constitutional}.
            RLAIF uses an LLM as a proxy judge to generate preference data following the idea of \textit{LLM-as-a-Judge} \cite{zheng2023judging-llm-as-a-judge,yu2025improve-llm-as-a-judge}.
            Like RLHF, RLAIF trains a scalar reward model on them and then uses it to fine-tune the language model.
            This pipeline substantially reduces reliance on expensive human annotations.
            \citet{lee2023rlaif-rlhf} have shown that RLAIF can achieve comparable performance with RLHF.

            Subsequent studies extend this strategy by incorporating diverse forms of automated feedback.
            \textit{Self-Taught} \cite{wang2024self-taught} prompts an LLM to synthesize contrastive synthetic pairs and uses them to fine-tune a reward model.
            \citet{dutta2024rlaif-codeapi} collect GPT-3.5's binary assessments about predefined code quality questions to train a reward model for code generation.
            \textit{VLM-RLAIF} \cite{ahn2024vlm-rlaif} trains a reward model with self-generated preference data for video-language tasks.
            It then uses the reward model to fine-tune a video-language model through PPO.
            Automated feedback can also come from various tools.
            \textit{RLTF} \cite{liu2023rltf-unittest} introduces an online reinforcement learning framework for code language models.
            It derives multi-granularity scalar rewards from real-time unit test execution results, and
            these rewards are integrated into a unified loss optimized via the REINFORCE algorithm \cite{sutton1999policy} to fine-tune the model.
            Similarly, \textit{StepCoder} \cite{dou2024stepcoder} designs scalar rewards based on compiler feedback, and it directly fine-tunes LLMs through token-level PPO with these rewards, with optimization restricted to executed code tokens.
            \textit{RLEF} \cite{gehring2024rlef} formulates code generation as a multi-turn decision-making process in an interactive environment.
            It fine-tunes the language model with the program execution results as scalar rewards through turn-level PPO and supports iterative refinement.

    \subsection{Training with Critique Rewards} \label{sec_training_critique_rewards}
        Besides scalar rewards, another line of work explores training with \textit{critique rewards}.
        This category commonly uses generative reward models to produce natural language critiques on the given response rather than predicting scalar scores.
        The generated critiques enjoy flexible scoring patterns: they could be pointwise scoring like \textit{``The score of this response is 3 out of 5"}
        or pairwise scoring like \textit{``Response 1 is better than Response 2"}.
        More importantly, the critiques can include explanations for the scoring and refinement suggestions for later improvements,
        which facilitates reward density and interpretability.

        Due to the above advantages, many studies work on generative reward models to produce critique rewards.
        \textit{Auto-J} \cite{li2023generative-judge} generates critiques that support pointwise and pairwise evaluation.
        It adopts GPT-4 to produce evaluation judgments as the training data.
        \textit{CompassJudger-1} \cite{cao2024compassjudger} and \textit{Con-J} \cite{ye2024con-j} follow a similar design.
        \textit{SFR-Judges} \cite{wang2024directjudgement} fine-tunes an LLM on the response deduction task to improve its judging ability.

        Because of the flexibility of generative models, recent studies concentrate on generalist reward models that support multiple tasks and scoring patterns.
        For instance, \textit{LLaVA-Critic} \cite{xiong2024llava-critic} is a trained reward model for multiple vision-language tasks, supporting both pointwise and pairwise scoring with natural language explanations.
        \textit{DeepSeek-GRM} \cite{liu2025deepseek-grm} is a generalist generative reward model that covers diverse scoring formats and NLP tasks and produces principles for its scoring.
        We discuss this promising future direction in \Cref{sec_future_directions}.

        In particular, generative reward models can work beyond critique generation.
        For example,  \textit{GenRM} \cite{zhang2024genrm} considers reward modeling as a token prediction task.
        Given a prompt like \textit{``Is the answer correct?''}, it uses the generation probability of answer indicator tokens, like \textit{Yes} or \textit{No}, as the scalar rewards.
        \citet{mahan2024genrm} follow a similar approach.

    \subsection{Training with Hybrid Rewards}
        Furthermore,
        several methods adopt hybrid rewards (or semi-scalar rewards) that include both scalars and critiques.
        Early attempts incorporate critiques into the training of a scalar reward model, such as \textit{CLoud} \cite{ankner2024critique-out-loud-rm}, \textit{MATH-Minos} \cite{gao2024math-minos-critics-bugs-math}, and \citet{ye2024improving-rm-synthetic-critiques}.
        \textit{Critic-RM} \cite{yu2024critic-rm} is a reward model with two heads, one for generating critiques and another for predicting scalars.
        It is trained on self-generated critiques of model outputs through a multi-task objective.
        \textit{Critique-GRPO} \cite{zhang2025critique-grpo} applies both scalar and critique rewards for reinforcement learning.
        \textit{MM-RLHF} \cite{zhang2025mm-rlhf} extends it to the multimodal field.
        It trains a semi-scalar reward model on the image, video understanding, and safety
        and then fine-tunes a multimodal language model with the trained reward model.

    \subsection{Training with Implicit Rewards} \label{sec_training_implicit_rewards}
        Rather than explicit scalars or critiques that evaluate the model outputs,
        another line of work adopts \textit{implicit rewards} for training.
        The reward signals are not provided directly but are implicitly embedded in the structure of the training data, such as curated preference data.
        We note that some methods may use a scalar reward model to construct the training data, but the reward model does not participate in fine-tuning language models.
        Instead, the reward signals for fine-tuning are implicitly encoded in the constructed training data, so we categorize them as training with implicit rewards.

        \paragraph{Implicit Rewards from Human Feedback.}
            A pioneering approach using implicit rewards from human feedback is Direct Preference Optimization (DPO) \cite{rafailov2023dpo}.
            DPO encodes implicit rewards via the log-likelihood difference between preferred and dispreferred responses and steers the generalization of the language model toward preferred ones.
            DPO proves that its objective is theoretically equivalent to optimizing the Bradley-Terry loss.
            As such, DPO effectively reduces complicated RLHF \cite{ouyang2022rlhf} into supervised fine-tuning, dramatically simplifying the alignment pipeline.

            Several variants have been proposed based on DPO to further simplify the training or expand its applicability.
            For instance, \textit{SimPO} \cite{meng2024simpo} eliminates the need for a reference model by directly optimizing preference pairs using a more straightforward and reference-free objective, further simplifying DPO.
            \textit{KTO} \cite{ethayarajh2024kto} models the reward signals implicitly by comparing the log-likelihood difference between the model output and a reference baseline.
            This difference is then transformed through a nonlinear value function to reflect human-like preferences.

            Some studies have extended training with implicit rewards in reasoning and multimodal areas.
            In mathematical reasoning, 
            \textit{Step-DPO} \cite{lai2024stepdpo} and \textit{Full-Step-DPO} \cite{xu2025fullstepdpo} build step-level preference data for DPO training to improve their reasoning abilities.
            In multimodal areas, \textit{mDPO} \cite{wang2024mdpo} proposes a multimodal extension of DPO.
            It introduces conditional and anchored preference objectives to better leverage visual inputs and preserve the likelihood of preferred responses in multimodal tasks.
            \textit{RLHF-V} \cite{yu2024rlhf-v} fine-tunes a multimodal LLM through DPO with human preferences regarding trustworthiness.
            \textit{UnifiedRM} \cite{wang2025unified} trains a unified multimodal reward model on human preferences for several key tasks, including image/video understanding and generation.
            Later, it builds preference data through the trained reward model for DPO training.
            Beyond language models, these techniques are applied to diffusion models due to their simplicity, such as \textit{Diffusion-DPO} \cite{wallace2024diffusion}.

            Apart from DPO, another line of work follows a Rejection-Sampling Fine-Tuning (RSFT) scheme.
            They typically select high-quality samples from a large number of candidate data for supervised fine-tuning on the language model.
            While reward models are usually used to evaluate the candidates, they are not involved in the actual fine-tuning;
            thus RSFT methods are within the category of training with implicit rewards. 
            \textit{RAFT} \cite{dong2023raft} trains a reward model on the HH-RLHF dataset \cite{bai2022training} and uses it to select high-reward samples.
            The model is then directly fine-tuned on these samples.
            \textit{ReST} \cite{gulcehre2023rest} adopts an iterative framework: it samples responses from the current model, filters high-quality ones with a reward model, and fine-tunes the model on the filtered set.
            \textit{RSO} \cite{liu2024rso} trains a reward model on human preferences and then uses it to generate new preference data via statistical rejection sampling from an estimated optimal policy.
            The language model is fine-tuned on the new preference data.
            Especially, some approaches directly depend on the ranking loss.
            \textit{RRHF} \cite{yuan2023rrhf-rank-responses-align} uses a reward model to rank multiple candidate responses.
            Then it fine-tunes the language model by jointly optimizing a ranking loss and a SFT loss on these ranked responses.

        \paragraph{Implicit Rewards from Automated Feedback.}
            Implicit rewards can originate from diverse automated feedback as well, such as AI feedback, feedback from external knowledge, and feedback from external tools.

            \begin{itemize}[leftmargin=*,itemsep=0pt,topsep=2pt]
            \item
                \textbf{AI Feedback.}
                AI feedback is a common source of implicit rewards, including self-rewarding and other trained models.
                \textit{Self-Rewarding} \cite{yuan2024self-rewarding} leverages the language model to evaluate its own outputs with a prompt like \textit{Rate this answer from 1 to 5}.
                Then it constructs preference data with high- and low-rated responses to fine-tune the LLM through iterative DPO.
                \textit{Meta-Rewarding} \cite{wu2024metarewarding} builds on this idea: the language model judges its outputs and evaluates the quality of its own judgments.
                This creates a two-level preference structure and enables joint optimization of acting and judging capabilities.
                \citet{zhang2025process-self-rewarding} extend self-rewarding to the process-level. It constructs step-wise preference data via self-rewarding to improve the language model's reasoning ability.

                Instead of direct self-assessment, some methods depend on self-consistency to model implicit rewards.
                \textit{SCPO} \cite{prasad2024scpo-self-consistency} samples multiple responses for each input and selects the consistent responses as preferred ones.
                Similarly, it then fine-tunes the language model through DPO on these constructed preference data.
                Similarly, \textit{PFPO} \cite{jiao2024pseudo-feedback-PFPO} constructs preference data by verifying model-generated solutions against the test cases derived by self-consistency.

                A list of extended studies focuses on the multimodal field.
                \textit{HA-DPO} \cite{zhao2023ha-dpo} evaluates model responses of image understanding with another trained model as a hallucination detector to produce preference data.
                With these preference data, it fine-tunes the multimodal language models via DPO to reduce hallucination.
                Many studies follow a similar way to this, such as \textit{POVID} \cite{zhou2024povid}, \textit{AMP} \cite{zhang2024amp} and \textit{RLAIF-V} \cite{yu2024rlaif-v}, and \textit{BPO} \cite{pi2024strengthening-bpo}.

            \item
                \textbf{Feedback from Knowledge and Tools.}
                External knowledge and tools can provide feedback to model implicit rewards.
                \citet{tian2023factuality} construct preference pairs by checking whether model outputs are supported by Wikipedia.
                These preference data are used to fine-tune a language model toward more factually accurate generations via DPO.
                \textit{FLAME} \cite{lin2024flame} follows a similar way.
                It fine-tunes the language model on factuality-aware preference data via both SFT and DPO to maintain the instruction-following ability.
                \textit{TRICE} \cite{qiao2023making} uses tool execution feedback to rank sampled outputs, and the language model is trained with a ranking loss combined with SFT on ranked samples toward accurate and appropriate tool usage.
                \textit{CodeLutra} \cite{tao2024codelutra} categorizes generated code into successful and failed samples based on code execution outcomes.
                It constructs preference data with these samples and fine-tunes the model by combining DPO and SFT objectives.
                \citet{xiong2025rewardingcorrection} leverages SymPy as the verifier to construct mathematical reasoning trajectory pairs for DPO training.
        \end{itemize}

        Note that some methods use hybrid automated feedback sources.
        For instance, \textit{RefineCoder} \cite{zhou2025refinecoder} incorporates critiques from self-rewarding, Elo ranking, and code execution feedback to build the training data for fine-tuning the language model.

    \subsection{Training with Rule-based Rewards}  \label{sec_training_rule-based_rewards}
        Recently,
        training with \textit{rule-based rewards}%
        ~\footnote{While rule-based rewards may overlap with previous scalar or critique rewards, here we present them separately due to their emerging significance.}
        has gained prominence due to the success of DeepSeek-R1 \cite{deepseekai2025deepseek-r1}.
        Rule-based rewards are derived by verifying outputs against specific rules, such as format constraints and evaluation metrics.
        By leveraging reinforcement learning with rule-based rewards,
        DeepSeek-R1 demonstrates that language models can acquire long Chain-of-Thoughts (long CoT) abilities for test-time scaling.
        This enables it to solve complex tasks such as mathematics and coding and show self-reflection and self-correction behaviors.
        This phenomenon, characterized by a sudden emergence of advanced reasoning capabilities, is referred to as RL scaling or the ``Aha moment".
        Note that the literature sometimes refers to rule-based rewards as \textit{verifiable rewards/outcomes} due to their clean evaluation criteria,
        for example Reinforcement Learning with Verifiable Rewards (RLVR).

        In detail, DeepSeek-R1 \cite{deepseekai2025deepseek-r1} defines two types of rule-based rewards:
        \begin{inparaenum}[(i)]
            \item
                \textit{Accuracy Rewards}, whether the output is factually or functionally correct, \eg the correctness of math solutions or code passing unit tests;
            \item
                \textit{Format Rewards}, which ensure the output follows specific structural constraints, such as containing special tokens \texttt{<think>}, \texttt{</think>}, \texttt{<answer>}, and \texttt{</answer>} to encourage long CoT reasoning.
        \end{inparaenum}
        With these rule-based rewards, it fine-tunes the language model through the RL algorithm GRPO \cite{shao2024deepseekmath}.
        GRPO eliminates the dependence on the reward and value model in PPO and the preference data construction in DPO.

        Later, many following studies have been proposed.
        \textit{DAPO} \cite{yu2025dapo} and \textit{Open-R1} \cite{huggingface2025open-r1} introduce open-source training frameworks,
        and some extended GRPO algorithms are introduced \cite{xu2025downsampling-rollouts-rl,zuo2025test-time-rl,feng2025grpo-curriculum-sampling,zhang2025right-question-half-answer}.
        \textit{Logic-RL} \cite{xie2025logicrl} combines accuracy and format rewards to improve logical reasoning through REINFORCE++ \cite{hu2025reinforce++}.
        \textit{Visual-RFT} \cite{liu2025visualrft} introduces a set of rule-based rewards for visual tasks, such as Intersection over Union (IoU), confidence-based scoring, and format compliance to fine-tune a vision language model via GRPO.
        These visual tasks include image classification, reasoning grounding, and object detection.
        Similarly, \textit{CLS-RL} \cite{li2025clsrl} designs rule-based rewards for image classification.
        \textit{R1-VL} \cite{zhang2025r1-vl} proposes StepGRPO that extends GRPO into step-level multimodal reasoning.

        Recently, \citet{shao2025spurious-rewards} challenges reinforcement learning with rule-based rewards:
        they find that random and incorrect rewards can work as correct rewards for some LLMs like QWen2.5-Math.

    \subsection{Training with Process Rewards}  \label{sec_training_process_rewards}
        The aforementioned strategies mostly depend on \textit{outcome rewards}---evaluating the holistic quality of outputs.
        An emerging line of work focuses on training with \textit{process rewards},%
        ~\footnote{Similarly, we discuss process rewards separately due to their increasing prevalence although they may overlap with previous scalar or critique rewards.}
        which evaluate the intermediate steps in a model's reasoning trajectory, such as the steps in mathematical problem solving.
        As shown in \refSubfigure{fig_reward_model_design}{d}, these methods commonly employ a Process Reward Model (PRM) to assess the intermediate steps.
        By providing step-level feedback, these methods enable more fine-grained supervision, which especially benefits complex reasoning tasks where intermediate reasoning directly influences the final result, such as mathematics and code.

        \paragraph{Process Rewards from Human Feedback.}
            Early studies leverage human annotations to train PRMs.
            For instance, \citet{uesato2022solving,lightman2023letsverify} train PRMs using human annotations on intermediate mathematical reasoning steps.
            \citet{uesato2022solving} then use the trained PRM to fine-tune the language model via reinforcement learning to improve its math reasoning.

        \paragraph{Process Rewards from Automated Feedback.}
            A key limitation of PRMs is their reliance on labor-intensive step-level human annotations.
            To address this limitation, recent efforts propose incorporating automated feedback to supervise PRMs training at scale.

            One major direction leverages strong LLMs to generate step-level annotations.
            For example, \textit{WizardMath} \cite{luo2023wizardmath} uses GPT-4 to label intermediate steps in math solutions and fine-tunes the language model through step-by-step PPO.
            \textit{ActPRM} \cite{duan2025actprm-efficient} similarly uses a strong LLM to annotate step-level correctness of reasoning trajectories, which enables step-by-step DPO fine-tuning.

            Alternatively, other methods estimate process rewards without explicit annotations.
            \textit{Math-Shepherd} \cite{wang2023math-shepherd} applies Monte Carlo estimation to infer step-level scores of mathematical reasoning based on the likelihood of reaching a correct final answer.
            It uses these estimated annotations to train a PRM and then fine-tunes the language model through step-by-step PPO.
            \citet{jiao2024planning-trajectories-collection} follow the same way but use DPO training with sampled reasoning trajectories as preference data.
            \textit{PQM} \cite{li2024pqm} reformulates the PRM training as a Q-value ranking problem.
            It trains a PRM through a margin-based ranking loss over step pairs, encouraging higher Q-values for steps leading to correct answers.
            \textit{OmegaPRM} \cite{luo2024omegareward} adopts an efficient and fully automated method to collect step-level supervision with a novel MCTS algorithm.
            This significantly reduces the annotation cost and enables to scale up the training of high-quality process reward models.
            \textit{HRM} \cite{wang2025hierarchical-rm} also leverages reasoning trajectories from MCTS for training.
            It evaluates both individual steps and consecutive step pairs to capture multi-step coherence and error recovery.
            Thus this avoids penalizing an early error regardless of subsequent correction.
            \textit{SCOPE} \cite{xu2025scope} compresses math reasoning steps into syntax trees, which avoids large-scale sampling and thus reduces annotation cost.

            Some methods attempt to derive process rewards from outcome rewards.
            \citet{yuan2024free-process-rewards} derive process rewards from an outcome reward model as the cumulative token-level log-ratio differences.
            Building on this idea,
            \textit{PRIME} \cite{cui2025prime-process-implicit-reward} trains a reward model online using only outcome-level supervision (\eg answer correctness) on sampled solution rollouts for mathematical problems.
            The reward model estimates token-level process rewards without requiring step-level annotations.
            It then fine-tunes the language model with these rewards through PPO (or REINFORCE variants).
            \textit{OREAL} \cite{lyu2025oreal} samples reasoning trajectories and their binary outcome rewards from a fine-tuned policy model.
            Afterward, it trains a PRM to assign importance scores to each token in the trajectories, where the weighted sum of the scores needs to approximate the outcome rewards.
    
\begin{figure*}[!ht]
    \centering
    \includegraphics[width=\linewidth]{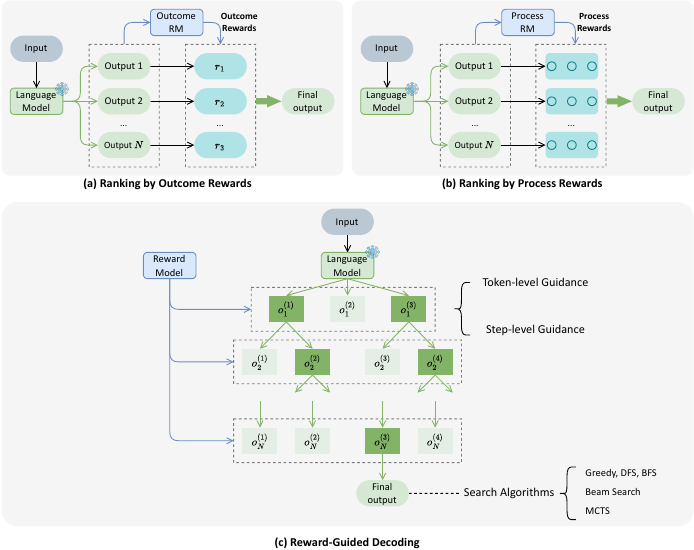}
    \caption{
        Illustrations of strategies for \textbf{Inference with Rewards}.
        (a,b): Generate-then-rank with outcome and process rewards.
        (c): Reward-guided decoding at the token and step level with search algorithms.
    }
    \label{fig_inference_rewards}
\end{figure*}

            Due to the popularity of generative reward models, generative PRMs have also been introduced.
            For instance, \textit{GenPRM} \cite{zhao2025genprm}
            predicts the correctness of a reasoning step by generating CoT reasoning and code verification.
            \textit{GenPRM} is trained via SFT on synthesized reasoning trajectories from the MATH dataset, where external LLMs generate the rationales and correctness labels.
            \textit{R-PRM} \cite{she2025reasoning-prm} and \textit{ThinkPRM} \cite{khalifa2025thinkprm} adopt a comparable generative PRM design by incorporating the reasoning process.

            In the multimodal field,
            \textit{M-STAR} \cite{liu2024diving-multimodal} trains a multimodal PRM on estimated step-level scores via Monte Carlo rollouts
            and then uses the reward model to sample high-quality responses to supervise the subsequent fine-tuning iterations.

\section{Inference with Rewards} \label{sec_inference_rewards}
    After the training stage, inference with rewards offers a flexible and lightweight mechanism to adapt and steer the model behavior without modifying model parameters.
    We identify two primary inference-with-rewards strategies:
    (i) \textit{Generate-then-Rank} and (ii) \textit{Reward-Guided Decoding}.    
    These strategies play a critical role for achieving \textbf{\textit{test-time scaling}}:
    They allow the language model to search, reflect, and revise its outputs on the fly.

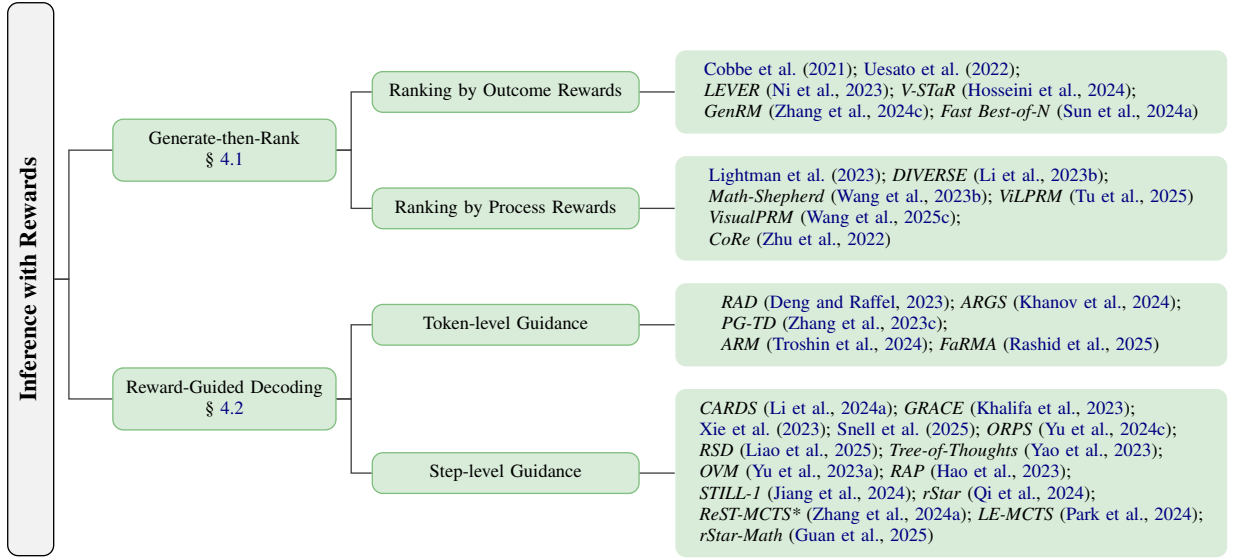
\begin{figure*}[!t]
    \centering
    \begin{forest}
        forked edges,
        for tree={
            grow=east,
            reversed=true,
            anchor=base west,
            parent anchor=east,
            child anchor=west,
            base=center,
            font=\small,
            rectangle,
            draw=black,
            rounded corners,
            align=center,
            text centered,
            minimum width=4em,
            edge+={darkgray, line width=0.5pt},
            s sep=8pt,
            line width=0.5pt,
            ver/.style={rotate=90, child anchor=north, parent anchor=south, anchor=center, minimum width=19em, fill=gray!10},
            leaf/.style={font=\scriptsize, align=left, draw=none, inner xsep=8pt}, %
            leaf2/.style={text width=10em, font=\scriptsize, draw=none, inner xsep=8pt} %
        },
        where level=1{text width=7em,align=center,font=\scriptsize}{},
        where level=2{text width=8.5em, align=center, font=\scriptsize}{},
        where level=3{text width=17.5em, align=left,font=\scriptsize}{},
        where level=4{align=left,font=\scriptsize}{},
        [
            \textbf{Inference with Rewards}, ver
            [
                Generate-then-Rank \\ \Cref{sec_generate-then-rank}, fill=inference-fill, draw=inference-border
                [
                    Ranking by Outcome Rewards, fill=inference-fill, draw=inference-border
                    [
                        \citet{cobbe2021training-verifiers};
                        \citet{uesato2022solving};\\
                        \textit{LEVER} \cite{ni2023lever};
                        \textit{V-STaR} \cite{hosseini2024vstar}; \\
                        \textit{GenRM} \cite{zhang2024genrm};
                        \textit{Fast Best-of-N} \cite{sun2024fast-best-of-n}
                        ,leaf, fill=inference-fill, draw=none
                    ]
                ]
                [
                    Ranking by Process Rewards, fill=inference-fill, draw=inference-border
                    [
                        \citet{lightman2023letsverify};
                        \textit{DIVERSE} \cite{li2023diverse-stepaware-verifier};\\
                        \textit{Math-Shepherd} \cite{wang2023math-shepherd}; 
                        \textit{ViLPRM} \cite{tu2025vilbench}\\
                        \textit{VisualPRM} \cite{wang2025visualprm};\\
                        \textit{CoRe} \cite{zhu2022cooperative}
                        , leaf, fill=inference-fill, draw=none
                    ]
                ]
            ]
            [
                Reward-Guided Decoding \\ \Cref{sec_reward_guided_decoding}, fill=inference-fill, draw=inference-border
                [
                    Token-level Guidance, fill=inference-fill, draw=inference-border
                    [
                        \textit{RAD} \cite{deng2023rad};
                        \textit{ARGS} \cite{khanov2024args};\\
                        \textit{PG-TD} \cite{zhang2023pg-td-planning-code-geneartion};\\
                        \textit{ARM} \cite{troshin2024lowrankrad};
                        \textit{FaRMA} \cite{rashid2025farma-costeffective-reward-guided-generation}
                        , leaf, fill=inference-fill, draw=none
                    ]
                ]
                [
                    Step-level Guidance, fill=inference-fill, draw=inference-border
                    [
                        \textit{CARDS} \cite{li2024cards};
                        \textit{GRACE} \cite{khalifa2023grace};\\
                        \citet{xie2023self};
                        \citet{snell2025scaling};
                        \textit{ORPS} \cite{yu2024orps}; \\
                        \textit{RSD} \cite{liao2025rewardspeculative};
                        \textit{Tree-of-Thoughts} \cite{yao2023treeofthoughts};\\
                        \textit{OVM} \cite{yu2023ovm};
                        \textit{RAP} \cite{hao2023rap};\\
                        \textit{STILL-1} \cite{jiang2024slow-thinking-still-1};
                        \textit{rStar} \cite{qi2024rstar};\\
                        \textit{ReST-MCTS*} \cite{zhang2024rest-mcts*};
                        \textit{LE-MCTS} \cite{park2024ensembling-le-mcts};\\
                        \textit{rStar-Math} \cite{guan2025rstar-math}
                        , leaf, fill=inference-fill, draw=none
                    ]
                ]
            ]
        ]
    \end{forest}
    \caption{
        Overview of \textbf{Inference with Rewards}.
    }
    \label{forest_inference_rewards}
\end{figure*}

    \subsection{Generate-then-Rank} \label{sec_generate-then-rank}
        The generate-then-rank approach, usually referred to as \textit{Best-of-N}, easily scales test-time compute to improve model outputs.
        It samples a number of candidate responses from the language model, scores them with a reward model, and selects the best one as the final output by ranking or voting \cite{wang2022self-consistency}.
        Based on the reward granularity, we distinguish two kinds of methods: (i) \textit{ranking by outcome rewards} and (ii) \textit{ranking by process rewards} as shown in \refSubfigure{fig_inference_rewards}{a,b}.
 
        \paragraph{Ranking by Outcome Rewards.}
            As shown in \refSubfigure{fig_inference_rewards}{a},
            this method adopts an outcome reward model (ORM) to assess the holistic quality of candidate responses.
            Early work by \citet{cobbe2021training-verifiers} trains a binary ORM to evaluate the correctness of candidate math solutions
            and selects the top-ranked one as the final output.
            \citet{uesato2022solving} adopt the same idea and conduct comprehensive experiments on ranking outputs by ORMs.
            \textit{LEVER} \cite{ni2023lever} trains a binary classifier as the ORM with code execution results as supervision.
            During inference, it ranks generated candidate code jointly based on the ORM's score and the generation probability.
            \textit{V-STaR} \cite{hosseini2024vstar} trains a verifier as the ORM on preference pairs through DPO to rank candidate math/code solutions during inference.
            Its ORM supports iterative training on dynamically collected preference data, which can progressively improve performance.
            \textit{GenRM} \cite{zhang2024genrm} follows a generative way.
            It scores multiple candidate solutions by computing the generation probability of the \textit{Yes/No} token as rewards
            and then selects the best solution.

            While simple and effective, this strategy becomes computationally expensive as the number of candidates increases.
            To address this, \textit{Fast Best-of-N} \cite{sun2024fast-best-of-n} follows a speculative rejection scheme.
            It queries the reward model multiple times throughout the generation of one candidate response and terminates the unpromising generation early based on the rewards.
            As such, this way accelerates the inference process without completely generating all candidates.
            \citet{jinnai2024regularized-best-of-n} investigate the reward backing problem of the generate-then-rank strategy.
            \citet{brown2024language-monkeys} systematically investigate the scaling of testing-time computing with various ORMs across multiple tasks.

        \paragraph{Ranking by Process Rewards.}
            As aforementioned, outcome reward models may struggle to discern the nuance among candidate responses.
            Thus many methods adopt process reward models (PRMs) for the generate-then-rank strategy.
            These methods score intermediate steps of candidate responses through a PRM
            and aggregate these step-level scores through multiplication or minimum to compute an overall score for ranking or voting \cite{zhang2025lessons-prm}.

            Early work by \citet{lightman2023letsverify} introduces a PRM trained on a large-scale human-labeled math dataset (PRM800K)
            and ranks candidate math solutions by the product of their step-level reward scores.
            \textit{DIVERSE} \cite{li2023diverse-stepaware-verifier} trains a reward model to assign scores to both the entire path and individual reasoning steps. Then it picks up the best one from multiple candidates through weighted voting.
            \textit{Math-Shepherd} \cite{wang2023math-shepherd} uses a PRM to score each step in a math solution and ranks solutions according to the lowest step-level score.
            Notably these methods can improve reasoning consistency---the chosen solution builds on a series of reliable steps rather than merely delivering a correct final answer.

            Some studies have explored multimodal PRMs for the generate-then-rank strategy.
            Examples include \textit{VisualPRM} \cite{wang2025visualprm} and \textit{ViLPRM} \cite{tu2025vilbench}.
            They incorporate both image and text inputs with step-wise evaluation in the multimodal reasoning.
            Some approaches combine outcome and process rewards to improve ranking quality.
            For example, \textit{CoRe} \cite{zhu2022cooperative} integrates these two signals to jointly verify model outputs during inference.

    \subsection{Reward-Guided Decoding} \label{sec_reward_guided_decoding}
\begin{figure*}[!ht]
    \centering
    \includegraphics[width=0.9\linewidth]{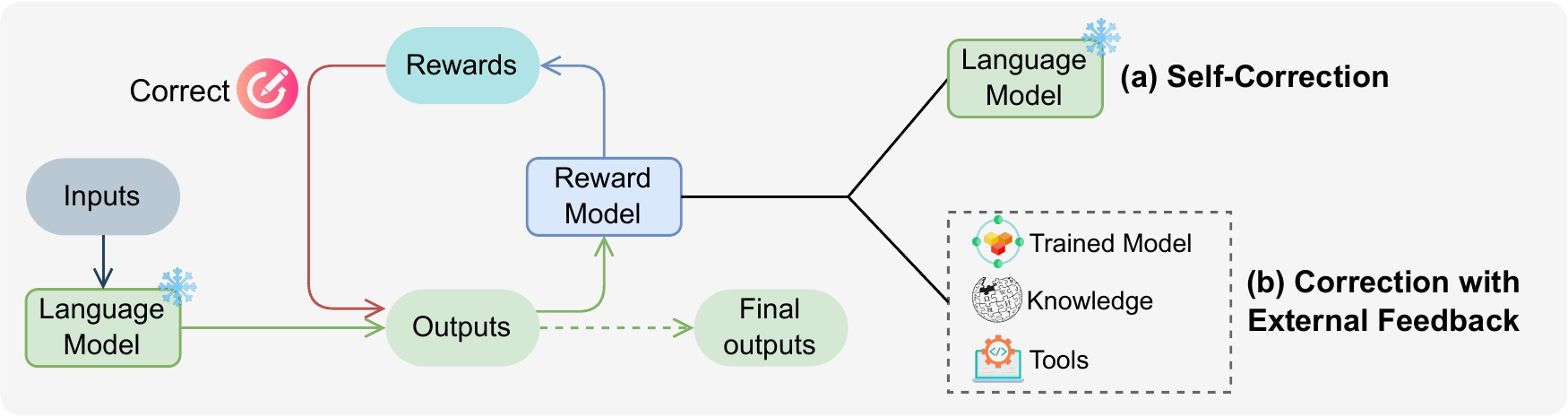}
    \caption{
        Illustration of \textbf{Post-Inference with Rewards}.
        (a): Self-Correction, using the language model itself.
        (b): Correction with External Feedback, such as trained model, external knowledge, and external tools.
    }
    \label{fig_post-inference_rewards}
\end{figure*}

        While the above generate-then-rank approach is simple and effective,
        it inherently decouples generation from evaluation, limiting its ability to refine outputs dynamically during decoding.
        In contrast, \textit{reward-guided decoding} tightly incorporates reward signals to guide the generation process of language models.
        Based on the granularity of guidance, we categorize this line of work into two strategies:
        \textit{token-level guidance} and \textit{step-level guidance}.
        As shown in \refSubfigure{fig_post-inference_rewards}{c},
        these strategies guide the language model's token-level or step-level decoding based on the reward signals through a search algorithm, such as greedy search, beam search, or Monte Carlo Tree Search (MCTS).
        This enables fine-grained control over output quality and alignment and can foster reasoning and planning abilities.

        \paragraph{Token-level Guidance.}
            Token-level guidance steers language model generation by incorporating reward signals into the token selection process at each decoding step.
            This strategy commonly combines the token likelihoods with the reward signals from a reward model.

            \textit{RAD} \cite{deng2023rad} adjusts token selection by combining the token's likelihood and the scalar rewards.
            It can control output attributes such as non-toxicity and sentiment.
            \textit{ARGS} \cite{khanov2024args} applies a similar strategy to align the helpfulness and harmless preferences.
            This work converts preference alignment from the previous training stage to the inference stage.
            \textit{PG-TD} \cite{zhang2023pg-td-planning-code-geneartion} targets code generation.
            It uses the pass rate over test cases as reward signals and integrates them into MCTS to guide token-level planning and generation
            and integrates it into MCTS to guide token-level planning and generation.

            However, the above methods are limited by low decoding efficiency since they require querying the reward model for each candidate token at every decoding step.
            To improve decoding efficiency, \textit{ARM} \cite{troshin2024lowrankrad} proposes a low-rank approximation of the reward model to score all token candidates within a single query.
            Similarly, \textit{FaRMA} \cite{rashid2025farma-costeffective-reward-guided-generation} trains a reward model that scores all token candidates in a single forward pass.

        \paragraph{Step-level Guidance.}
            Beyond token-level guidance, step-level guidance operates on intermediate steps of the generation process.
            As illustrated in \refSubfigure{fig_inference_rewards}{d}, the generation is decomposed into multiple intermediate steps.
            At each step, a search algorithm, such as beam search and Monte Carlo Tree Search (MCTS), explores the output space and selects the appropriate steps with the guidance of reward signals.
            This mechanism enables the model to recover from earlier errors and enhance reasoning.

            In some cases, a step may correspond to a semantic segment.
            For example, \textit{CARDS} \cite{li2024cards} samples candidate semantic segments at each step.
            Then it uses a reward model to score the resulting output with a segment
            and choose the high-reward segment for continuing decoding.

            More studies focus on guiding reasoning steps with reward signals.
            \textit{GRACE} \cite{khalifa2023grace} trains a discriminator as the reward model that scores the correctness of candidate reasoning steps.
            During decoding, it combines the step-level reward scores with the language model's likelihood to guide step selection toward more accurate reasoning paths.
            \citet{xie2023self} employs a similar approach by prompting the language model itself to evaluate the reasoning steps during beam search.
            \citet{snell2025scaling} use a process reward model to guide which reasoning steps are retained in the beam search and look-ahead search during decoding.
            \textit{ORPS} \cite{yu2024orps} derives outcome rewards from code execution feedback and process rewards from LLM-generated self-critiques about code reasoning quality.
            It combines them to guide a tree-based search process, \eg selecting, expanding, and refining candidate solutions throughout the generation.
            \textit{RSD} \cite{liao2025rewardspeculative} combines rewards with speculative decoding.
            During its speculative decoding, it leverages a process reward model to determine whether to accept a draft model's output or invoke the target model.

            Some studies guide the decoding based on the step-level value, \ie cumulative future rewards.
            \textit{Tree-of-Thoughts} \cite{yao2023treeofthoughts} prompts the language model to assess the value of the current state by producing a scalar score or short phrases.
            It then guides the search algorithms, like BFS and DFS, to explore diverse reasoning trajectories.
            \textit{OVM} \cite{yu2023ovm} trains a value model to estimate the probability that a partial reasoning path leads to a correct final answer.
            During inference, it guides the decoding via value-based beam search to select the most promising reasoning trajectories.

            Other methods use reward signals to guide Monte Carlo Tree Search (MCTS).
            \textit{RAP} \cite{hao2023rap} defines task-specific reward functions for MCTS.
            This encourages the model to simulate future states and select high-reward reasoning paths for planning, math, and logic tasks.
            \textit{STILL-1} \cite{jiang2024slow-thinking-still-1} follows a similar reward-guided MCTS way.
            \textit{rStar} \cite{qi2024rstar} assigns rewards to reasoning trajectories by combining the correctness of the final answer with self-consistency confidence.
            Then it back-propagates these rewards to the steps in the trajectories to guide future MCTS exploration toward more promising reasoning paths.
            Several extensions leverage process reward models to precisely guide MCTS,
            including \textit{ReST-MCTS*} \cite{zhang2024rest-mcts*}, \textit{LE-MCTS} \cite{park2024ensembling-le-mcts}, and \textit{rStar-Math} \cite{guan2025rstar-math}.
            They use step-level scoring from process reward models to select and expand high-quality reasoning trajectories during the search.

\begin{figure*}[!t]
    \centering
    \begin{forest}
        forked edges,
        for tree={
            grow=east,
            reversed=true,
            anchor=base west,
            parent anchor=east,
            child anchor=west,
            base=center,
            font=\small,
            rectangle,
            draw=black,
            rounded corners,
            align=center,
            text centered,
            minimum width=4em,
            edge+={darkgray, line width=0.5pt},
            s sep=8pt,
            line width=0.5pt,
            ver/.style={rotate=90, child anchor=north, parent anchor=south, anchor=center, minimum width=15em, fill=gray!10},
            leaf/.style={font=\scriptsize, align=left, draw=none, inner xsep=8pt}, %
            leaf2/.style={text width=10em, font=\scriptsize, draw=none, inner xsep=8pt} %
        },
        where level=1{text width=6em,align=center,font=\scriptsize}{},
        where level=2{text width=8.5em, align=center, font=\scriptsize}{},
        where level=3{text width=17.5em, align=left,font=\scriptsize}{},
        where level=4{align=left,font=\scriptsize}{},
        [
            \textbf{Post-Inference with Rewards}, ver
            [
                Self-Correction \\ \Cref{sec_self-correction}, fill=post-inference-fill, draw=post-inference-border
                [
                    \textit{Self-Refine} \cite{madaan2023self-refine};
                    \textit{Reflexion} \cite{shinn2023reflexion};\\
                    \textit{CoVe} \cite{dhuliawala2023chain-of-verification};
                    \textit{SCoRE} \cite{kumar2024score-self-correct};\\
                    \textit{RISE} \cite{qu2024introspection-self-improve}
                    ,leaf, fill=post-inference-fill, draw=none, text width=17.5em
                ]
            ]
            [
                Correction with \\External Feedback \\ \Cref{sec_correction_with_external_feedback}, fill=post-inference-fill, draw=post-inference-border
                [
                    Trained Models, fill=post-inference-fill, draw=post-inference-border
                    [
                        \textit{CodeRL} \cite{le2022coderl};
                        \textit{RL4F} \cite{akyurek2023rl4f}; \\
                        \textit{Shepherd} \cite{wang2023shepherd};
                        \textit{A2R} \cite{lee2024ask};\\
                        \textit{CTRL} \cite{xie2025teaching-ctrl};
                        \textit{CriticGPT} \cite{mcaleese2024criticgpt-catch-bugs};\\
                        \textit{DARS} \cite{li2025dars-verbal-reflection};
                        \textit{REFINER} \cite{paul2023refiner};\\
                        \textit{AutoMathCritique} \cite{xi2024automathcritique};
                        \textit{MAD} \cite{liang2023divergent-thinking-multiagent-debate};\\
                        \citet{cohen2023lm-vs-lm};
                        \citet{du2023factulaity-multiagent-debate}
                        ,leaf, fill=post-inference-fill, draw=none
                    ]
                ]
                [
                    External Knowledge, fill=post-inference-fill, draw=post-inference-border
                    [
                        \textit{RARR} \cite{gao2022rarr};
                        \textit{ReFeed} \cite{yu2023refeed};\\
                        \textit{LLM-Augmenter} \cite{peng2023llmaugmenter};
                        \citet{varshney2023stitch};\\
                        \textit{FACTOOL} \cite{chern2023factool}
                        ,leaf, fill=post-inference-fill, draw=none
                    ]
                ]
                [
                    External Tools, fill=post-inference-fill, draw=post-inference-border
                    [
                        \textit{Self-Edit} \cite{zhang2023self-edit};
                        \textit{Self-Debug} \cite{chen2023self-debug};\\
                        \textit{CYCLE}  \cite{ding2024cycle};
                        \textit{Logic-LM} \cite{pan2023logic-lm}; \\
                        \textit{IHR} \cite{qiu2023ihr-symbolic-interpreter};
                        \textit{Baldur} \cite{first2023baldur-proof-geneartion};\\
                        \textit{CRITIC} \cite{gou2023critic-tool-critiquing};
                        \textit{RCI} \cite{kim2023rci-computer-tasks}
                        ,leaf, fill=post-inference-fill, draw=none
                    ]
                ]
            ]
        ]
    \end{forest}
    \caption{
        Overview of \textbf{Post-Inference with Rewards}.
    }
    \label{forest_post-inference_rewards}
\end{figure*}
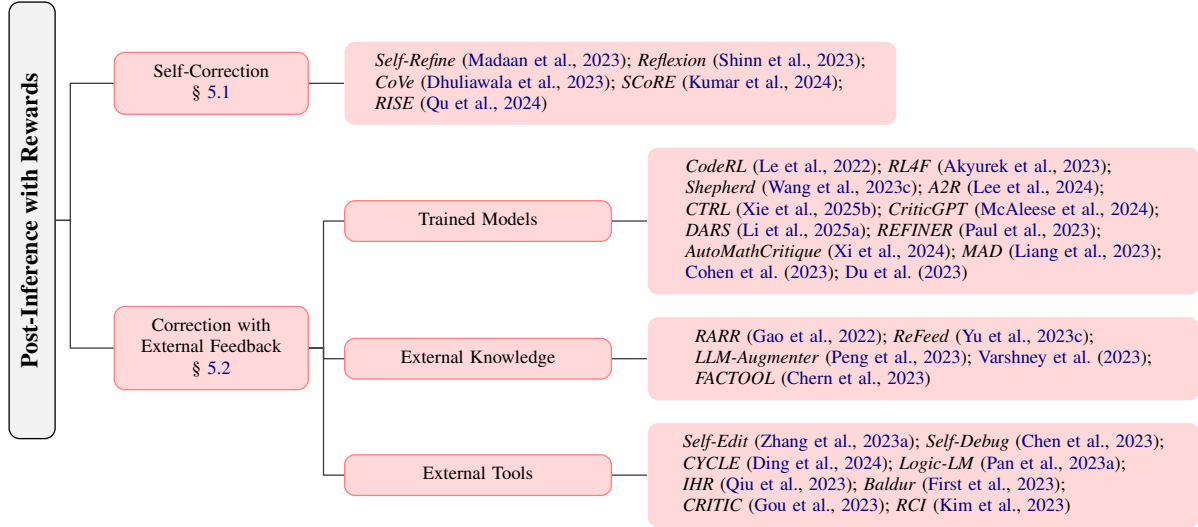

\section{Post-Inference with Rewards} \label{sec_post-inference_rewards}
    Post-inference with rewards aims to correct and refine the model outputs after they have been generated with reward signals.
    This approach enables iterative enhancement without updating model parameters,
    offering a lightweight and compatible mechanism for \textit{test-time scaling}.
    Compared to sparse scalar rewards, post-inference methods favor critique rewards that provide fine-grained and dense signals for correction.
    These critiques typically identify error locations, explain reasoning flaws, and suggest revisions.
    To leverage such rich feedback, the language model commonly incorporates them as augmented context and iteratively generates revised outputs.
    According to the source of rewards, we categorize these methods into two main strategies:
    \textit{Self-Correction} and \textit{Correction with external rewards}.

    \subsection{Self-Correction} \label{sec_self-correction}
        As illustrated in \refSubfigure{fig_post-inference_rewards}{a},
        self-correction leverages the language model itself as a generative reward model to evaluate and revise its own outputs, similar to the aforementioned self-rewarding strategy.
        \textit{Self-Refine} \cite{madaan2023self-refine} prompts the language model itself to produce natural language feedback on its own outputs.
        It then leverages this feedback as reward signals to refine outputs iteratively.
        Similarly, \textit{Reflexion} \cite{shinn2023reflexion} generates reflection feedback through the language model itself.
        It additionally maintains a memory bank to store previous feedback, outputs, and scalar feedback from evaluation metrics.
        These reflections serve as auxiliary contexts to refine subsequent generations.
        \textit{CoVe} \cite{dhuliawala2023chain-of-verification} prompts the language model to generate and answer verification questions about its own outputs to identify factual errors.
        This verification feedback guides the model to correct its responses to reduce hallucination.

        In addition, \textit{SCoRE} \cite{kumar2024score-self-correct} trains the language model via reinforcement learning to enhance its self-correction capability.
        Similarly, \textit{RISE} \cite{qu2024introspection-self-improve} bootstraps the language model with DPO training for better self-correction.

    \subsection{Correction with External Feedback} \label{sec_correction_with_external_feedback}
        While self-correction is simple, prior studies argue that general language models struggle to identify and correct their errors without external feedback, especially for less capable small-scale models \cite{huang2023cannot-self-correct,kamoi2024can-correct,madaan2023self-refine,pan2023automatically}.
        Owing to this, increasing attention has been devoted to incorporating external feedback as reward signals to refine model outputs as shown in \refSubfigure{fig_post-inference_rewards}{b}.
        We classify these works according to the feedback source: \textit{Trained Models}, \textit{External Knowledge}, and \textit{External Tools}.

        \paragraph{Trained Model.}
            Many methods rely on more capable trained models (commonly referred to as critic models) to provide feedback as reward signals.
            The feedback is mostly natural language critiques containing quality assessments and correction suggestions on model outputs.
            The early work \textit{CodeRL} \cite{le2022coderl} uses a trained critic model to predict the functional correctness of the generated code.
            Afterward, this feedback serves as reward signals to guide the language model to regenerate the code.
            \citet{welleck2022generating-self-correct} trains a critic model to provide both scalar and critique feedback for mathematics, constrained generation, and toxicity control.
            \textit{RL4F} \cite{akyurek2023rl4f} uses a critic model to generate critiques for topic-based summarization, action planning, and alphabetization tasks.
            \textit{Shepherd} \cite{wang2023shepherd} presents a critic model that can identify factual, logical, or stylistic issues and suggest actionable refinements for the language model.
            \textit{A2R} \cite{lee2024ask} incorporates factual critiques from a critic model to mitigate the hallucination issue.
            \textit{CTRL} \cite{xie2025teaching-ctrl} focuses on code generation. It fine-tunes a critic model with code execution results via GRPO, and the model provides actionable critiques to guide the refinement of generated code.
            \citet{mcaleese2024criticgpt-catch-bugs} use a trained critic model, \textit{CriticGPT}, to identify the flaws in generated code.
            It can output structured critiques that disclose bugs and reasoning errors in the generated code.
            \textit{DARS} \cite{li2025dars-verbal-reflection} refines model outputs by guiding iterative reasoning with the reflection critiques from a trained critic model.

            Moreover, some studies focus on fine-grained step-level feedback for correction.
            \textit{REFINER} \cite{paul2023refiner} adopts a critic model to provide fine-grained feedback on the intermediate reasoning steps.
            It then uses these feedback to iteratively refine the language model's reasoning for math and moral story.
            \textit{AutoMathCritique} \cite{xi2024automathcritique} similarly trains a critic model to provide process-level critiques,
            which supports iterative refinement of mathematical reasoning outputs.

            Some methods follow the multi-agent debate design, where critiques from peer agents support reflection and improvement.
            These methods include \textit{MAD} \cite{liang2023divergent-thinking-multiagent-debate}, \citet{cohen2023lm-vs-lm}, and \citet{du2023factulaity-multiagent-debate}.

            In the multimodal field, \textit{DRESS} \cite{chen2024dress} leverages GPT-4 to generate feedback on vision-language model outputs, including critiques and refinement suggestions. It then uses these feedback to guide the model to refine its outputs iteratively.

        \paragraph{External Knowledge.}
            External knowledge sources mainly provide factual critiques along with retrieved evidence,
            which can improve factuality and reduce hallucinations.
            \textit{RARR} \cite{gao2022rarr} derives hybrid rewards based on the entailment-based agreement between the model output and retrieved evidence from external knowledge.
            These reward signals then guide the post-hoc correction of the language model to improve factual attribution while preserving the original text's intent and structure.
            \textit{ReFeed} \cite{yu2023refeed} applies a similar method to knowledge-intensive QA tasks.
            \textit{LLM-Augmenter} \cite{peng2023llmaugmenter} computes the Knowledge F1 scores between model outputs and retrieved evidence from external knowledge as reward signals.
            These signals guide the language model's decision to either finalize or continue to refine its response.
            \citet{varshney2023stitch} formulates verification questions regarding the low-confidence concepts in the model outputs.
            To verify these questions, it retrieves evidence from external knowledge and takes the results and evidence as feedback to guide the language model's refinement and reduce hallucination.
            As a general factuality tool, \textit{FACTOOL} \cite{chern2023factool} broadens this idea to an enormous scope, including knowledge-based QA, code generation, mathematical reasoning, and scientific literature review.

        \paragraph{External Tools.}
            External tools can execute and verify the language model outputs, and their feedback can work as reward signals for correction.
            \textit{Self-Edit} \cite{zhang2023self-edit} and \textit{Self-Evolve} \cite{jiang2023self-evolve} use program execution feedback from code compilers to guide the refinement of the language model.
            \textit{Self-Debug} \cite{chen2023self-debug} and \textit{CYCLE}  \cite{ding2024cycle} extend them with more feedback, for instance, unit test results and program explanations.

            Apart from code compilers,
            \textit{Logic-LM} \cite{pan2023logic-lm} uses the feedback from symbolic logic reasoner as critique rewards to refine the model's answers to logic reasoning problems.
            \textit{IHR} \cite{qiu2023ihr-symbolic-interpreter} depends on the feedback from a symbolic interpreter for inductive reasoning, and \textit{Baldur} \cite{first2023baldur-proof-geneartion} incorporates feedback from a proof checker for automated formal verification.
            \textit{CRITIC} \cite{gou2023critic-tool-critiquing} and \textit{RCI} \cite{kim2023rci-computer-tasks} can leverage feedback from diverse external tools, such as search engines, code interpreters, and calculators.

\begin{figure*}[!t]
    \centering
    \includegraphics[width=0.8\linewidth]{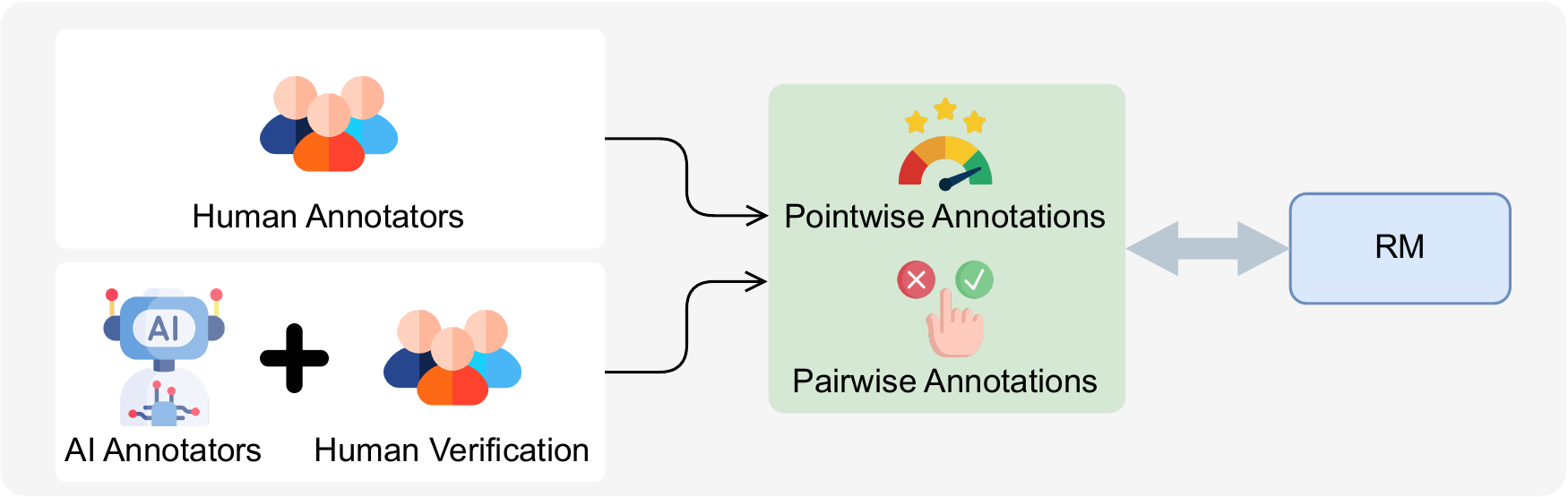}
    \caption{
        Illustration of \textbf{Benchmarking Reward Models}.
        Annotations come from human annotators or AI annotators with human verification.
        The annotations are mainly pointwise (\eg a scalar score for each sample) or pairwise (\eg chosen and rejected responses).
    }
    \label{fig_benchmarking_reward_models}
\end{figure*}

\begin{figure*}[!t]
    \centering
    \begin{forest}
        forked edges,
        for tree={
            grow=east,
            reversed=true,
            anchor=base west,
            parent anchor=east,
            child anchor=west,
            base=center,
            font=\small,
            rectangle,
            draw=black,
            rounded corners,
            align=center,
            text centered,
            minimum width=4em,
            edge+={darkgray, line width=0.5pt},
            s sep=8pt,
            line width=0.5pt,
            ver/.style={rotate=90, child anchor=north, parent anchor=south, anchor=center, minimum width=18em, fill=gray!10},
            leaf/.style={font=\scriptsize, align=left, draw=none, inner xsep=8pt}, %
            leaf2/.style={text width=10em, font=\scriptsize, draw=none, inner xsep=8pt} %
        },
        where level=1{text width=10em,align=center,font=\scriptsize}{},
        where level=2{text width=21.5em, align=center, font=\scriptsize}{},
        where level=3{text width=17.5em, align=left,font=\scriptsize}{},
        where level=4{align=left,font=\scriptsize}{},
        [
            \textbf{Benchmarking Reward Models}, ver
            [
                Benchmarking \\Outcome Reward Models \\ \Cref{sec_benchmark_outcome_reward_models}, fill=benchmark-fill, draw=benchmark-border
                [
                    \citet{zheng2023judging-llm-as-a-judge};
                    \textit{RewardBench} \cite{lambert2024rewardbench}; \\
                    \textit{RM-Bench} \cite{liu2024rmbench};
                    \textit{AceMath-RewardBench} \cite{liu2024acemath};\\
                    \textit{RMB} \cite{zhou2024rmb};
                    \textit{CriticBench} \cite{lin2024criticbench};\\
                    \textit{MetaCritique} \cite{sun2024metacritique}
                    ,leaf, fill=benchmark-fill, draw=none
                ]
            ]
            [
                Benchmarking \\Process Reward Models \\ \Cref{sec_benchmark_process_reward_models}, fill=benchmark-fill, draw=benchmark-border
                [
                    \textit{MathCheck-GSM} \cite{zhou2024mathcheck-gsm};
                    \textit{MR-GSM8K} \cite{zeng2023mr-gsm8k};\\
                    \textit{ProcessBench} \cite{zheng2024processbench};
                    \textit{PRMBench} \cite{song2025prmbench};\\
                    \textit{Big-Bench Mistake} \cite{tyen2023bigbench-mistake};
                    \textit{MR-Ben} \cite{zeng2024mr-ben}
                    ,leaf, fill=benchmark-fill, draw=none
                ]
            ]
            [
                Benchmarking \\Multimodal Reward Models \\ \Cref{sec_benchmark_multimodal_reward_models}, fill=benchmark-fill, draw=benchmark-border
                [
                    \textit{MJ-Bench} \cite{chen2024mj-bench};
                    \textit{MLLM-as-a-Judge} \cite{chen2024mllm-as-a-judge};\\
                     \textit{VL-RewardBench} \cite{li2024vlrewardbench}; \\
                    \textit{Multimodal-RewardBench} \cite{yasunaga2025multimodal-rewardbench} ;\\
                    \textit{SVIP} \cite{gao2025benchmarking-svip};
                    \textit{VLRMBench} \cite{ruan2025vlrmbench}
                    ,leaf, fill=benchmark-fill, draw=none
                ]
            ]
            [
                Other Benchmarks \\ \Cref{sec_other_benchmarks}, fill=benchmark-fill, draw=benchmark-border
                [
                    \textit{RAG-RewardBench} \cite{jin2024ragrewardbench};
                    \textit{M-RewardBench} \cite{gureja2024multilingual-rewardbench};\\
                    \textit{PPE} \cite{frick2024evaluate-ppe}
                    ,leaf, fill=benchmark-fill, draw=none
                ]
            ]
        ]
    \end{forest}
    \caption{
        Overviews of \textbf{Benchmarking Reward Models}.
    }
    \label{forest_benchmark_rewards}
\end{figure*}
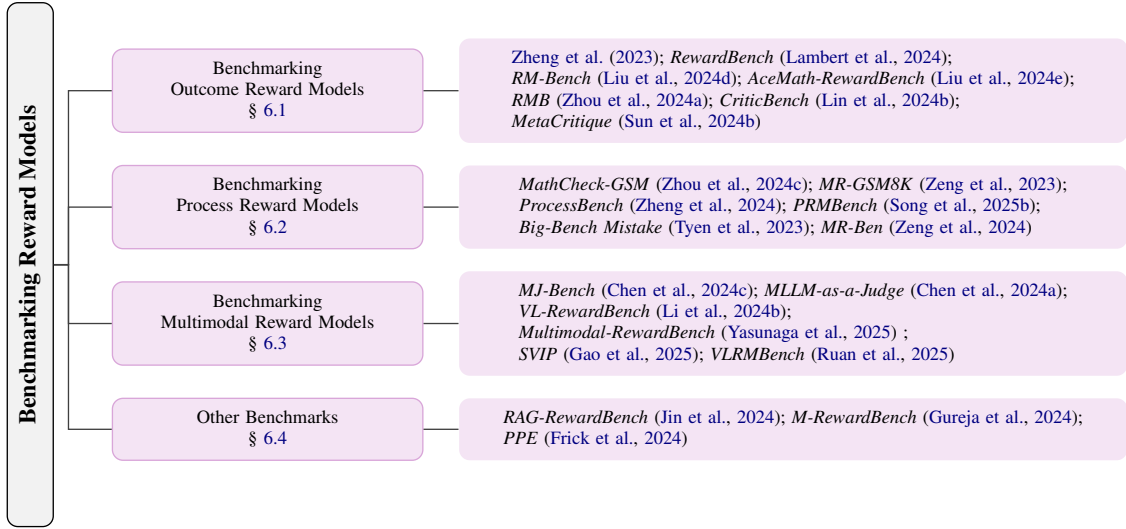

\section{Benchmarking Reward Models} \label{sec_benchmarks_rm}
    As reward models play a central role in the learning-from-rewards paradigm for post-training and test-time scaling,
    rigorous and diverse benchmarks are essential for evaluating their capabilities.
    As illustrated in \Cref{fig_benchmarking_reward_models},
    recent benchmarking efforts primarily rely on expert human annotators or AI annotators (\eg LLM-as-a-Judge frameworks) followed by human verification to ensure reliability.
    The resulting annotations are mainly pointwise (\eg scalar scoring) or pairwise (\eg selecting the preferred response given two options).
    These benchmarks vary in task coverage, evaluation protocols, annotation sources, and reward formats.
    We discuss the representative benchmarks in the literature as follows.

    \subsection{Benchmarking Outcome Reward Models} \label{sec_benchmark_outcome_reward_models}
        A dominant line of benchmarking studies centers on outcome reward models that evaluate the overall quality of generated outputs.
        \citet{zheng2023judging-llm-as-a-judge} is an early work that evaluates LLMs' judging ability by directly prompting them.
        As LLMs can naturally function as generative reward models, this study also represents one of the earliest benchmarks for reward models.
        \citet{zhou2025evaluating-judges-as-evaluators} follow a similar idea and focus on evaluation for test-time scaling.

        \textit{RewardBench} \cite{lambert2024rewardbench} is the first comprehensive benchmarks for reward models.
        It aggregates preference data from existing datasets, such as AlpacaEval and MTBench, to evaluate reward model performance in chatting, reasoning, and safety.
        \textit{RM-Bench} \cite{liu2024rmbench} introduces evaluation for reward models on sensitivity to subtle content changes and robustness to style biases.
        It constructs preference pairs across chat, code, math, and safety domains using GPT-4o.
        \textit{AceMath-RewardBench} \cite{liu2024acemath} focuses on math-specific evaluations. It tests whether reward models can identify correct solutions from candidates across various mathematical tasks and difficulty levels.
        \textit{RMB} \cite{zhou2024rmb} furthermore broadens the evaluation scope to 49 real-world scenarios.

        Apart from evaluating with preference data, some benchmarks focus on the critique ability of reward models.
        \textit{CriticBench} \cite{lin2024criticbench} assess whether reward models can generate critiques that accurately identify the correctness of a response and effectively guide the correction.
        Similarly, \textit{MetaCritique} \cite{sun2024metacritique} benchmarks LLM-generated critiques by decomposing them into atomic information units and assessing their correctness.

    \subsection{Benchmarking Process Reward Models} \label{sec_benchmark_process_reward_models}
        Recently more benchmarks focus on process reward models due to their increasing significance.
        In detail, several benchmarks focus on math reasoning,
        such as \textit{MathCheck-GSM} \cite{zhou2024mathcheck-gsm}, \textit{MR-GSM8K} \cite{zeng2023mr-gsm8k}, and \textit{MR-MATH} \cite{xia2024mr-math}.
        They require reward models to locate the first error step in a math reasoning solution.
        Their testing samples are adapted from existing math datasets, including GSM8K \cite{cobbe2021training-verifiers} and MATH \cite{hendrycks2021math-dataset}.
        Furthermore, \textit{ProcessBench} \cite{zheng2024processbench} features diversity and higher difficulty levels by scaling this up to Olympiad- and competition-level math problems \cite{he2024olympiadbench,gao2024omni-math}.
        Beyond step correctness, \textit{PRMBench} \cite{song2025prmbench} offers a more fine-grained benchmark.
        It annotates each step in the reasoning path with specific error types grouped into three dimensions: simplicity, soundness, and sensitivity.
        The annotations come from LLM-generated perturbations and are subsequently verified by human annotators.

        Besides mathematical reasoning, \textit{Big-Bench Mistake} \cite{tyen2023bigbench-mistake} targets logical reasoning.
        It annotates chain-of-thought trajectories from BIG-Bench \cite{srivastava2023beyond}, each labeled with the first logical error.
        Furthermore, \textit{MR-Ben} \cite{zeng2024mr-ben} expands this to the reasoning process of seven domains: math, logic, physics, chemistry, medicine, biology and code.

    \subsection{Benchmarking Multimodal Reward Models} \label{sec_benchmark_multimodal_reward_models}
        Due to the prevalence of multimodal language models,
        another vital line of benchmarks focuses on multimodal reward models with diverse evaluation protocols.

        \textit{MJ-Bench} \cite{chen2024mj-bench} depends on text-to-image generation tasks for evaluation.
        It builds preference data across four dimensions: text-image alignment, safety, image quality, and social bias.
        \textit{MLLM-as-a-Judge} \cite{chen2024mllm-as-a-judge}
        uses image understanding tasks for benchmarking and includes pointwise and pairwise scoring.
        \textit{VL-RewardBench} \cite{li2024vlrewardbench} includes three tasks: general multimodal instructions, hallucination detection, and multimodal reasoning.
        \textit{Multimodal-RewardBench} \cite{yasunaga2025multimodal-rewardbench} spans six key capabilities of multimodal reward models: general correctness, human preference, factual knowledge, reasoning, safety, and VQA.

        Beyond the outcome level, current benchmarks also assess multimodal process reward models.
        \textit{SVIP} \cite{gao2025benchmarking-svip} targets process-level evaluation on relevance, logic, and attribute correctness of diverse multimodal tasks. 
        It transforms reasoning paths into executable visual programs and automatically annotates each step.
        \textit{VLRMBench} \cite{ruan2025vlrmbench} further integrates evaluation on three dimensions: reasoning steps, whole outcomes, and critiques on error analysis.
        It collects testing data of multimodal understanding through AI annotations and human verification.

    \subsection{Other Benchmarks} \label{sec_other_benchmarks}
        In addition to general-purpose evaluations, several benchmarks aim to address domain-specific or emerging challenges in reward modeling.
        \textit{RAG-RewardBench} \cite{jin2024ragrewardbench} targets reward model evaluation in RAG.
        It constructs preference data for RAG-specific scenarios, including multi-hop reasoning, fine-grained citation, appropriate abstention, and conflict robustness.
        \textit{M-RewardBench} \cite{gureja2024multilingual-rewardbench} extends the evaluation to multilingual contexts.
        Instead of direct evaluation, \textit{PPE} \cite{frick2024evaluate-ppe} indirectly evaluates reward models through RLHF pipelines.
        It measures the performance of trained LLMs with a reward model, offering a practical perspective.

\section{Applications} \label{sec_applications}
    The strategies described above for learning from rewards have been widely adopted across diverse applications.
    Early applications focus on preference alignment, such as RLHF \cite{ouyang2022rlhf} and RLAIF \cite{bai2022constitutional}.
    In particular, the recent DeepSeek-R1 \cite{deepseekai2025deepseek-r1} has demonstrated the effectiveness of reinforcement learning to develop \textit{large reasoning models}, which has inspired a wave of R-1 style applications for diverse areas.
    In this section, we review the primary applications following these strategies.

    \subsection{Preference Alignment}
        Learning-from-rewards strategies have become the cornerstone for aligning LLMs with human preferences.
        These strategies design diverse reward signals to encourage desirable attributes, such as factuality, harmlessness, and helpfulness, while penalizing undesired behaviors like toxicity, bias, and hallucination.
        We summarize three major objectives of preference alignment as follows.
        \begin{itemize}[leftmargin=*,itemsep=0pt]
            \item
                \textbf{Factuality and Reducing hallucination.}
                Hallucination refers to generating fluent but factually incorrect or fabricated content \cite{tian2023factuality}.
                It is a pervasive issue for language models, especially in knowledge-intensive tasks such as healthcare and scientific research.
                The methods for this alignment span the training, inference, and post-inference stages \cite{sun2023fact-rlhf,lin2024flame,zhao2023ha-dpo,peng2023llmaugmenter,wang2023shepherd}.
                The rewards mainly stem from human preferences about factuality as well as external knowledge sources.
                For instance, \textit{Fact-RLHF} \cite{sun2023fact-rlhf} trains a factuality-aware reward model on human preferences and additional supervision from image captions and multiple-choice answers
                The reward model is then used to fine-tune the multimodal language model via PPO, guiding the model to reduce hallucinations.
                \textit{RLFH} \cite{wen2024policy-rlfh} decomposes the model responses into atomic statements, verifies their truthfulness against external knowledge, and converts them into dense token-level scalar rewards.
                To reduce hallucination, it directly uses these reward signals to fine-tune the model via PPO.
            \item
                \textbf{Safety and Harmlessness.}
                Safety and harmlessness constitute another critical axis of alignment, particularly in adversarial or socially sensitive contexts \cite{bai2022constitutional,ji2023beavertails}.
                Language models must be discouraged from producing toxic, offensive, or biased content before being deployed in real-world systems.
                To this end, the methods primarily focus on the training \cite{ouyang2022rlhf,bai2022training} and inference stages \cite{deng2023rad,khanov2024args}.
                For instance, \textit{RAD} \cite{deng2023rad} depends on reward signals to produce non-toxicity content during decoding.
            \item
                \textbf{Helpfulness.}
                Meanwhile, helpfulness emphasizes that language models are expected to provide relevant, informative, and context-aware responses to fulfill user intent \cite{taori2023stanford-alpaca}.
                This alignment is imperative in areas like instruction-following and dialogue systems.
                Reward signals are generally sourced from human preferences and task-specific quality metrics \cite{bai2022training}.
        \end{itemize}

    \subsection{Mathematical Reasoning}
        Mathematical reasoning is vital to measure the language model's ability to solve complex reasoning problems.
        Some methods build reward models and fine-tune the language model for math reasoning \cite{shao2024deepseekmath,deepseekproverv2-2025},
        particularly using process reward models \cite{uesato2022solving,luo2023wizardmath} like \textit{Math-Shepherd} \cite{wang2023math-shepherd}.
        They can provide step-level reward signals for a math reasoning solution.
        Moreover, some approaches construct preference data for math reasoning, \ie correct and incorrect solutions, and then fine-tune the language model through DPO \cite{lai2024stepdpo,xu2025fullstepdpo}.
        Others include inference-time scaling strategies, such as generate-then-rank \cite{cobbe2021training-verifiers,lightman2023letsverify}, and reward-guided decoding with search algorithms like MCTS \cite{hao2023rap,guan2025rstar-math}.

    \subsection{Code Generation}
        The code generation task has made significant strides due to the development of LLMs, which improves software engineering productivity by a large margin.
        To improve the code language model through fine-tuning,
        the reward signals can come from various sources, including \cite{zhu2024deepseek-coder-v2},
        and code compiler feedback, unit test results, and code analysis \cite{liu2023rltf-unittest,dou2024stepcoder,tao2024codelutra,zhou2025refinecoder}.
        For example, DeepSeek-Coder-V2 \cite{zhu2024deepseek-coder-v2} trains a reward model for code generation and fine-tunes the language model via the reinforcement learning algorithm GRPO \cite{shao2024deepseekmath}.
        Additionally, some approaches guide the inference of language models during code generation with reward models,
        including the generate-then-rank \cite{ni2023lever,hosseini2024vstar}
        and reward-guided decoding \cite{yu2024orps}.
        Another popular direction refines the generated code to correct errors and bugs through the language model itself \cite{shinn2023reflexion,zhang2023self-edit,chen2023self-debug} or external feedback \cite{xie2025teaching-ctrl}.

    \subsection{Multimodal Tasks}
        Learning-from-rewards strategies have been widely applied to multimodal tasks, including multimodal understanding and generation.
        Most studies adopt reinforcement learning and reward-guided decoding methods.
        For instance, \textit{Q-Insight} \cite{li2025q-insight-image-quality} focuses on improving comprehensive image quality understanding with reinforcement learning.
        \textit{VLM-R1} \cite{shen2025vlm-r1} applies reinforcement learning to fine-tune vision-language models and focuses on two tasks: referring expression compression and object detection.
        \textit{Vision-R1} \cite{huang2025vision-r1-reasoning} enhances multimodal reasoning of vision-language models for mathematical VQA.
        \citet{zhan2025vision-r1-object-localization} proposes another \textit{Vision-R1}, but it mainly facilitates object localization tasks with vision-language models.

        \textit{Video-R1} \cite{feng2025video-r1}, \textit{VideoChat-R1} \cite{li2025videochat-r1}, and \textit{TinyLLaVA-Video-R1} \cite{zhang2025tinyllava-video-r1} apply GRPO into video reasoning.
        \textit{R1-V} \cite{chen2025r1v} and \textit{CrowdVLM-R1} \cite{wang2025crowdvlm-r1} focus on visual counting.
        More example applications include multimodal reasoning \cite{zhou2025r1-zero-aha-moment,meng2025mm-eureka,tan2025reason-rft,li2025relation-r1,liu2025othink-mr1}, object detection \cite{liu2025visualrft}, segmentation \cite{liu2025seg-zero}, and image/video generation \cite{guo2025parm-image-generation,liu2025video-t1}.

    \subsection{Agents}
        LLM Agent is an autonomous system that automatically performs complex tasks through task decomposition and action execution in dynamic environments \cite{wang2024survey-agents}.
        Various learning-from-rewards strategies have been applied to training or guiding the agents.
        \textit{AgentRM} \cite{xia2025agentrm} targets general-purpose decision-making agents across domains such as web navigation, embodied planning, text games, and tool use.
        During inference, a reward model guides the agents to choose candidate actions or trajectories.
        \textit{AgentPRM} \cite{choudhury2025agentprm} trains LLM agents with a process reward model.
        \textit{KBQA-o1} \cite{luo2025kbqa-o1} guides MCTS with a reward model for the knowledge base question answering task with agents.
        \textit{DeepResearch} \cite{openai-deepresearch} and \textit{DeepResearcher} \cite{zheng2025deepresearcher} design agents for research tasks.
        They both use reinforcement learning to fine-tune the agents.
        \textit{UI-R1} \cite{lu2025ui-r1} introduces a rule-based reinforcement learning framework for GUI action prediction with multimodal agents.
        \textit{InfiGUI-R1} \cite{liu2025infigui} is a similar work with GUI agents.
        \textit{RAGEN} \cite{wang2025ragen-self-evolution-agents} propose training the agents via multi-turn reinforcement learning with a new algorithm based on GRPO.

    \subsection{Other Applications}
        Many other applications have been developed following the learning-from-rewards strategies.

        Embodied AI is essential for the development of artificial general intelligence.
        AI systems, such as embodied robots, must interact with the physical world and complete complex tasks through high-level planning and low-level control.
        They generally aim to enhance the embodied reasoning abilities with reinforcement learning,
        such as 
        \textit{Cosmos-reason1} \cite{azzolini2025cosmos-reason1},
        \textit{iRe-VLA} \cite{guo2025vision-language-action},
        \textit{Embodied-Reasoner} \cite{zhang2025embodied-reasoner},
        and \textit{Embodied-R} \cite{zhao2025embodied-r}.

        Several approaches apply reinforcement learning to reason with information retrieval from knowledge databases or the real-world web.
        These approaches include \textit{R1-Searcher} \cite{song2025r1-searcher}, \textit{Search-R1} \cite{jin2025search-r1}, \textit{DeepRetrieval} \cite{jiang2025deepretrieval}, \textit{ReSearch} \cite{chen2025re-search}, and \textit{WebThinker} \cite{li2025webthinker}.
        They adopt different reward designs to improve search performance.

        Applications for other applications also emerge.
        \textit{ToRL} \cite{li2025torl}, \textit{ReTool} \cite{feng2025retool}, \textit{SWi-RL} \cite{goldie2025swirl-tooluse}, \textit{ToolRL} \cite{qian2025toolrl} and \textit{OTC} \cite{wang2025tool-calls}
        are proposed to improve LLMs' reasoning ability to call various tools through reinforcement learning.
        \textit{Rec-R1} \cite{lin2025rec-r1} applies reinforcement learning for recommendation system.
        \textit{SWE-RL} \cite{wei2025swe-rl} aims at software engineering with reinforcement learning.
        \textit{SQL-R1} \cite{ma2025sql-r1} focuses on natural language to SQL reasoning.
        It uses a composite reward function with format correctness, execution success, result accuracy, and reasoning completeness.

        Some applications are designed for specific areas.
        \textit{Med-R1} \citet{lai2025med-r1} and \textit{MedVLM-R1} \cite{pan2025medvlm-r1} are proposed for medical field.
        They target medical VQA across various imaging modalities (\eg CT, MRT, and X-ray) and several clinical tasks, like diagnosis, and anatomy identification.
        \textit{Fin-R1} \cite{liu2025fin-r1} develops LLMs for the financial field, targeting financial QA and decision-making.
        It leverages accuracy and format rule-based rewards to train a language model on domain-specific data.
        \textit{DianJin-R1} \cite{zhu2025dianjin-r1} is another LLM for the financial field with reinforcement learning.

\begin{figure*}[!t]
    \centering
    \includegraphics[width=0.8\linewidth]{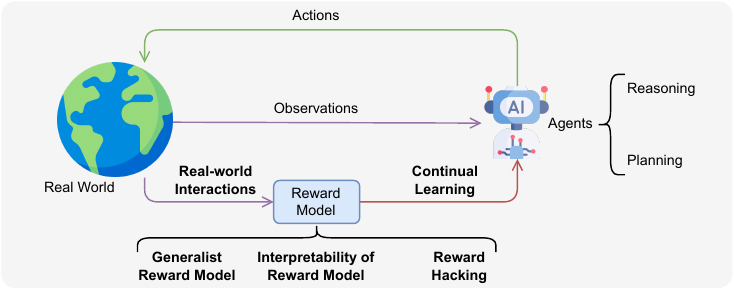}
    \caption{
        Illustration of challenges and future directions.
    }
    \label{fig_future_directions}
\end{figure*}

\section{Challenges and Future Directions} \label{sec_future_directions}
    In this section, we discuss the current challenges and future directions of learning from rewards.
    \Cref{fig_future_directions} summarizes the key challenges and future directions from the perspective of reward models and learning strategies.
    Ultimately, we envision the development of interpretable, robust, and continually evolving agent systems capable of interacting with and adapting to the complexities of the real world.

    \subsection{Interpretability of Reward Models}
        Interpretability of reward models remains an open challenge for the learning-from-rewards strategies \cite{russell2019explaining-reward-functions,zhang2023interpretable-reward-distribution,jenner2022preprocessing-reward-interpretability}.
        Most reward models are typically treated as black boxes that produce scalars or critiques without exposing human-interpretable explanations.
        Such opacity hinders human trust and oversight and may lead to misaligned optimization.
        In consequence, enhancing reward model interpretability is essential for reliable alignment, enabling humans to inspect and verify the internal decision process and steer models toward desired behavior.
        Recent efforts have attempted to address this issue.
        For instance, \textit{ArmoRM} \cite{wang2024armorm} improves the interpretability with multi-objective reward modeling,
        where each objective corresponds to a human-interpretable dimension, such as helpfulness, correctness, coherence, complexity, and verbosity.
        While this approach is effective, its interpretability is limited to these predefined objectives.
        In addition, emerging generative reward models can disclose their rationales of reward scoring \cite{zhao2025genprm,khalifa2025thinkprm}.
        While promising, their interpretability remains limited and demands further investigation into consistency, reliability, and faithfulness.

    \subsection{Generalist Reward Models}
        A promising future direction is the development of \textit{generalist reward models}.
        Most existing reward models are designed for narrow domains;
        thus they often suffer from weak generalization across tasks.
        Moreover, their reward outputs are typically static and lack support for inference-time scalability, hindering their application in diverse and open-ended scenarios \cite{liu2024skywork,zhang2024genrm,snell2025scaling}.

        In contrast, a generalist reward model seeks to overcome these limitations.
        They demand flexibility for input types, including single, paired, or multiple responses,
        and also require accurate reward generation in various domains, such as question answering, math reasoning, and code generation.
        Besides, they are expected to generate higher-quality reward signals with increased inference-time computing.
        Such models offer a unified interface for reward modeling across domains and enable scalable, interpretable reward generation.
        For example, \textit{DeepSeek-GRM} \cite{liu2025deepseek-grm}, a recent attempt in this direction, proposes a pointwise generative reward model.
        Rather than only scalars, it can generate evaluative natural language principles and critiques, enabling effective inference-time scaling through multi-sample voting and meta-reward filtering.

    \subsection{Reward Hacking}
        Reward hacking is a fundamental challenge in learning from rewards \cite{everitt2021reward-tampering,amodei2016concrete-reward-hacking,weng2024rewardhack,liu2025robust-reward-model}.
        It occurs when models exploit unintended shortcuts in the reward function to obtain high rewards without truly learning the desired behaviors
        or completing the task as designed.
        This phenomenon has been observed across domains.
        For instance, LLMs may fabricate plausible yet incorrect answers, and code LLMs subtly modify unit tests to pass evaluations \cite{denison2024sycophancy}.
        Reward backing can also happen during inference, called \textit{in-context reward hacking} \cite{pan2024reward-hacking-self-refinement,pan2024feedback-loops-in-context-reward-hacking}.
        It arises in self-refinement loops where the same model acts as both the generator and the judge.
        In such cases, the model may learn to produce outputs that exploit its own evaluation heuristics, leading to inflated internal scores while 
        deviating from true objectives.

        Reward hacking fundamentally arises from the difficulty of specifying a reward function that perfectly captures the true objectives.
        As articulated by Goodhart's Laws---\textit{When a measure becomes a target, it ceases to be a good measure}---any proxy metric used as a reward will eventually be exploited once applying optimization pressure.
        To mitigate reward hacking, the following directions are worth exploring:
        \begin{inparaenum}[(i)]
            \item Designing more robust and tamper-resistant reward functions \cite{razin2025what-rm-good-teacher,shen2025data-scaling-rlhf,peng2025agentic-reward-modeling};
            \item Detecting misalignment via behavioral or distributional anomaly detection \cite{pan2022effects-reard-misspecification};
            \item Decoupling feedback mechanisms to prevent contamination \cite{uesato2020avoiding-decoupled-approval};
            \item Auditing the dataset for training reward models to reduce reward hacking risks \cite{revel2025seal}.
        \end{inparaenum}

    \subsection{Grounded Rewards from Real-World Interactions}
        Despite recent advances in learning from rewards for LLMs, most methods fundamentally rely on human preferences or well-curated automated feedback.
        The LLMs are typically optimized to maximize the rewards derived from these feedback.
        In consequence, this inherently limits the agent's ability to surpass existing human knowledge and adapt to complex environments.

        Due to these limitations, moving beyond chat-driven rewards toward grounded real-world rewards is another promising direction.
        This movement requires LLMs to be integrated into agentic frameworks,
        and agents should increasingly interact directly with their environment and derive reward signals from observed outcomes.
        For example, a health assistant could optimize behavior based on physiological signals rather than user ratings, and a scientific agent could refine hypotheses based on experimental data rather than expert approval \cite{silver2025welcome-era-experience}.
        This shift would enable agents to close the feedback loop with the real world, allowing for autonomous discovery, adaptation, and pursuit of goals beyond human understanding.
        The transition to real-world interactions raises substantial technical challenges.
        Agents must handle noisy, delayed, or partial feedback from complex environments, requiring advances in credit assignment, robust exploration, and uncertainty modeling.

    \subsection{Continual Learning from Rewards}
        Current learning-from-rewards strategies often assume a fixed dataset, a predefined reward model, and short episodic interactions.
        Once trained, models typically exhibit limited abilities to adapt to new tasks or evolving environments \cite{zhang2024continual-cppo,silver2025welcome-era-experience}.
        This episodic and offline paradigm sharply contrasts with real-world intelligence's dynamic, ongoing nature, where agents must continually learn from experience and recalibrate based on new feedback.

        As such, a vital direction is continual learning from rewards.
        It is a crucial foundation for building lifelong competent and aligned agents.
        By abandoning the traditional assumption of fixed objectives, models can remain responsive to changing reward signals,
        avoid performance degradation under distributional shifts, and better reflect long-term user intent.
        Notably, it is a broader idea of continual reinforcement learning \cite{abel2023definition-continual-rl,li2024continual-multi-objective-rl,bowling2025continual-rl}.
        Achieving continual learning from rewards presents significant challenges.
        It requires addressing catastrophic forgetting, maintaining stability while enabling plasticity, and designing dynamic reward modeling mechanisms.

\section{Conclusion}
    In this paper, we present a comprehensive survey of learning from rewards in LLMs.
    We categorize the landscape into three key stages---training, inference, and post-inference, each reflecting a distinct paradigm for integrating reward signals into steering LLMs' behavior.
    For each stage, we review representative studies in terms of reward models and learning strategies.
    In addition, we summarize recent progress in benchmarking reward models and applications.
    Finally we identify core challenges and outline promising future directions.
    We hope this survey provides a structured understanding of the field and inspires future research.

\bibliography{lib}

\begin{thebibliography}{288}
\providecommand{\natexlab}[1]{#1}

\bibitem[{Abel et~al.(2023)Abel, Barreto, Van~Roy, Precup, van Hasselt, and Singh}]{abel2023definition-continual-rl}
David Abel, Andr{\'e} Barreto, Benjamin Van~Roy, Doina Precup, Hado~P van Hasselt, and Satinder Singh. 2023.
\newblock \href {https://arxiv.org/pdf/2307.11046} {A definition of continual reinforcement learning}.
\newblock \emph{Advances in Neural Information Processing Systems}, 36:50377--50407.

\bibitem[{Ahn et~al.(2024)Ahn, Choi, Yu, Kang, and Choi}]{ahn2024vlm-rlaif}
Daechul Ahn, Yura Choi, Youngjae Yu, Dongyeop Kang, and Jonghyun Choi. 2024.
\newblock \href {{https://arxiv.org/pdf/2402.03746}} {Tuning large multimodal models for videos using reinforcement learning from ai feedback}.
\newblock \emph{arXiv preprint arXiv:2402.03746}.

\bibitem[{Akyurek et~al.(2023)Akyurek, Akyurek, Kalyan, Clark, Wijaya, and Tandon}]{akyurek2023rl4f}
Afra~Feyza Akyurek, Ekin Akyurek, Ashwin Kalyan, Peter Clark, Derry~Tanti Wijaya, and Niket Tandon. 2023.
\newblock \href {https://doi.org/10.18653/v1/2023.acl-long.427} {{RL}4{F}: Generating natural language feedback with reinforcement learning for repairing model outputs}.
\newblock In \emph{Proceedings of the 61st Annual Meeting of the Association for Computational Linguistics (Volume 1: Long Papers)}, pages 7716--7733, Toronto, Canada. Association for Computational Linguistics.

\bibitem[{Amodei et~al.(2016)Amodei, Olah, Steinhardt, Christiano, Schulman, and Man{\'e}}]{amodei2016concrete-reward-hacking}
Dario Amodei, Chris Olah, Jacob Steinhardt, Paul Christiano, John Schulman, and Dan Man{\'e}. 2016.
\newblock \href {{https://arxiv.org/pdf/1606.06565}} {Concrete problems in ai safety}.
\newblock \emph{arXiv preprint arXiv:1606.06565}.

\bibitem[{Ankner et~al.(2024)Ankner, Paul, Cui, Chang, and Ammanabrolu}]{ankner2024critique-out-loud-rm}
Zachary Ankner, Mansheej Paul, Brandon Cui, Jonathan~D Chang, and Prithviraj Ammanabrolu. 2024.
\newblock \href {{https://arxiv.org/pdf/2408.11791}} {Critique-out-loud reward models}.
\newblock \emph{arXiv preprint arXiv:2408.11791}.

\bibitem[{Anthropic(2025)}]{claude3.7sonnet}
Anthropic. 2025.
\newblock \href {https://www.anthropic.com/news/claude-3-7-sonnet} {Introducing deep research}.

\bibitem[{Azzolini et~al.(2025)Azzolini, Brandon, Chattopadhyay, Chen, Chu, Cui, Diamond, Ding, Ferroni, Govindaraju et~al.}]{azzolini2025cosmos-reason1}
Alisson Azzolini, Hannah Brandon, Prithvijit Chattopadhyay, Huayu Chen, Jinju Chu, Yin Cui, Jenna Diamond, Yifan Ding, Francesco Ferroni, Rama Govindaraju, et~al. 2025.
\newblock \href {{https://arxiv.org/pdf/2503.15558}} {Cosmos-reason1: From physical common sense to embodied reasoning}.
\newblock \emph{arXiv preprint arXiv:2503.15558}.

\bibitem[{Bai et~al.(2022{\natexlab{a}})Bai, Jones, Ndousse, Askell, Chen, DasSarma, Drain, Fort, Ganguli, Henighan et~al.}]{bai2022training}
Yuntao Bai, Andy Jones, Kamal Ndousse, Amanda Askell, Anna Chen, Nova DasSarma, Dawn Drain, Stanislav Fort, Deep Ganguli, Tom Henighan, et~al. 2022{\natexlab{a}}.
\newblock \href {{https://arxiv.org/pdf/2204.05862}} {Training a helpful and harmless assistant with reinforcement learning from human feedback}.
\newblock \emph{arXiv preprint arXiv:2204.05862}.

\bibitem[{Bai et~al.(2022{\natexlab{b}})Bai, Kadavath, Kundu, Askell, Kernion, Jones, Chen, Goldie, Mirhoseini, McKinnon et~al.}]{bai2022constitutional}
Yuntao Bai, Saurav Kadavath, Sandipan Kundu, Amanda Askell, Jackson Kernion, Andy Jones, Anna Chen, Anna Goldie, Azalia Mirhoseini, Cameron McKinnon, et~al. 2022{\natexlab{b}}.
\newblock \href {{https://arxiv.org/pdf/2212.08073}} {Constitutional {AI}: Harmlessness from {AI} feedback}.
\newblock \emph{arXiv preprint arXiv:2212.08073}.

\bibitem[{bench authors(2023)}]{srivastava2023beyond}
BIG bench authors. 2023.
\newblock \href {https://openreview.net/forum?id=uyTL5Bvosj} {Beyond the imitation game: Quantifying and extrapolating the capabilities of language models}.
\newblock \emph{Transactions on Machine Learning Research}.

\bibitem[{Bowling and Elelimy(2025)}]{bowling2025continual-rl}
Michael Bowling and Esraa Elelimy. 2025.
\newblock \href {{https://arxiv.org/pdf/2504.08161}} {Rethinking the foundations for continual reinforcement learning}.
\newblock \emph{arXiv preprint arXiv:2504.08161}.

\bibitem[{Bradley and Terry(1952)}]{bradley1952rank}
Ralph~Allan Bradley and Milton~E Terry. 1952.
\newblock \href {https://www.jstor.org/stable/2334029} {Rank analysis of incomplete block designs: I. the method of paired comparisons}.
\newblock \emph{Biometrika}, 39(3/4):324--345.

\bibitem[{Brown et~al.(2024)Brown, Juravsky, Ehrlich, Clark, Le, R{\'e}, and Mirhoseini}]{brown2024language-monkeys}
Bradley Brown, Jordan Juravsky, Ryan Ehrlich, Ronald Clark, Quoc~V Le, Christopher R{\'e}, and Azalia Mirhoseini. 2024.
\newblock \href {{https://arxiv.org/pdf/2407.21787}} {Large language monkeys: Scaling inference compute with repeated sampling}.
\newblock \emph{arXiv preprint arXiv:2407.21787}.

\bibitem[{Cao et~al.(2024)Cao, Lam, Duan, Liu, Zhang, and Chen}]{cao2024compassjudger}
Maosong Cao, Alexander Lam, Haodong Duan, Hongwei Liu, Songyang Zhang, and Kai Chen. 2024.
\newblock \href {{https://arxiv.org/pdf/2410.16256}} {Compassjudger-1: All-in-one judge model helps model evaluation and evolution}.
\newblock \emph{arXiv preprint arXiv:2410.16256}.

\bibitem[{Chen et~al.(2024{\natexlab{a}})Chen, Chen, Zhang, Wang, Liu, Zhou, Zhang, Wan, Zhou, and Sun}]{chen2024mllm-as-a-judge}
Dongping Chen, Ruoxi Chen, Shilin Zhang, Yaochen Wang, Yinuo Liu, Huichi Zhou, Qihui Zhang, Yao Wan, Pan Zhou, and Lichao Sun. 2024{\natexlab{a}}.
\newblock \href {https://arxiv.org/pdf/2402.04788} {Mllm-as-a-judge: Assessing multimodal llm-as-a-judge with vision-language benchmark}.
\newblock In \emph{Forty-first International Conference on Machine Learning}.

\bibitem[{Chen et~al.(2025{\natexlab{a}})Chen, Li, Zhao, Song, and Vinci}]{chen2025r1v}
Liang Chen, Lei Li, Haozhe Zhao, Yifan Song, and Vinci. 2025{\natexlab{a}}.
\newblock \href {https://github.com/Deep-Agent/R1-V} {R1-v: Reinforcing super generalization ability in vision-language models with less than \$3}.

\bibitem[{Chen et~al.(2025{\natexlab{b}})Chen, Li, Sun, Zhou, Zhu, Wang, Pan, Zhang, Chen, Yang, Zhou, and Chen}]{chen2025re-search}
Mingyang Chen, Tianpeng Li, Haoze Sun, Yijie Zhou, Chenzheng Zhu, Haofen Wang, Jeff~Z. Pan, Wen Zhang, Huajun Chen, Fan Yang, Zenan Zhou, and Weipeng Chen. 2025{\natexlab{b}}.
\newblock \href {https://arxiv.org/pdf/2503.19470} {Research: Learning to reason with search for llms via reinforcement learning}.
\newblock \emph{arXiv preprint arXiv:2503.19470}.

\bibitem[{Chen et~al.(2023)Chen, Lin, Sch{\"a}rli, and Zhou}]{chen2023self-debug}
Xinyun Chen, Maxwell Lin, Nathanael Sch{\"a}rli, and Denny Zhou. 2023.
\newblock \href {{https://arxiv.org/pdf/2304.05128}} {Teaching large language models to self-debug}.
\newblock \emph{arXiv preprint arXiv:2304.05128}.

\bibitem[{Chen et~al.(2024{\natexlab{b}})Chen, Sikka, Cogswell, Ji, and Divakaran}]{chen2024dress}
Yangyi Chen, Karan Sikka, Michael Cogswell, Heng Ji, and Ajay Divakaran. 2024{\natexlab{b}}.
\newblock \href {https://arxiv.org/pdf/2311.10081} {Dress: Instructing large vision-language models to align and interact with humans via natural language feedback}.
\newblock In \emph{Proceedings of the IEEE/CVF Conference on Computer Vision and Pattern Recognition}, pages 14239--14250.

\bibitem[{Chen et~al.(2024{\natexlab{c}})Chen, Du, Wen, Zhou, Cui, Weng, Tu, Wang, Tong, Huang et~al.}]{chen2024mj-bench}
Zhaorun Chen, Yichao Du, Zichen Wen, Yiyang Zhou, Chenhang Cui, Zhenzhen Weng, Haoqin Tu, Chaoqi Wang, Zhengwei Tong, Qinglan Huang, et~al. 2024{\natexlab{c}}.
\newblock \href {{https://arxiv.org/pdf/2407.04842}} {Mj-bench: Is your multimodal reward model really a good judge for text-to-image generation?}
\newblock \emph{arXiv preprint arXiv:2407.04842}.

\bibitem[{Chern et~al.(2023)Chern, Chern, Chen, Yuan, Feng, Zhou, He, Neubig, Liu et~al.}]{chern2023factool}
I~Chern, Steffi Chern, Shiqi Chen, Weizhe Yuan, Kehua Feng, Chunting Zhou, Junxian He, Graham Neubig, Pengfei Liu, et~al. 2023.
\newblock \href {{https://arxiv.org/pdf/2307.13528}} {Factool: Factuality detection in generative ai--a tool augmented framework for multi-task and multi-domain scenarios}.
\newblock \emph{arXiv preprint arXiv:2307.13528}.

\bibitem[{Choudhury(2025)}]{choudhury2025agentprm}
Sanjiban Choudhury. 2025.
\newblock \href {{https://arxiv.org/pdf/2502.10325}} {Process reward models for llm agents: Practical framework and directions}.
\newblock \emph{arXiv preprint arXiv:2502.10325}.

\bibitem[{Cobbe et~al.(2021)Cobbe, Kosaraju, Bavarian, Chen, Jun, Kaiser, Plappert, Tworek, Hilton, Nakano et~al.}]{cobbe2021training-verifiers}
Karl Cobbe, Vineet Kosaraju, Mohammad Bavarian, Mark Chen, Heewoo Jun, Lukasz Kaiser, Matthias Plappert, Jerry Tworek, Jacob Hilton, Reiichiro Nakano, et~al. 2021.
\newblock \href {{https://arxiv.org/pdf/2110.14168}} {Training verifiers to solve math word problems}.
\newblock \emph{arXiv preprint arXiv:2110.14168}.

\bibitem[{Cohen et~al.(2023)Cohen, Hamri, Geva, and Globerson}]{cohen2023lm-vs-lm}
Roi Cohen, May Hamri, Mor Geva, and Amir Globerson. 2023.
\newblock \href {{https://arxiv.org/pdf/2305.13281}} {Lm vs lm: Detecting factual errors via cross examination}.
\newblock \emph{arXiv preprint arXiv:2305.13281}.

\bibitem[{Cui et~al.(2025)Cui, Yuan, Wang, Wang, Li, He, Fan, Yu, Xu, Chen et~al.}]{cui2025prime-process-implicit-reward}
Ganqu Cui, Lifan Yuan, Zefan Wang, Hanbin Wang, Wendi Li, Bingxiang He, Yuchen Fan, Tianyu Yu, Qixin Xu, Weize Chen, et~al. 2025.
\newblock \href {{https://arxiv.org/pdf/2502.01456}} {Process reinforcement through implicit rewards}.
\newblock \emph{arXiv preprint arXiv:2502.01456}.

\bibitem[{Dai et~al.(2023)Dai, Pan, Sun, Ji, Xu, Liu, Wang, and Yang}]{dai2023safe-rlhf}
Josef Dai, Xuehai Pan, Ruiyang Sun, Jiaming Ji, Xinbo Xu, Mickel Liu, Yizhou Wang, and Yaodong Yang. 2023.
\newblock \href {{https://arxiv.org/pdf/2310.12773}} {Safe rlhf: Safe reinforcement learning from human feedback}.
\newblock \emph{arXiv preprint arXiv:2310.12773}.

\bibitem[{DeepSeek-AI(2025)}]{deepseekproverv2-2025}
DeepSeek-AI. 2025.
\newblock \href {https://arxiv.org/pdf/2504.21801} {Deepseek-prover-v2: Advancing formal mathematical reasoning via reinforcement learning for subgoal decomposition}.
\newblock \emph{arXiv preprint arXiv:2504.21801}.

\bibitem[{DeepSeek-AI et~al.(2025)}]{deepseekai2025deepseek-r1}
DeepSeek-AI et~al. 2025.
\newblock \href {{https://arxiv.org/pdf/2501.12948}} {Deepseek-r1: Incentivizing reasoning capability in llms via reinforcement learning}.
\newblock \emph{arXiv preprint arXiv:2501.12948}.

\bibitem[{Deng and Raffel(2023)}]{deng2023rad}
Haikang Deng and Colin Raffel. 2023.
\newblock \href {{https://arxiv.org/pdf/2310.09520}} {Reward-augmented decoding: Efficient controlled text generation with a unidirectional reward model}.
\newblock \emph{arXiv preprint arXiv:2310.09520}.

\bibitem[{Denison et~al.(2024)Denison, MacDiarmid, Barez, Duvenaud, Kravec, Marks, Schiefer, Soklaski, Tamkin, Kaplan et~al.}]{denison2024sycophancy}
Carson Denison, Monte MacDiarmid, Fazl Barez, David Duvenaud, Shauna Kravec, Samuel Marks, Nicholas Schiefer, Ryan Soklaski, Alex Tamkin, Jared Kaplan, et~al. 2024.
\newblock \href {{https://arxiv.org/pdf/2406.10162}} {Sycophancy to subterfuge: Investigating reward-tampering in large language models}.
\newblock \emph{arXiv preprint arXiv:2406.10162}.

\bibitem[{Deshpande et~al.(2023)Deshpande, Murahari, Rajpurohit, Kalyan, and Narasimhan}]{deshpande2023toxicity-chatgpt}
Ameet Deshpande, Vishvak Murahari, Tanmay Rajpurohit, Ashwin Kalyan, and Karthik Narasimhan. 2023.
\newblock \href {{https://arxiv.org/pdf/2304.05335}} {Toxicity in chatgpt: Analyzing persona-assigned language models}.
\newblock \emph{arXiv preprint arXiv:2304.05335}.

\bibitem[{Dhuliawala et~al.(2023)Dhuliawala, Komeili, Xu, Raileanu, Li, Celikyilmaz, and Weston}]{dhuliawala2023chain-of-verification}
Shehzaad Dhuliawala, Mojtaba Komeili, Jing Xu, Roberta Raileanu, Xian Li, Asli Celikyilmaz, and Jason Weston. 2023.
\newblock \href {{https://arxiv.org/pdf/2309.11495}} {Chain-of-verification reduces hallucination in large language models}.
\newblock \emph{arXiv preprint arXiv:2309.11495}.

\bibitem[{Ding et~al.(2024)Ding, Min, Kaiser, and Ray}]{ding2024cycle}
Yangruibo Ding, Marcus~J Min, Gail Kaiser, and Baishakhi Ray. 2024.
\newblock \href {https://arxiv.org/pdf/2403.18746} {Cycle: Learning to self-refine the code generation}.
\newblock \emph{Proceedings of the ACM on Programming Languages}, 8(OOPSLA1):392--418.

\bibitem[{Dong et~al.(2023)Dong, Xiong, Goyal, Zhang, Chow, Pan, Diao, Zhang, Shum, and Zhang}]{dong2023raft}
Hanze Dong, Wei Xiong, Deepanshu Goyal, Yihan Zhang, Winnie Chow, Rui Pan, Shizhe Diao, Jipeng Zhang, Kashun Shum, and Tong Zhang. 2023.
\newblock \href {{https://arxiv.org/pdf/2304.06767}} {Raft: Reward ranked finetuning for generative foundation model alignment}.
\newblock \emph{arXiv preprint arXiv:2304.06767}.

\bibitem[{Dou et~al.(2024)Dou, Liu, Jia, Xiong, Zhou, Shen, Shan, Huang, Wang, Fan et~al.}]{dou2024stepcoder}
Shihan Dou, Yan Liu, Haoxiang Jia, Limao Xiong, Enyu Zhou, Wei Shen, Junjie Shan, Caishuang Huang, Xiao Wang, Xiaoran Fan, et~al. 2024.
\newblock \href {{https://arxiv.org/pdf/2402.01391}} {Stepcoder: Improve code generation with reinforcement learning from compiler feedback}.
\newblock \emph{arXiv preprint arXiv:2402.01391}.

\bibitem[{Du et~al.(2023)Du, Li, Torralba, Tenenbaum, and Mordatch}]{du2023factulaity-multiagent-debate}
Yilun Du, Shuang Li, Antonio Torralba, Joshua~B Tenenbaum, and Igor Mordatch. 2023.
\newblock \href {{https://arxiv.org/pdf/2305.14325}} {Improving factuality and reasoning in language models through multiagent debate}.
\newblock \emph{arXiv preprint arXiv:2305.14325}.

\bibitem[{Duan et~al.(2025)Duan, Liu, Mao, Pang, Chen, Chen, Shieh, and Dou}]{duan2025actprm-efficient}
Keyu Duan, Zichen Liu, Xin Mao, Tianyu Pang, Changyu Chen, Qiguang Chen, Michael~Qizhe Shieh, and Longxu Dou. 2025.
\newblock \href {{https://arxiv.org/pdf/2504.10559}} {Efficient process reward model training via active learning}.
\newblock \emph{arXiv preprint arXiv:2504.10559}.

\bibitem[{Dutta et~al.(2024)Dutta, Mahinder, Anantha, and Bandyopadhyay}]{dutta2024rlaif-codeapi}
Sujan Dutta, Sayantan Mahinder, Raviteja Anantha, and Bortik Bandyopadhyay. 2024.
\newblock \href {https://aclanthology.org/2024.nlrse-1.4/} {Applying {RLAIF} for code generation with {API}-usage in lightweight {LLM}s}.
\newblock In \emph{Proceedings of the 2nd Workshop on Natural Language Reasoning and Structured Explanations (@ACL 2024)}, pages 39--45, Bangkok, Thailand. Association for Computational Linguistics.

\bibitem[{Ethayarajh et~al.(2024)Ethayarajh, Xu, Muennighoff, Jurafsky, and Kiela}]{ethayarajh2024kto}
Kawin Ethayarajh, Winnie Xu, Niklas Muennighoff, Dan Jurafsky, and Douwe Kiela. 2024.
\newblock \href {{https://arxiv.org/pdf/2402.01306}} {{KTO:} model alignment as prospect theoretic optimization}.
\newblock \emph{arXiv preprint arXiv:2402.01306}.

\bibitem[{Everitt et~al.(2021)Everitt, Hutter, Kumar, and Krakovna}]{everitt2021reward-tampering}
Tom Everitt, Marcus Hutter, Ramana Kumar, and Victoria Krakovna. 2021.
\newblock \href {https://arxiv.org/pdf/1908.04734} {Reward tampering problems and solutions in reinforcement learning: A causal influence diagram perspective}.
\newblock \emph{Synthese}, 198(Suppl 27):6435--6467.

\bibitem[{Face(2025)}]{huggingface2025open-r1}
Hugging Face. 2025.
\newblock \href {https://github.com/huggingface/open-r1} {Open r1: A fully open reproduction of deepseek-r1}.

\bibitem[{Feng et~al.(2025{\natexlab{a}})Feng, Huang, Qu, Zhang, Qin, Zhong, Jiang, Chi, and Zhong}]{feng2025retool}
Jiazhan Feng, Shijue Huang, Xingwei Qu, Ge~Zhang, Yujia Qin, Baoquan Zhong, Chengquan Jiang, Jinxin Chi, and Wanjun Zhong. 2025{\natexlab{a}}.
\newblock \href {{https://arxiv.org/pdf/2504.11536}} {Retool: Reinforcement learning for strategic tool use in llms}.
\newblock \emph{arXiv preprint arXiv:2504.11536}.

\bibitem[{Feng et~al.(2025{\natexlab{b}})Feng, Gong, Li, Guo, Wang, Peng, Wang, and Yue}]{feng2025video-r1}
Kaituo Feng, Kaixiong Gong, Bohao Li, Zonghao Guo, Yibing Wang, Tianshuo Peng, Benyou Wang, and Xiangyu Yue. 2025{\natexlab{b}}.
\newblock \href {{https://arxiv.org/pdf/2503.21776}} {Video-r1: Reinforcing video reasoning in mllms}.
\newblock \emph{arXiv preprint arXiv:2503.21776}.

\bibitem[{Feng et~al.(2025{\natexlab{c}})Feng, Wang, Bai, Su, Wu, Yu, and Wang}]{feng2025grpo-curriculum-sampling}
Zihao Feng, Xiaoxue Wang, Ziwei Bai, Donghang Su, Bowen Wu, Qun Yu, and Baoxun Wang. 2025{\natexlab{c}}.
\newblock \href {{https://arxiv.org/pdf/2504.13592}} {Improving generalization in intent detection: Grpo with reward-based curriculum sampling}.
\newblock \emph{arXiv preprint arXiv:2504.13592}.

\bibitem[{First et~al.(2023)First, Rabe, Ringer, and Brun}]{first2023baldur-proof-geneartion}
Emily First, Markus~N Rabe, Talia Ringer, and Yuriy Brun. 2023.
\newblock \href {https://arxiv.org/pdf/2303.04910} {Baldur: Whole-proof generation and repair with large language models}.
\newblock In \emph{Proceedings of the 31st ACM Joint European Software Engineering Conference and Symposium on the Foundations of Software Engineering}, pages 1229--1241.

\bibitem[{Frick et~al.(2024)Frick, Li, Chen, Chiang, Angelopoulos, Jiao, Zhu, Gonzalez, and Stoica}]{frick2024evaluate-ppe}
Evan Frick, Tianle Li, Connor Chen, Wei-Lin Chiang, Anastasios~N Angelopoulos, Jiantao Jiao, Banghua Zhu, Joseph~E Gonzalez, and Ion Stoica. 2024.
\newblock \href {{https://arxiv.org/pdf/2410.14872}} {How to evaluate reward models for rlhf}.
\newblock \emph{arXiv preprint arXiv:2410.14872}.

\bibitem[{Gao et~al.(2024{\natexlab{a}})Gao, Cai, Xu, Wang, Zheng, Lin, Lu, Liu, Zhou, Xiao et~al.}]{gao2024math-minos-critics-bugs-math}
Bofei Gao, Zefan Cai, Runxin Xu, Peiyi Wang, Ce~Zheng, Runji Lin, Keming Lu, Dayiheng Liu, Chang Zhou, Wen Xiao, et~al. 2024{\natexlab{a}}.
\newblock \href {{https://arxiv.org/pdf/2406.14024}} {Llm critics help catch bugs in mathematics: Towards a better mathematical verifier with natural language feedback}.
\newblock \emph{arXiv preprint arXiv:2406.14024}.

\bibitem[{Gao et~al.(2024{\natexlab{b}})Gao, Song, Yang, Cai, Miao, Dong, Li, Ma, Chen, Xu et~al.}]{gao2024omni-math}
Bofei Gao, Feifan Song, Zhe Yang, Zefan Cai, Yibo Miao, Qingxiu Dong, Lei Li, Chenghao Ma, Liang Chen, Runxin Xu, et~al. 2024{\natexlab{b}}.
\newblock \href {{https://arxiv.org/pdf/2410.07985}} {Omni-math: A universal olympiad level mathematic benchmark for large language models}.
\newblock \emph{arXiv preprint arXiv:2410.07985}.

\bibitem[{Gao et~al.(2022)Gao, Dai, Pasupat, Chen, Chaganty, Fan, Zhao, Lao, Lee, Juan et~al.}]{gao2022rarr}
Luyu Gao, Zhuyun Dai, Panupong Pasupat, Anthony Chen, Arun~Tejasvi Chaganty, Yicheng Fan, Vincent~Y Zhao, Ni~Lao, Hongrae Lee, Da-Cheng Juan, et~al. 2022.
\newblock \href {{https://arxiv.org/pdf/2210.08726}} {Rarr: Researching and revising what language models say, using language models}.
\newblock \emph{arXiv preprint arXiv:2210.08726}.

\bibitem[{Gao et~al.(2025)Gao, Liu, Yue, Wu, Chen, Li, Tang, Wu, Chua, and Zhuang}]{gao2025benchmarking-svip}
Minghe Gao, Xuqi Liu, Zhongqi Yue, Yang Wu, Shuang Chen, Juncheng Li, Siliang Tang, Fei Wu, Tat-Seng Chua, and Yueting Zhuang. 2025.
\newblock \href {{https://arxiv.org/pdf/2504.06606}} {Benchmarking multimodal cot reward model stepwise by visual program}.
\newblock \emph{arXiv preprint arXiv:2504.06606}.

\bibitem[{Gehring et~al.(2024)Gehring, Zheng, Copet, Mella, Cohen, and Synnaeve}]{gehring2024rlef}
Jonas Gehring, Kunhao Zheng, Jade Copet, Vegard Mella, Taco Cohen, and Gabriel Synnaeve. 2024.
\newblock \href {{https://arxiv.org/pdf/2410.02089}} {{RLEF:} grounding code llms in execution feedback with reinforcement learning}.
\newblock \emph{arXiv preprint arXiv:2410.02089}.

\bibitem[{Glaese et~al.(2022)Glaese, McAleese, Trebacz, Aslanides, Firoiu, Ewalds, Rauh, Weidinger, Chadwick, Thacker, Campbell{-}Gillingham, Uesato, Huang, Comanescu, Yang, See, Dathathri, Greig, Chen, Fritz, Elias, Green, Mokr{\'{a}}, Fernando, Wu, Foley, Young, Gabriel, Isaac, Mellor, Hassabis, Kavukcuoglu, Hendricks, and Irving}]{glase2022sparrow}
Amelia Glaese, Nat McAleese, Maja Trebacz, John Aslanides, Vlad Firoiu, Timo Ewalds, Maribeth Rauh, Laura Weidinger, Martin~J. Chadwick, Phoebe Thacker, Lucy Campbell{-}Gillingham, Jonathan Uesato, Po{-}Sen Huang, Ramona Comanescu, Fan Yang, Abigail See, Sumanth Dathathri, Rory Greig, Charlie Chen, Doug Fritz, Jaume~Sanchez Elias, Richard Green, Sona Mokr{\'{a}}, Nicholas Fernando, Boxi Wu, Rachel Foley, Susannah Young, Iason Gabriel, William Isaac, John Mellor, Demis Hassabis, Koray Kavukcuoglu, Lisa~Anne Hendricks, and Geoffrey Irving. 2022.
\newblock \href {https://doi.org/10.48550/arXiv.2209.14375} {Improving alignment of dialogue agents via targeted human judgements}.
\newblock \emph{arXiv preprint arXiv:2209.14375}.

\bibitem[{Goldie et~al.(2025)Goldie, Mirhoseini, Zhou, Cai, and Manning}]{goldie2025swirl-tooluse}
Anna Goldie, Azalia Mirhoseini, Hao Zhou, Irene Cai, and Christopher~D Manning. 2025.
\newblock \href {{https://arxiv.org/pdf/2504.04736}} {Synthetic data generation \& multi-step rl for reasoning \& tool use}.
\newblock \emph{arXiv preprint arXiv:2504.04736}.

\bibitem[{Gou et~al.(2023)Gou, Shao, Gong, Shen, Yang, Duan, and Chen}]{gou2023critic-tool-critiquing}
Zhibin Gou, Zhihong Shao, Yeyun Gong, Yelong Shen, Yujiu Yang, Nan Duan, and Weizhu Chen. 2023.
\newblock \href {{https://arxiv.org/pdf/2305.11738}} {Critic: Large language models can self-correct with tool-interactive critiquing}.
\newblock \emph{arXiv preprint arXiv:2305.11738}.

\bibitem[{Guan et~al.(2025)Guan, Zhang, Liu, Shang, Sun, Zhu, Yang, and Yang}]{guan2025rstar-math}
Xinyu Guan, Li~Lyna Zhang, Yifei Liu, Ning Shang, Youran Sun, Yi~Zhu, Fan Yang, and Mao Yang. 2025.
\newblock \href {{https://arxiv.org/pdf/2501.04519}} {rstar-math: Small llms can master math reasoning with self-evolved deep thinking}.
\newblock \emph{arXiv preprint arXiv:2501.04519}.

\bibitem[{Gulcehre et~al.(2023)Gulcehre, Paine, Srinivasan, Konyushkova, Weerts, Sharma, Siddhant, Ahern, Wang, Gu, Macherey, Doucet, Firat, and de~Freitas}]{gulcehre2023rest}
Caglar Gulcehre, Tom~Le Paine, Srivatsan Srinivasan, Ksenia Konyushkova, Lotte Weerts, Abhishek Sharma, Aditya Siddhant, Alexa Ahern, Miaosen Wang, Chenjie Gu, Wolfgang Macherey, A.~Doucet, Orhan Firat, and Nando de~Freitas. 2023.
\newblock \href {{https://arxiv.org/pdf/2308.08998}} {Reinforced self-training (rest) for language modeling}.
\newblock \emph{arXiv preprint arXiv:2308.08998}.

\bibitem[{Guo et~al.(2025{\natexlab{a}})Guo, Chi, Dong, Dong, Wu, Huang, and Wei}]{guo2025reward-reasoning-model}
Jiaxin Guo, Zewen Chi, Li~Dong, Qingxiu Dong, Xun Wu, Shaohan Huang, and Furu Wei. 2025{\natexlab{a}}.
\newblock \href {{https://arxiv.org/pdf/2505.14674}} {Reward reasoning model}.
\newblock \emph{arXiv preprint arXiv:2505.14674}.

\bibitem[{Guo et~al.(2025{\natexlab{b}})Guo, Zhang, Chen, Ji, Wang, Hu, and Chen}]{guo2025vision-language-action}
Yanjiang Guo, Jianke Zhang, Xiaoyu Chen, Xiang Ji, Yen-Jen Wang, Yucheng Hu, and Jianyu Chen. 2025{\natexlab{b}}.
\newblock \href {{https://arxiv.org/pdf/2501.16664}} {Improving vision-language-action model with online reinforcement learning}.
\newblock \emph{arXiv preprint arXiv:2501.16664}.

\bibitem[{Guo et~al.(2025{\natexlab{c}})Guo, Zhang, Tong, Zhao, Gao, Li, and Heng}]{guo2025parm-image-generation}
Ziyu Guo, Renrui Zhang, Chengzhuo Tong, Zhizheng Zhao, Peng Gao, Hongsheng Li, and Pheng-Ann Heng. 2025{\natexlab{c}}.
\newblock \href {{https://arxiv.org/pdf/2501.13926}} {Can we generate images with cot? let's verify and reinforce image generation step by step}.
\newblock \emph{arXiv preprint arXiv:2501.13926}.

\bibitem[{Gureja et~al.(2024)Gureja, Miranda, Islam, Maheshwary, Sharma, Winata, Lambert, Ruder, Hooker, and Fadaee}]{gureja2024multilingual-rewardbench}
Srishti Gureja, Lester James~V. Miranda, Shayekh~Bin Islam, Rishabh Maheshwary, Drishti Sharma, Gusti Winata, Nathan Lambert, Sebastian Ruder, Sara Hooker, and Marzieh Fadaee. 2024.
\newblock \href {https://arxiv.org/pdf/2410.15522} {M-rewardbench: Evaluating reward models in multilingual settings}.
\newblock \emph{arXiv preprint arXiv:2410.15522}.

\bibitem[{Hao et~al.(2023)Hao, Gu, Ma, Hong, Wang, Wang, and Hu}]{hao2023rap}
Shibo Hao, Yi~Gu, Haodi Ma, Joshua~Jiahua Hong, Zhen Wang, Daisy~Zhe Wang, and Zhiting Hu. 2023.
\newblock \href {{https://arxiv.org/pdf/2305.14992}} {Reasoning with language model is planning with world model}.
\newblock \emph{arXiv preprint arXiv:2305.14992}.

\bibitem[{He et~al.(2024)He, Luo, Bai, Hu, Thai, Shen, Hu, Han, Huang, Zhang et~al.}]{he2024olympiadbench}
Chaoqun He, Renjie Luo, Yuzhuo Bai, Shengding Hu, Zhen~Leng Thai, Junhao Shen, Jinyi Hu, Xu~Han, Yujie Huang, Yuxiang Zhang, et~al. 2024.
\newblock \href {{https://arxiv.org/pdf/2402.14008}} {Olympiadbench: A challenging benchmark for promoting agi with olympiad-level bilingual multimodal scientific problems}.
\newblock \emph{arXiv preprint arXiv:2402.14008}.

\bibitem[{Hendrycks et~al.(2021)Hendrycks, Burns, Kadavath, Arora, Basart, Tang, Song, and Steinhardt}]{hendrycks2021math-dataset}
Dan Hendrycks, Collin Burns, Saurav Kadavath, Akul Arora, Steven Basart, Eric Tang, Dawn Song, and Jacob Steinhardt. 2021.
\newblock \href {{https://arxiv.org/pdf/2103.03874}} {Measuring mathematical problem solving with the math dataset}.
\newblock \emph{arXiv preprint arXiv:2103.03874}.

\bibitem[{Hosseini et~al.(2024)Hosseini, Yuan, Malkin, Courville, Sordoni, and Agarwal}]{hosseini2024vstar}
Arian Hosseini, Xingdi Yuan, Nikolay Malkin, Aaron Courville, Alessandro Sordoni, and Rishabh Agarwal. 2024.
\newblock \href {{https://arxiv.org/pdf/2402.06457}} {V-star: Training verifiers for self-taught reasoners}.
\newblock \emph{arXiv preprint arXiv:2402.06457}.

\bibitem[{Hu(2025)}]{hu2025reinforce++}
Jian Hu. 2025.
\newblock \href {{https://arxiv.org/pdf/2501.03262}} {Reinforce++: A simple and efficient approach for aligning large language models}.
\newblock \emph{arXiv preprint arXiv:2501.03262}.

\bibitem[{Huang et~al.(2025{\natexlab{a}})Huang, He, Zhou, Zhang, Liu, Wang, Su, Zheng, and Liu}]{huang2025think-judge}
Hui Huang, Yancheng He, Hongli Zhou, Rui Zhang, Wei Liu, Weixun Wang, Wenbo Su, Bo~Zheng, and Jiaheng Liu. 2025{\natexlab{a}}.
\newblock \href {{https://arxiv.org/pdf/2505.14268}} {Think-j: Learning to think for generative llm-as-a-judge}.
\newblock \emph{arXiv preprint arXiv:2505.14268}.

\bibitem[{Huang et~al.(2023)Huang, Chen, Mishra, Zheng, Yu, Song, and Zhou}]{huang2023cannot-self-correct}
Jie Huang, Xinyun Chen, Swaroop Mishra, Huaixiu~Steven Zheng, Adams~Wei Yu, Xinying Song, and Denny Zhou. 2023.
\newblock \href {{https://arxiv.org/pdf/2310.01798}} {Large language models cannot self-correct reasoning yet}.
\newblock \emph{arXiv preprint arXiv:2310.01798}.

\bibitem[{Huang et~al.(2025{\natexlab{b}})Huang, Jia, Zhai, Cao, Ye, Zhao, Hu, and Lin}]{huang2025vision-r1-reasoning}
Wenxuan Huang, Bohan Jia, Zijie Zhai, Shaosheng Cao, Zheyu Ye, Fei Zhao, Yao Hu, and Shaohui Lin. 2025{\natexlab{b}}.
\newblock \href {{https://arxiv.org/pdf/2503.06749}} {Vision-r1: Incentivizing reasoning capability in multimodal large language models}.
\newblock \emph{arXiv preprint arXiv:2503.06749}.

\bibitem[{Jenner and Gleave(2022)}]{jenner2022preprocessing-reward-interpretability}
Erik Jenner and Adam Gleave. 2022.
\newblock \href {{https://arxiv.org/pdf/2203.13553}} {Preprocessing reward functions for interpretability}.
\newblock \emph{arXiv preprint arXiv:2203.13553}.

\bibitem[{Ji et~al.(2023)Ji, Liu, Dai, Pan, Zhang, Bian, Chen, Sun, Wang, and Yang}]{ji2023beavertails}
Jiaming Ji, Mickel Liu, Josef Dai, Xuehai Pan, Chi Zhang, Ce~Bian, Boyuan Chen, Ruiyang Sun, Yizhou Wang, and Yaodong Yang. 2023.
\newblock \href {https://arxiv.org/pdf/2307.04657} {Beavertails: Towards improved safety alignment of llm via a human-preference dataset}.
\newblock \emph{Advances in Neural Information Processing Systems}, 36:24678--24704.

\bibitem[{Jiang et~al.(2024)Jiang, Chen, Min, Chen, Cheng, Wang, Tang, Sun, Deng, Zhao et~al.}]{jiang2024slow-thinking-still-1}
Jinhao Jiang, Zhipeng Chen, Yingqian Min, Jie Chen, Xiaoxue Cheng, Jiapeng Wang, Yiru Tang, Haoxiang Sun, Jia Deng, Wayne~Xin Zhao, et~al. 2024.
\newblock \href {{https://arxiv.org/pdf/2411.11694}} {Enhancing llm reasoning with reward-guided tree search}.
\newblock \emph{arXiv preprint arXiv:2411.11694}.

\bibitem[{Jiang et~al.(2025)Jiang, Lin, Cao, Tian, Kang, Wang, Sun, and Han}]{jiang2025deepretrieval}
Pengcheng Jiang, Jiacheng Lin, Lang Cao, Runchu Tian, SeongKu Kang, Zifeng Wang, Jimeng Sun, and Jiawei Han. 2025.
\newblock \href {https://arxiv.org/pdf/2503.00223} {Deepretrieval: Hacking real search engines and retrievers with large language models via reinforcement learning}.
\newblock \emph{arXiv preprint arXiv:2503.00223}.

\bibitem[{Jiang et~al.(2023)Jiang, Wang, and Wang}]{jiang2023self-evolve}
Shuyang Jiang, Yuhao Wang, and Yu~Wang. 2023.
\newblock \href {{https://arxiv.org/pdf/2306.02907}} {Selfevolve: A code evolution framework via large language models}.
\newblock \emph{arXiv preprint arXiv:2306.02907}.

\bibitem[{Jiao et~al.(2024{\natexlab{a}})Jiao, Guo, Zhang, Chen, Joty, and Wei}]{jiao2024pseudo-feedback-PFPO}
Fangkai Jiao, Geyang Guo, Xingxing Zhang, Nancy~F Chen, Shafiq Joty, and Furu Wei. 2024{\natexlab{a}}.
\newblock \href {{https://arxiv.org/pdf/2411.16345}} {Preference optimization for reasoning with pseudo feedback}.
\newblock \emph{arXiv preprint arXiv:2411.16345}.

\bibitem[{Jiao et~al.(2024{\natexlab{b}})Jiao, Qin, Liu, Chen, and Joty}]{jiao2024planning-trajectories-collection}
Fangkai Jiao, Chengwei Qin, Zhengyuan Liu, Nancy~F Chen, and Shafiq Joty. 2024{\natexlab{b}}.
\newblock \href {{https://arxiv.org/pdf/2402.00658}} {Learning planning-based reasoning by trajectories collection and process reward synthesizing}.
\newblock \emph{arXiv preprint arXiv:2402.00658}.

\bibitem[{Jin et~al.(2025)Jin, Zeng, Yue, Wang, Zamani, and Han}]{jin2025search-r1}
Bowen Jin, Hansi Zeng, Zhenrui Yue, Dong Wang, Hamed Zamani, and Jiawei Han. 2025.
\newblock \href {{https://arxiv.org/pdf/2503.09516}} {Search-r1: Training llms to reason and leverage search engines with reinforcement learning}.
\newblock \emph{arXiv preprint arXiv:2503.09516}.

\bibitem[{Jin et~al.(2024)Jin, Yuan, Men, Cao, Chen, Liu, and Zhao}]{jin2024ragrewardbench}
Zhuoran Jin, Hongbang Yuan, Tianyi Men, Pengfei Cao, Yubo Chen, Kang Liu, and Jun Zhao. 2024.
\newblock \href {https://arxiv.org/pdf/2412.13746} {Rag-rewardbench: Benchmarking reward models in retrieval augmented generation for preference alignment}.
\newblock \emph{arXiv preprint arXiv:2412.13746}.

\bibitem[{Jinnai et~al.(2024)Jinnai, Morimura, Ariu, and Abe}]{jinnai2024regularized-best-of-n}
Yuu Jinnai, Tetsuro Morimura, Kaito Ariu, and Kenshi Abe. 2024.
\newblock \href {{https://arxiv.org/pdf/2404.01054}} {Regularized best-of-n sampling with minimum bayes risk objective for language model alignment}.
\newblock \emph{arXiv preprint arXiv:2404.01054}.

\bibitem[{Kamoi et~al.(2024)Kamoi, Zhang, Zhang, Han, and Zhang}]{kamoi2024can-correct}
Ryo Kamoi, Yusen Zhang, Nan Zhang, Jiawei Han, and Rui Zhang. 2024.
\newblock \href {https://arxiv.org/pdf/2406.01297} {When can llms actually correct their own mistakes? a critical survey of self-correction of llms}.
\newblock \emph{Transactions of the Association for Computational Linguistics}, 12:1417--1440.

\bibitem[{Kaplan et~al.(2020)Kaplan, McCandlish, Henighan, Brown, Chess, Child, Gray, Radford, Wu, and Amodei}]{kaplan2020scaling}
Jared Kaplan, Sam McCandlish, Tom Henighan, Tom~B Brown, Benjamin Chess, Rewon Child, Scott Gray, Alec Radford, Jeffrey Wu, and Dario Amodei. 2020.
\newblock \href {{https://arxiv.org/pdf/2001.08361}} {Scaling laws for neural language models}.
\newblock \emph{arXiv preprint arXiv:2001.08361}.

\bibitem[{Khalifa et~al.(2025)Khalifa, Agarwal, Logeswaran, Kim, Peng, Lee, Lee, and Wang}]{khalifa2025thinkprm}
Muhammad Khalifa, Rishabh Agarwal, Lajanugen Logeswaran, Jaekyeom Kim, Hao Peng, Moontae Lee, Honglak Lee, and Lu~Wang. 2025.
\newblock \href {{https://arxiv.org/pdf/2504.16828}} {Process reward models that think}.
\newblock \emph{arXiv preprint arXiv:2504.16828}.

\bibitem[{Khalifa et~al.(2023)Khalifa, Logeswaran, Lee, Lee, and Wang}]{khalifa2023grace}
Muhammad Khalifa, Lajanugen Logeswaran, Moontae Lee, Honglak Lee, and Lu~Wang. 2023.
\newblock \href {{https://arxiv.org/pdf/2305.14934}} {Grace: Discriminator-guided chain-of-thought reasoning}.
\newblock \emph{arXiv preprint arXiv:2305.14934}.

\bibitem[{Khanov et~al.(2024)Khanov, Burapacheep, and Li}]{khanov2024args}
Maxim Khanov, Jirayu Burapacheep, and Yixuan Li. 2024.
\newblock \href {https://arxiv.org/pdf/2402.01694} {Args: Alignment as reward-guided search}.
\newblock In \emph{The Twelfth International Conference on Learning Representations}.

\bibitem[{Kim et~al.(2023)Kim, Baldi, and McAleer}]{kim2023rci-computer-tasks}
Geunwoo Kim, Pierre Baldi, and Stephen McAleer. 2023.
\newblock \href {https://arxiv.org/pdf/2303.17491} {Language models can solve computer tasks}.
\newblock \emph{Advances in Neural Information Processing Systems}, 36:39648--39677.

\bibitem[{Kumar et~al.(2024)Kumar, Zhuang, Agarwal, Su, Co-Reyes, Singh, Baumli, Iqbal, Bishop, Roelofs et~al.}]{kumar2024score-self-correct}
Aviral Kumar, Vincent Zhuang, Rishabh Agarwal, Yi~Su, John~D Co-Reyes, Avi Singh, Kate Baumli, Shariq Iqbal, Colton Bishop, Rebecca Roelofs, et~al. 2024.
\newblock \href {{https://arxiv.org/pdf/2409.12917}} {Training language models to self-correct via reinforcement learning}.
\newblock \emph{arXiv preprint arXiv:2409.12917}.

\bibitem[{Kwon et~al.(2023)Kwon, Xie, Bullard, and Sadigh}]{kwon2023reward-design-language-models}
Minae Kwon, Sang~Michael Xie, Kalesha Bullard, and Dorsa Sadigh. 2023.
\newblock \href {{https://arxiv.org/pdf/2303.00001}} {Reward design with language models}.
\newblock \emph{arXiv preprint arXiv:2303.00001}.

\bibitem[{Lai et~al.(2024)Lai, Tian, Chen, Yang, Peng, and Jia}]{lai2024stepdpo}
Xin Lai, Zhuotao Tian, Yukang Chen, Senqiao Yang, Xiangru Peng, and Jiaya Jia. 2024.
\newblock \href {{https://arxiv.org/pdf/2406.18629}} {{Step-DPO}: Step-wise preference optimization for long-chain reasoning of llms}.
\newblock \emph{arXiv preprint arXiv:2406.18629}.

\bibitem[{Lai et~al.(2025)Lai, Zhong, Li, Zhao, and Yang}]{lai2025med-r1}
Yuxiang Lai, Jike Zhong, Ming Li, Shitian Zhao, and Xiaofeng Yang. 2025.
\newblock \href {{https://arxiv.org/pdf/2503.13939}} {Med-r1: Reinforcement learning for generalizable medical reasoning in vision-language models}.
\newblock \emph{arXiv preprint arXiv:2503.13939}.

\bibitem[{Lambert et~al.(2024)Lambert, Pyatkin, Morrison, Miranda, Lin, Chandu, Dziri, Kumar, Zick, Choi et~al.}]{lambert2024rewardbench}
Nathan Lambert, Valentina Pyatkin, Jacob Morrison, LJ~Miranda, Bill~Yuchen Lin, Khyathi Chandu, Nouha Dziri, Sachin Kumar, Tom Zick, Yejin Choi, et~al. 2024.
\newblock \href {{https://arxiv.org/pdf/2403.13787}} {Rewardbench: Evaluating reward models for language modeling}.
\newblock \emph{arXiv preprint arXiv:2403.13787}.

\bibitem[{Le et~al.(2022)Le, Wang, Gotmare, Savarese, and Hoi}]{le2022coderl}
Hung Le, Yue Wang, Akhilesh~Deepak Gotmare, Silvio Savarese, and Steven Chu~Hong Hoi. 2022.
\newblock \href {https://arxiv.org/pdf/2207.01780} {Coderl: Mastering code generation through pretrained models and deep reinforcement learning}.
\newblock \emph{Advances in Neural Information Processing Systems}, 35:21314--21328.

\bibitem[{Lee et~al.(2024)Lee, Park, Lee, and Lim}]{lee2024ask}
Dongyub Lee, Eunhwan Park, Hodong Lee, and Heui-Seok Lim. 2024.
\newblock \href {https://aclanthology.org/2024.eacl-long.149/} {Ask, assess, and refine: Rectifying factual consistency and hallucination in llms with metric-guided feedback learning}.
\newblock In \emph{Proceedings of the 18th Conference of the European Chapter of the Association for Computational Linguistics (Volume 1: Long Papers)}, pages 2422--2433.

\bibitem[{Lee et~al.(2023{\natexlab{a}})Lee, Phatale, Mansoor, Mesnard, Ferret, Lu, Bishop, Hall, Carbune, Rastogi, and Prakash}]{lee2023rlaif-rlhf}
Harrison Lee, Samrat Phatale, Hassan Mansoor, Thomas Mesnard, Johan Ferret, Kellie Lu, Colton Bishop, Ethan Hall, Victor Carbune, Abhinav Rastogi, and Sushant Prakash. 2023{\natexlab{a}}.
\newblock \href {{https://arxiv.org/pdf/2309.00267}} {{RLAIF} vs. {RLHF}: Scaling reinforcement learning from human feedback with {AI} feedback}.
\newblock \emph{arXiv preprint arXiv:2309.00267}.

\bibitem[{Lee et~al.(2023{\natexlab{b}})Lee, Liu, Ryu, Watkins, Du, Boutilier, Abbeel, Ghavamzadeh, and Gu}]{lee2023aligning}
Kimin Lee, Hao Liu, Moonkyung Ryu, Olivia Watkins, Yuqing Du, Craig Boutilier, Pieter Abbeel, Mohammad Ghavamzadeh, and Shixiang~Shane Gu. 2023{\natexlab{b}}.
\newblock \href {{https://arxiv.org/pdf/2302.12192}} {Aligning text-to-image models using human feedback}.
\newblock \emph{arXiv preprint arXiv:2302.12192}.

\bibitem[{Li et~al.(2024{\natexlab{a}})Li, Wang, Grama, and Zhang}]{li2024cards}
Bolian Li, Yifan Wang, Ananth Grama, and Ruqi Zhang. 2024{\natexlab{a}}.
\newblock \href {{https://arxiv.org/pdf/2406.16306}} {Cascade reward sampling for efficient decoding-time alignment}.
\newblock \emph{arXiv preprint arXiv:2406.16306}.

\bibitem[{Li et~al.(2025{\natexlab{a}})Li, Zhou, Lu, Tyen, Gui, Aloisi, and He}]{li2025dars-verbal-reflection}
Jiazheng Li, Yuxiang Zhou, Junru Lu, Gladys Tyen, Lin Gui, Cesare Aloisi, and Yulan He. 2025{\natexlab{a}}.
\newblock \href {{https://arxiv.org/pdf/2502.19230}} {Two heads are better than one: Dual-model verbal reflection at inference-time}.
\newblock \emph{arXiv preprint arXiv:2502.19230}.

\bibitem[{Li et~al.(2023{\natexlab{a}})Li, Sun, Yuan, Fan, Zhao, and Liu}]{li2023generative-judge}
Junlong Li, Shichao Sun, Weizhe Yuan, Run-Ze Fan, Hai Zhao, and Pengfei Liu. 2023{\natexlab{a}}.
\newblock \href {{https://arxiv.org/pdf/2310.05470}} {Generative judge for evaluating alignment}.
\newblock \emph{arXiv preprint arXiv:2310.05470}.

\bibitem[{Li et~al.(2024{\natexlab{b}})Li, Wei, Xie, Yang, Song, Wang, An, Liu, Li, Lin et~al.}]{li2024vlrewardbench}
Lei Li, Yuancheng Wei, Zhihui Xie, Xuqing Yang, Yifan Song, Peiyi Wang, Chenxin An, Tianyu Liu, Sujian Li, Bill~Yuchen Lin, et~al. 2024{\natexlab{b}}.
\newblock \href {{https://arxiv.org/pdf/2411.17451}} {Vlrewardbench: A challenging benchmark for vision-language generative reward models}.
\newblock \emph{arXiv preprint arXiv:2411.17451}.

\bibitem[{Li et~al.(2024{\natexlab{c}})Li, Chen, Zhang, Wu, Li, Guan, Yu, and Yuan}]{li2024continual-multi-objective-rl}
Lihe Li, Ruotong Chen, Ziqian Zhang, Zhichao Wu, Yi-Chen Li, Cong Guan, Yang Yu, and Lei Yuan. 2024{\natexlab{c}}.
\newblock \href {https://www.ijcai.org/proceedings/2024/490} {Continual multi-objective reinforcement learning via reward model rehearsal}.
\newblock In \emph{Proceedings of the Thirty-Third International Joint Conference on Artificial Intelligence}, pages 4434--4442.

\bibitem[{Li et~al.(2025{\natexlab{b}})Li, Chen, Li, and Chen}]{li2025relation-r1}
Lin Li, Wei Chen, Jiahui Li, and Long Chen. 2025{\natexlab{b}}.
\newblock \href {{https://arxiv.org/pdf/2504.14642}} {Relation-r1: Cognitive chain-of-thought guided reinforcement learning for unified relational comprehension}.
\newblock \emph{arXiv preprint arXiv:2504.14642}.

\bibitem[{Li et~al.(2025{\natexlab{c}})Li, Zhao, Zhong, Lai, and Zhang}]{li2025clsrl}
Ming Li, Shitian Zhao, Jike Zhong, Yuxiang Lai, and Kaipeng Zhang. 2025{\natexlab{c}}.
\newblock \href {{https://arxiv.org/pdf/2503.16188}} {Cls-rl: Image classification with rule-based reinforcement learning}.
\newblock \emph{arXiv preprint arXiv:2503.16188}.

\bibitem[{Li et~al.(2025{\natexlab{d}})Li, Zhang, Zhao, Zhang, Li, Zhang, and Zhang}]{li2025q-insight-image-quality}
Weiqi Li, Xuanyu Zhang, Shijie Zhao, Yabin Zhang, Junlin Li, Li~Zhang, and Jian Zhang. 2025{\natexlab{d}}.
\newblock \href {{https://arxiv.org/pdf/2503.22679}} {Q-insight: Understanding image quality via visual reinforcement learning}.
\newblock \emph{arXiv preprint arXiv:2503.22679}.

\bibitem[{Li and Li(2024)}]{li2024pqm}
Wendi Li and Yixuan Li. 2024.
\newblock \href {{https://arxiv.org/pdf/2410.11287}} {Process reward model with q-value rankings}.
\newblock \emph{arXiv preprint arXiv:2410.11287}.

\bibitem[{Li et~al.(2025{\natexlab{e}})Li, Jin, Dong, Qian, Zhu, Wu, Wen, and Dou}]{li2025webthinker}
Xiaoxi Li, Jiajie Jin, Guanting Dong, Hongjin Qian, Yutao Zhu, Yongkang Wu, Ji-Rong Wen, and Zhicheng Dou. 2025{\natexlab{e}}.
\newblock \href {https://arxiv.org/pdf/2504.21776} {Webthinker: Empowering large reasoning models with deep research capability}.

\bibitem[{Li et~al.(2025{\natexlab{f}})Li, Yan, Meng, Dong, Zeng, He, Wang, Qiao, Wang, and Wang}]{li2025videochat-r1}
Xinhao Li, Ziang Yan, Desen Meng, Lu~Dong, Xiangyu Zeng, Yinan He, Yali Wang, Yu~Qiao, Yi~Wang, and Limin Wang. 2025{\natexlab{f}}.
\newblock \href {{https://arxiv.org/pdf/2504.06958}} {Videochat-r1: Enhancing spatio-temporal perception via reinforcement fine-tuning}.
\newblock \emph{arXiv preprint arXiv:2504.06958}.

\bibitem[{Li et~al.(2025{\natexlab{g}})Li, Zou, and Liu}]{li2025torl}
Xuefeng Li, Haoyang Zou, and Pengfei Liu. 2025{\natexlab{g}}.
\newblock \href {{https://arxiv.org/pdf/2503.23383}} {Torl: Scaling tool-integrated rl}.
\newblock \emph{arXiv preprint arXiv:2503.23383}.

\bibitem[{Li et~al.(2023{\natexlab{b}})Li, Lin, Zhang, Fu, Chen, Lou, and Chen}]{li2023diverse-stepaware-verifier}
Yifei Li, Zeqi Lin, Shizhuo Zhang, Qiang Fu, Bei Chen, Jian-Guang Lou, and Weizhu Chen. 2023{\natexlab{b}}.
\newblock \href {https://aclanthology.org/2023.acl-long.291/} {Making language models better reasoners with step-aware verifier}.
\newblock In \emph{Proceedings of the 61st Annual Meeting of the Association for Computational Linguistics (Volume 1: Long Papers)}, pages 5315--5333.

\bibitem[{Liang et~al.(2023)Liang, He, Jiao, Wang, Wang, Wang, Yang, Shi, and Tu}]{liang2023divergent-thinking-multiagent-debate}
Tian Liang, Zhiwei He, Wenxiang Jiao, Xing Wang, Yan Wang, Rui Wang, Yujiu Yang, Shuming Shi, and Zhaopeng Tu. 2023.
\newblock \href {{https://arxiv.org/pdf/2305.19118}} {Encouraging divergent thinking in large language models through multi-agent debate}.
\newblock \emph{arXiv preprint arXiv:2305.19118}.

\bibitem[{Liang et~al.(2024)Liang, He, Li, Li, Klimovskiy, Carolan, Sun, Pont-Tuset, Young, Yang et~al.}]{liang2024rich-texttoimage}
Youwei Liang, Junfeng He, Gang Li, Peizhao Li, Arseniy Klimovskiy, Nicholas Carolan, Jiao Sun, Jordi Pont-Tuset, Sarah Young, Feng Yang, et~al. 2024.
\newblock \href {https://arxiv.org/pdf/2312.10240} {Rich human feedback for text-to-image generation}.
\newblock In \emph{Proceedings of the IEEE/CVF Conference on Computer Vision and Pattern Recognition}, pages 19401--19411.

\bibitem[{Liao et~al.(2025)Liao, Xu, Dong, Li, Monz, Savarese, Sahoo, and Xiong}]{liao2025rewardspeculative}
Baohao Liao, Yuhui Xu, Hanze Dong, Junnan Li, Christof Monz, Silvio Savarese, Doyen Sahoo, and Caiming Xiong. 2025.
\newblock \href {{https://arxiv.org/pdf/2501.19324}} {Reward-guided speculative decoding for efficient llm reasoning}.
\newblock \emph{arXiv preprint arXiv:2501.19324}.

\bibitem[{Lightman et~al.(2023)Lightman, Kosaraju, Burda, Edwards, Baker, Lee, Leike, Schulman, Sutskever, and Cobbe}]{lightman2023letsverify}
Hunter Lightman, Vineet Kosaraju, Yuri Burda, Harrison Edwards, Bowen Baker, Teddy Lee, Jan Leike, John Schulman, Ilya Sutskever, and Karl Cobbe. 2023.
\newblock \href {https://arxiv.org/pdf/2305.20050} {Let's verify step by step}.
\newblock In \emph{The Twelfth International Conference on Learning Representations}.

\bibitem[{Lin et~al.(2025)Lin, Wang, and Qian}]{lin2025rec-r1}
Jiacheng Lin, Tian Wang, and Kun Qian. 2025.
\newblock \href {{https://arxiv.org/pdf/2503.24289}} {Rec-r1: Bridging generative large language models and user-centric recommendation systems via reinforcement learning}.
\newblock \emph{arXiv preprint arXiv:2503.24289}.

\bibitem[{Lin et~al.(2024{\natexlab{a}})Lin, Gao, Oguz, Xiong, Lin, Yih, and Chen}]{lin2024flame}
Sheng-Chieh Lin, Luyu Gao, Barlas Oguz, Wenhan Xiong, Jimmy Lin, Scott Yih, and Xilun Chen. 2024{\natexlab{a}}.
\newblock \href {https://arxiv.org/pdf/2405.01525} {Flame: Factuality-aware alignment for large language models}.
\newblock \emph{Advances in Neural Information Processing Systems}, 37:115588--115614.

\bibitem[{Lin et~al.(2024{\natexlab{b}})Lin, Gou, Liang, Luo, Liu, and Yang}]{lin2024criticbench}
Zicheng Lin, Zhibin Gou, Tian Liang, Ruilin Luo, Haowei Liu, and Yujiu Yang. 2024{\natexlab{b}}.
\newblock \href {{https://arxiv.org/pdf/2402.14809}} {Criticbench: Benchmarking llms for critique-correct reasoning}.
\newblock \emph{arXiv preprint arXiv:2402.14809}.

\bibitem[{Liu et~al.(2024{\natexlab{a}})Liu, Zeng, Liu, Yan, He, Wang, Yan, Liu, and Zhou}]{liu2024skywork}
Chris~Yuhao Liu, Liang Zeng, Jiacai Liu, Rui Yan, Jujie He, Chaojie Wang, Shuicheng Yan, Yang Liu, and Yahui Zhou. 2024{\natexlab{a}}.
\newblock \href {{https://arxiv.org/pdf/2410.18451}} {Skywork-reward: Bag of tricks for reward modeling in llms}.
\newblock \emph{arXiv preprint arXiv:2410.18451}.

\bibitem[{Liu et~al.(2025{\natexlab{a}})Liu, Wang, Cai, Zhang, Zhan, and Duan}]{liu2025video-t1}
Fangfu Liu, Hanyang Wang, Yimo Cai, Kaiyan Zhang, Xiaohang Zhan, and Yueqi Duan. 2025{\natexlab{a}}.
\newblock \href {{https://arxiv.org/pdf/2503.18942}} {Video-t1: Test-time scaling for video generation}.
\newblock \emph{arXiv preprint arXiv:2503.18942}.

\bibitem[{Liu et~al.(2023)Liu, Zhu, Xiao, Fu, Han, Yang, and Ye}]{liu2023rltf-unittest}
Jiate Liu, Yiqin Zhu, Kaiwen Xiao, Qiang Fu, Xiao Han, Wei Yang, and Deheng Ye. 2023.
\newblock \href {https://openreview.net/forum?id=hjYmsV6nXZ} {{RLTF:} reinforcement learning from unit test feedback}.
\newblock \emph{Trans. Mach. Learn. Res.}, 2023.

\bibitem[{Liu et~al.(2025{\natexlab{b}})Liu, Xiong, Ren, Chen, Wu, Joshi, Gao, Shen, Qin, Yu, Sohn, Makarova, Liu, Liu, Piot, Ittycheriah, Kumar, and Saleh}]{liu2025robust-reward-model}
Tianqi Liu, Wei Xiong, Jie Ren, Lichang Chen, Junru Wu, Rishabh Joshi, Yang Gao, Jiaming Shen, Zhen Qin, Tianhe Yu, Daniel Sohn, Anastasiia Makarova, Jeremiah Liu, Yuan Liu, Bilal Piot, Abe Ittycheriah, Aviral Kumar, and Mohammad Saleh. 2025{\natexlab{b}}.
\newblock \href {https://arxiv.org/pdf/2409.13156} {Rrm: Robust reward model training mitigates reward hacking}.
\newblock \emph{arXiv preprint arXiv:2409.13156}.

\bibitem[{Liu et~al.(2024{\natexlab{b}})Liu, Zhao, Joshi, Khalman, Saleh, Liu, and Liu}]{liu2024rso}
Tianqi Liu, Yao Zhao, Rishabh Joshi, Misha Khalman, Mohammad Saleh, Peter~J Liu, and Jialu Liu. 2024{\natexlab{b}}.
\newblock \href {https://openreview.net/forum?id=xbjSwwrQOe} {Statistical rejection sampling improves preference optimization}.
\newblock In \emph{The Twelfth International Conference on Learning Representations}.

\bibitem[{Liu et~al.(2024{\natexlab{c}})Liu, Li, Zhang, Zhou, Cheng, and He}]{liu2024diving-multimodal}
Wei Liu, Junlong Li, Xiwen Zhang, Fan Zhou, Yu~Cheng, and Junxian He. 2024{\natexlab{c}}.
\newblock \href {{https://arxiv.org/pdf/2412.17451}} {Diving into self-evolving training for multimodal reasoning}.
\newblock \emph{arXiv preprint arXiv:2412.17451}.

\bibitem[{Liu et~al.(2024{\natexlab{d}})Liu, Yao, Min, Cao, Hou, and Li}]{liu2024rmbench}
Yantao Liu, Zijun Yao, Rui Min, Yixin Cao, Lei Hou, and Juanzi Li. 2024{\natexlab{d}}.
\newblock \href {{https://arxiv.org/pdf/2410.16184}} {Rm-bench: Benchmarking reward models of language models with subtlety and style}.
\newblock \emph{arXiv preprint arXiv:2410.16184}.

\bibitem[{Liu et~al.(2025{\natexlab{c}})Liu, Li, Xie, Hu, Han, Zhang, Yang, and Wu}]{liu2025infigui}
Yuhang Liu, Pengxiang Li, Congkai Xie, Xavier Hu, Xiaotian Han, Shengyu Zhang, Hongxia Yang, and Fei Wu. 2025{\natexlab{c}}.
\newblock \href {https://arxiv.org/pdf/2504.14239} {Infigui-r1: Advancing multimodal gui agents from reactive actors to deliberative reasoners}.
\newblock \emph{arXiv preprint arXiv:2504.14239}.

\bibitem[{Liu et~al.(2025{\natexlab{d}})Liu, Peng, Zhong, Yue, Lu, Yu, and Jia}]{liu2025seg-zero}
Yuqi Liu, Bohao Peng, Zhisheng Zhong, Zihao Yue, Fanbin Lu, Bei Yu, and Jiaya Jia. 2025{\natexlab{d}}.
\newblock \href {{https://arxiv.org/pdf/2503.06520}} {Seg-zero: Reasoning-chain guided segmentation via cognitive reinforcement}.
\newblock \emph{arXiv preprint arXiv:2503.06520}.

\bibitem[{Liu et~al.(2025{\natexlab{e}})Liu, Guo, Lou, Zeng, Niu, Wang, Xu, Cai, Yang, Zhao et~al.}]{liu2025fin-r1}
Zhaowei Liu, Xin Guo, Fangqi Lou, Lingfeng Zeng, Jinyi Niu, Zixuan Wang, Jiajie Xu, Weige Cai, Ziwei Yang, Xueqian Zhao, et~al. 2025{\natexlab{e}}.
\newblock \href {{https://arxiv.org/pdf/2503.16252}} {Fin-r1: A large language model for financial reasoning through reinforcement learning}.
\newblock \emph{arXiv preprint arXiv:2503.16252}.

\bibitem[{Liu et~al.(2025{\natexlab{f}})Liu, Zhang, Liu, Zhang, Sun, and Wang}]{liu2025othink-mr1}
Zhiyuan Liu, Yuting Zhang, Feng Liu, Changwang Zhang, Ying Sun, and Jun Wang. 2025{\natexlab{f}}.
\newblock \href {{https://arxiv.org/pdf/2503.16081}} {Othink-mr1: Stimulating multimodal generalized reasoning capabilities through dynamic reinforcement learning}.
\newblock \emph{arXiv preprint arXiv:2503.16081}.

\bibitem[{Liu et~al.(2024{\natexlab{e}})Liu, Chen, Shoeybi, Catanzaro, and Ping}]{liu2024acemath}
Zihan Liu, Yang Chen, Mohammad Shoeybi, Bryan Catanzaro, and Wei Ping. 2024{\natexlab{e}}.
\newblock \href {{https://arxiv.org/pdf/2412.15084}} {Acemath: Advancing frontier math reasoning with post-training and reward modeling}.
\newblock \emph{arXiv preprint arXiv:2412.15084}.

\bibitem[{Liu et~al.(2025{\natexlab{g}})Liu, Wang, Xu, Ma, Ruan, Li, Liu, and Wu}]{liu2025deepseek-grm}
Zijun Liu, Peiyi Wang, Runxin Xu, Shirong Ma, Chong Ruan, Peng Li, Yang Liu, and Yu~Wu. 2025{\natexlab{g}}.
\newblock \href {{https://arxiv.org/pdf/2504.02495}} {Inference-time scaling for generalist reward modeling}.
\newblock \emph{arXiv preprint arXiv:2504.02495}.

\bibitem[{Liu et~al.(2025{\natexlab{h}})Liu, Sun, Zang, Dong, Cao, Duan, Lin, and Wang}]{liu2025visualrft}
Ziyu Liu, Zeyi Sun, Yuhang Zang, Xiaoyi Dong, Yuhang Cao, Haodong Duan, Dahua Lin, and Jiaqi Wang. 2025{\natexlab{h}}.
\newblock \href {{https://arxiv.org/pdf/2503.01785}} {Visual-rft: Visual reinforcement fine-tuning}.
\newblock \emph{arXiv preprint arXiv:2503.01785}.

\bibitem[{Lu et~al.(2025)Lu, Chai, Guo, Yin, Liu, Wang, Xiong, and Li}]{lu2025ui-r1}
Zhengxi Lu, Yuxiang Chai, Yaxuan Guo, Xi~Yin, Liang Liu, Hao Wang, Guanjing Xiong, and Hongsheng Li. 2025.
\newblock \href {{https://arxiv.org/pdf/2503.21620}} {Ui-r1: Enhancing action prediction of gui agents by reinforcement learning}.
\newblock \emph{arXiv preprint arXiv:2503.21620}.

\bibitem[{Luo et~al.(2023)Luo, Sun, Xu, Zhao, Lou, Tao, Geng, Lin, Chen, and Zhang}]{luo2023wizardmath}
Haipeng Luo, Qingfeng Sun, Can Xu, Pu~Zhao, Jianguang Lou, Chongyang Tao, Xiubo Geng, Qingwei Lin, Shifeng Chen, and Dongmei Zhang. 2023.
\newblock \href {https://doi.org/10.48550/arXiv.2308.09583} {Wizardmath: Empowering mathematical reasoning for large language models via reinforced evol-instruct}.
\newblock \emph{arXiv preprint arXiv:2308.09583}.

\bibitem[{Luo et~al.(2025)Luo, Guo, Lin, Wu, Mu, Liu, Song, Zhu, Tuan et~al.}]{luo2025kbqa-o1}
Haoran Luo, Yikai Guo, Qika Lin, Xiaobao Wu, Xinyu Mu, Wenhao Liu, Meina Song, Yifan Zhu, Luu~Anh Tuan, et~al. 2025.
\newblock \href {{https://arxiv.org/pdf/2501.18922}} {Kbqa-o1: Agentic knowledge base question answering with monte carlo tree search}.
\newblock \emph{arXiv preprint arXiv:2501.18922}.

\bibitem[{Luo et~al.(2024)Luo, Liu, Liu, Phatale, Lara, Li, Shu, Zhu, Meng, Sun et~al.}]{luo2024omegareward}
Liangchen Luo, Yinxiao Liu, Rosanne Liu, Samrat Phatale, Harsh Lara, Yunxuan Li, Lei Shu, Yun Zhu, Lei Meng, Jiao Sun, et~al. 2024.
\newblock \href {{https://arxiv.org/pdf/2406.06592}} {Improve mathematical reasoning in language models by automated process supervision}.
\newblock \emph{arXiv preprint arXiv:2406.06592}, 2.

\bibitem[{Lyu et~al.(2025)Lyu, Gao, Gu, Zhang, Gao, Liu, Wang, Li, Zhao, Huang et~al.}]{lyu2025oreal}
Chengqi Lyu, Songyang Gao, Yuzhe Gu, Wenwei Zhang, Jianfei Gao, Kuikun Liu, Ziyi Wang, Shuaibin Li, Qian Zhao, Haian Huang, et~al. 2025.
\newblock \href {{https://arxiv.org/pdf/2502.06781}} {Exploring the limit of outcome reward for learning mathematical reasoning}.
\newblock \emph{arXiv preprint arXiv:2502.06781}.

\bibitem[{Lyu et~al.(2023)Lyu, Havaldar, Stein, Zhang, Rao, Wong, Apidianaki, and Callison-Burch}]{lyu2023faithful-cot}
Qing Lyu, Shreya Havaldar, Adam Stein, Li~Zhang, Delip Rao, Eric Wong, Marianna Apidianaki, and Chris Callison-Burch. 2023.
\newblock \href {https://arxiv.org/pdf/2301.13379} {Faithful chain-of-thought reasoning}.
\newblock In \emph{The 13th International Joint Conference on Natural Language Processing and the 3rd Conference of the Asia-Pacific Chapter of the Association for Computational Linguistics (IJCNLP-AACL 2023)}.

\bibitem[{Ma et~al.(2025)Ma, Zhuang, Xu, Jiang, Chen, and Guo}]{ma2025sql-r1}
Peixian Ma, Xialie Zhuang, Chengjin Xu, Xuhui Jiang, Ran Chen, and Jian Guo. 2025.
\newblock \href {{https://arxiv.org/pdf/2504.08600}} {Sql-r1: Training natural language to sql reasoning model by reinforcement learning}.
\newblock \emph{arXiv preprint arXiv:2504.08600}.

\bibitem[{Madaan et~al.(2023)Madaan, Tandon, Gupta, Hallinan, Gao, Wiegreffe, Alon, Dziri, Prabhumoye, Yang et~al.}]{madaan2023self-refine}
Aman Madaan, Niket Tandon, Prakhar Gupta, Skyler Hallinan, Luyu Gao, Sarah Wiegreffe, Uri Alon, Nouha Dziri, Shrimai Prabhumoye, Yiming Yang, et~al. 2023.
\newblock \href {https://arxiv.org/pdf/2303.17651} {Self-refine: Iterative refinement with self-feedback}.
\newblock \emph{Advances in Neural Information Processing Systems}, 36:46534--46594.

\bibitem[{Mahan et~al.(2024)Mahan, Van~Phung, Rafailov, Blagden, Lile, Castricato, Fr{\"a}nken, Finn, and Albalak}]{mahan2024genrm}
Dakota Mahan, Duy Van~Phung, Rafael Rafailov, Chase Blagden, Nathan Lile, Louis Castricato, Jan-Philipp Fr{\"a}nken, Chelsea Finn, and Alon Albalak. 2024.
\newblock \href {{https://arxiv.org/pdf/2410.12832}} {Generative reward models}.
\newblock \emph{arXiv preprint arXiv:2410.12832}.

\bibitem[{McAleese et~al.(2024)McAleese, Pokorny, Uribe, Nitishinskaya, Trebacz, and Leike}]{mcaleese2024criticgpt-catch-bugs}
Nat McAleese, Rai~Michael Pokorny, Juan Felipe~Ceron Uribe, Evgenia Nitishinskaya, Maja Trebacz, and Jan Leike. 2024.
\newblock \href {{https://arxiv.org/pdf/2407.00215}} {Llm critics help catch llm bugs}.
\newblock \emph{arXiv preprint arXiv:2407.00215}.

\bibitem[{Meng et~al.(2025)Meng, Du, Liu, Zhou, Lu, Fu, Han, Shi, Wang, He, Zhang, Luo, Qiao, Zhang, and Shao}]{meng2025mm-eureka}
Fanqing Meng, Lingxiao Du, Zongkai Liu, Zhixiang Zhou, Quanfeng Lu, Daocheng Fu, Tiancheng Han, Botian Shi, Wenhai Wang, Junjun He, Kaipeng Zhang, Ping Luo, Yu~Qiao, Qiaosheng Zhang, and Wenqi Shao. 2025.
\newblock \href {https://arxiv.org/pdf/2503.07365} {Mm-eureka: Exploring the frontiers of multimodal reasoning with rule-based reinforcement learning}.
\newblock \emph{arXiv preprint arXiv:2503.07365}.

\bibitem[{Meng et~al.(2024)Meng, Xia, and Chen}]{meng2024simpo}
Yu~Meng, Mengzhou Xia, and Danqi Chen. 2024.
\newblock \href {https://arxiv.org/pdf/2405.14734} {Simpo: Simple preference optimization with a reference-free reward}.
\newblock \emph{Advances in Neural Information Processing Systems}, 37:124198--124235.

\bibitem[{Meta(2023)}]{llama2-2023}
Meta. 2023.
\newblock \href {https://arxiv.org/pdf/2307.09288.pdf} {Llama: Open and efficient foundation language models}.
\newblock \emph{arXiv preprint arXiv:2307.09288}.

\bibitem[{Meta(2024)}]{llama3herdmodels2024}
Meta. 2024.
\newblock \href {https://arxiv.org/pdf/2407.21783} {The llama 3 herd of models}.
\newblock \emph{arXiv preprint arXiv:2407.21783}.

\bibitem[{Mirzadeh et~al.(2024)Mirzadeh, Alizadeh, Shahrokhi, Tuzel, Bengio, and Farajtabar}]{mirzadeh2024gsm-symbolic}
Iman Mirzadeh, Keivan Alizadeh, Hooman Shahrokhi, Oncel Tuzel, Samy Bengio, and Mehrdad Farajtabar. 2024.
\newblock \href {{https://arxiv.org/pdf/2410.05229}} {Gsm-symbolic: Understanding the limitations of mathematical reasoning in large language models}.
\newblock \emph{arXiv preprint arXiv:2410.05229}.

\bibitem[{Nakano et~al.(2021)Nakano, Hilton, Balaji, Wu, Ouyang, Kim, Hesse, Jain, Kosaraju, Saunders, Jiang, Cobbe, Eloundou, Krueger, Button, Knight, Chess, and Schulman}]{nakano2021webgpt}
Reiichiro Nakano, Jacob Hilton, Suchir Balaji, Jeff Wu, Long Ouyang, Christina Kim, Christopher Hesse, Shantanu Jain, Vineet Kosaraju, William Saunders, Xu~Jiang, Karl Cobbe, Tyna Eloundou, Gretchen Krueger, Kevin Button, Matthew Knight, Benjamin Chess, and John Schulman. 2021.
\newblock \href {https://arxiv.org/pdf/2112.09332} {Webgpt: Browser-assisted question-answering with human feedback}.
\newblock \emph{arXiv preprint arXiv:2112.09332}.

\bibitem[{Ni et~al.(2023)Ni, Iyer, Radev, Stoyanov, Yih, Wang, and Lin}]{ni2023lever}
Ansong Ni, Srini Iyer, Dragomir Radev, Veselin Stoyanov, Wen-tau Yih, Sida Wang, and Xi~Victoria Lin. 2023.
\newblock \href {https://proceedings.mlr.press/v202/ni23b/ni23b.pdf} {Lever: Learning to verify language-to-code generation with execution}.
\newblock In \emph{International Conference on Machine Learning}, pages 26106--26128. PMLR.

\bibitem[{OpenAI(2023)}]{openai2023gpt4}
OpenAI. 2023.
\newblock \href {https://arxiv.org/abs/2303.08774} {Gpt-4 technical report}.
\newblock \emph{Preprint}, arXiv:2303.08774.

\bibitem[{OpenAI(2025)}]{openai-deepresearch}
OpenAI. 2025.
\newblock \href {https://openai.com/index/introducing-deep-research/} {Introducing deep research}.
\newblock openai.com.

\bibitem[{Ouyang et~al.(2022)Ouyang, Wu, Jiang, Almeida, Wainwright, Mishkin, Zhang, Agarwal, Slama, Ray, Schulman, Hilton, Kelton, Miller, Simens, Askell, Welinder, Christiano, Leike, and Lowe}]{ouyang2022rlhf}
Long Ouyang, Jeffrey Wu, Xu~Jiang, Diogo Almeida, Carroll Wainwright, Pamela Mishkin, Chong Zhang, Sandhini Agarwal, Katarina Slama, Alex Ray, John Schulman, Jacob Hilton, Fraser Kelton, Luke Miller, Maddie Simens, Amanda Askell, Peter Welinder, Paul~F Christiano, Jan Leike, and Ryan Lowe. 2022.
\newblock \href {https://proceedings.neurips.cc/paper_files/paper/2022/file/b1efde53be364a73914f58805a001731-Paper-Conference.pdf} {Training language models to follow instructions with human feedback}.
\newblock In \emph{Advances in Neural Information Processing Systems}, volume~35, pages 27730--27744. Curran Associates, Inc.

\bibitem[{Pan et~al.(2022)Pan, Bhatia, and Steinhardt}]{pan2022effects-reard-misspecification}
Alexander Pan, Kush Bhatia, and Jacob Steinhardt. 2022.
\newblock \href {{https://arxiv.org/pdf/2201.03544}} {The effects of reward misspecification: Mapping and mitigating misaligned models}.
\newblock \emph{arXiv preprint arXiv:2201.03544}.

\bibitem[{Pan et~al.(2024{\natexlab{a}})Pan, Jones, Jagadeesan, and Steinhardt}]{pan2024feedback-loops-in-context-reward-hacking}
Alexander Pan, Erik Jones, Meena Jagadeesan, and Jacob Steinhardt. 2024{\natexlab{a}}.
\newblock \href {{https://arxiv.org/pdf/2402.06627}} {Feedback loops with language models drive in-context reward hacking}.
\newblock \emph{arXiv preprint arXiv:2402.06627}.

\bibitem[{Pan et~al.(2024{\natexlab{b}})Pan, He, Bowman, and Feng}]{pan2024reward-hacking-self-refinement}
Jane Pan, He~He, Samuel~R Bowman, and Shi Feng. 2024{\natexlab{b}}.
\newblock \href {{https://arxiv.org/pdf/2407.04549}} {Spontaneous reward hacking in iterative self-refinement}.
\newblock \emph{arXiv preprint arXiv:2407.04549}.

\bibitem[{Pan et~al.(2025)Pan, Liu, Wu, Liu, Zhu, Li, Chen, Ouyang, and Rueckert}]{pan2025medvlm-r1}
Jiazhen Pan, Che Liu, Junde Wu, Fenglin Liu, Jiayuan Zhu, Hongwei~Bran Li, Chen Chen, Cheng Ouyang, and Daniel Rueckert. 2025.
\newblock \href {{https://arxiv.org/pdf/2502.19634}} {Medvlm-r1: Incentivizing medical reasoning capability of vision-language models (vlms) via reinforcement learning}.
\newblock \emph{arXiv preprint arXiv:2502.19634}.

\bibitem[{Pan et~al.(2023{\natexlab{a}})Pan, Albalak, Wang, and Wang}]{pan2023logic-lm}
Liangming Pan, Alon Albalak, Xinyi Wang, and William~Yang Wang. 2023{\natexlab{a}}.
\newblock \href {{https://arxiv.org/pdf/2305.12295}} {Logic-lm: Empowering large language models with symbolic solvers for faithful logical reasoning}.
\newblock \emph{arXiv preprint arXiv:2305.12295}.

\bibitem[{Pan et~al.(2023{\natexlab{b}})Pan, Saxon, Xu, Nathani, Wang, and Wang}]{pan2023automatically}
Liangming Pan, Michael Saxon, Wenda Xu, Deepak Nathani, Xinyi Wang, and William~Yang Wang. 2023{\natexlab{b}}.
\newblock \href {{https://arxiv.org/pdf/2308.03188}} {Automatically correcting large language models: Surveying the landscape of diverse self-correction strategies}.
\newblock \emph{arXiv preprint arXiv:2308.03188}.

\bibitem[{Park et~al.(2024)Park, Liu, Gong, and Choi}]{park2024ensembling-le-mcts}
Sungjin Park, Xiao Liu, Yeyun Gong, and Edward Choi. 2024.
\newblock \href {{https://arxiv.org/pdf/2412.15797}} {Ensembling large language models with process reward-guided tree search for better complex reasoning}.
\newblock \emph{arXiv preprint arXiv:2412.15797}.

\bibitem[{Paul et~al.(2023)Paul, Ismayilzada, Peyrard, Borges, Bosselut, West, and Faltings}]{paul2023refiner}
Debjit Paul, Mete Ismayilzada, Maxime Peyrard, Beatriz Borges, Antoine Bosselut, Robert West, and Boi Faltings. 2023.
\newblock \href {{https://arxiv.org/pdf/2304.01904}} {Refiner: Reasoning feedback on intermediate representations}.
\newblock \emph{arXiv preprint arXiv:2304.01904}.

\bibitem[{Peng et~al.(2023)Peng, Galley, He, Cheng, Xie, Hu, Huang, Liden, Yu, Chen et~al.}]{peng2023llmaugmenter}
Baolin Peng, Michel Galley, Pengcheng He, Hao Cheng, Yujia Xie, Yu~Hu, Qiuyuan Huang, Lars Liden, Zhou Yu, Weizhu Chen, et~al. 2023.
\newblock \href {{https://arxiv.org/pdf/2302.12813}} {Check your facts and try again: Improving large language models with external knowledge and automated feedback}.
\newblock \emph{arXiv preprint arXiv:2302.12813}.

\bibitem[{Peng et~al.(2025)Peng, Qi, Wang, Yao, Xu, Hou, and Li}]{peng2025agentic-reward-modeling}
Hao Peng, Yunjia Qi, Xiaozhi Wang, Zijun Yao, Bin Xu, Lei Hou, and Juanzi Li. 2025.
\newblock \href {{https://arxiv.org/pdf/2502.19328}} {Agentic reward modeling: Integrating human preferences with verifiable correctness signals for reliable reward systems}.
\newblock \emph{arXiv preprint arXiv:2502.19328}.

\bibitem[{Pi et~al.(2024)Pi, Han, Xiong, Zhang, Liu, Pan, and Zhang}]{pi2024strengthening-bpo}
Renjie Pi, Tianyang Han, Wei Xiong, Jipeng Zhang, Runtao Liu, Rui Pan, and Tong Zhang. 2024.
\newblock \href {{https://arxiv.org/pdf/2403.08730}} {Strengthening multimodal large language model with bootstrapped preference optimization}.
\newblock \emph{arXiv preprint arXiv:2403.08730}.

\bibitem[{Prasad et~al.(2024)Prasad, Yuan, Pang, Xu, Fazel-Zarandi, Bansal, Sukhbaatar, Weston, and Yu}]{prasad2024scpo-self-consistency}
Archiki Prasad, Weizhe Yuan, Richard~Yuanzhe Pang, Jing Xu, Maryam Fazel-Zarandi, Mohit Bansal, Sainbayar Sukhbaatar, Jason Weston, and Jane Yu. 2024.
\newblock \href {https://arxiv.org/pdf/2411.04109} {Self-consistency preference optimization}.

\bibitem[{Qi et~al.(2024)Qi, Ma, Xu, Zhang, Yang, and Yang}]{qi2024rstar}
Zhenting Qi, Mingyuan Ma, Jiahang Xu, Li~Lyna Zhang, Fan Yang, and Mao Yang. 2024.
\newblock \href {{https://arxiv.org/pdf/2408.06195}} {Mutual reasoning makes smaller llms stronger problem-solvers}.
\newblock \emph{arXiv preprint arXiv:2408.06195}.

\bibitem[{Qian et~al.(2025)Qian, Acikgoz, He, Wang, Chen, Hakkani-Tür, Tur, and Ji}]{qian2025toolrl}
Cheng Qian, Emre~Can Acikgoz, Qi~He, Hongru Wang, Xiusi Chen, Dilek Hakkani-Tür, Gokhan Tur, and Heng Ji. 2025.
\newblock \href {https://arxiv.org/pdf/2504.13958} {Toolrl: Reward is all tool learning needs}.
\newblock \emph{arXiv preprint arXiv:2504.13958}.

\bibitem[{Qiao et~al.(2023)Qiao, Gui, Lv, Jia, Chen, and Zhang}]{qiao2023making}
Shuofei Qiao, Honghao Gui, Chengfei Lv, Qianghuai Jia, Huajun Chen, and Ningyu Zhang. 2023.
\newblock \href {{https://arxiv.org/pdf/2305.13068}} {Making language models better tool learners with execution feedback}.
\newblock \emph{arXiv preprint arXiv:2305.13068}.

\bibitem[{Qiu et~al.(2023)Qiu, Jiang, Lu, Sclar, Pyatkin, Bhagavatula, Wang, Kim, Choi, Dziri et~al.}]{qiu2023ihr-symbolic-interpreter}
Linlu Qiu, Liwei Jiang, Ximing Lu, Melanie Sclar, Valentina Pyatkin, Chandra Bhagavatula, Bailin Wang, Yoon Kim, Yejin Choi, Nouha Dziri, et~al. 2023.
\newblock \href {{https://arxiv.org/pdf/2310.08559}} {Phenomenal yet puzzling: Testing inductive reasoning capabilities of language models with hypothesis refinement}.
\newblock \emph{arXiv preprint arXiv:2310.08559}.

\bibitem[{Qu et~al.(2024)Qu, Zhang, Garg, and Kumar}]{qu2024introspection-self-improve}
Yuxiao Qu, Tianjun Zhang, Naman Garg, and Aviral Kumar. 2024.
\newblock \href {https://proceedings.neurips.cc/paper_files/paper/2024/file/639d992f819c2b40387d4d5170b8ffd7-Paper-Conference.pdf} {Recursive introspection: Teaching language model agents how to self-improve}.
\newblock \emph{Advances in Neural Information Processing Systems}, 37:55249--55285.

\bibitem[{Rafailov et~al.(2023)Rafailov, Sharma, Mitchell, Manning, Ermon, and Finn}]{rafailov2023dpo}
Rafael Rafailov, Archit Sharma, Eric Mitchell, Christopher~D Manning, Stefano Ermon, and Chelsea Finn. 2023.
\newblock \href {https://arxiv.org/pdf/2305.18290} {Direct preference optimization: Your language model is secretly a reward model}.
\newblock \emph{Advances in Neural Information Processing Systems}, 36:53728--53741.

\bibitem[{Rashid et~al.(2025)Rashid, Wu, Fan, Li, Kristiadi, and Poupart}]{rashid2025farma-costeffective-reward-guided-generation}
Ahmad Rashid, Ruotian Wu, Rongqi Fan, Hongliang Li, Agustinus Kristiadi, and Pascal Poupart. 2025.
\newblock \href {{https://arxiv.org/pdf/2502.04517}} {Towards cost-effective reward guided text generation}.
\newblock \emph{arXiv preprint arXiv:2502.04517}.

\bibitem[{Razin et~al.(2025)Razin, Wang, Strauss, Wei, Lee, and Arora}]{razin2025what-rm-good-teacher}
Noam Razin, Zixuan Wang, Hubert Strauss, Stanley Wei, Jason~D Lee, and Sanjeev Arora. 2025.
\newblock \href {{https://arxiv.org/pdf/2503.15477}} {What makes a reward model a good teacher? an optimization perspective}.
\newblock \emph{arXiv preprint arXiv:2503.15477}.

\bibitem[{Revel et~al.(2025)Revel, Cargnelutti, Eloundou, and Leppert}]{revel2025seal}
Manon Revel, Matteo Cargnelutti, Tyna Eloundou, and Greg Leppert. 2025.
\newblock \href {https://arxiv.org/pdf/2408.10270} {Seal: Systematic error analysis for value alignment}.
\newblock In \emph{Proceedings of the AAAI Conference on Artificial Intelligence}, volume~39, pages 27599--27607.

\bibitem[{Ruan et~al.(2025)Ruan, Yuan, Gao, Guo, Zhang, Xu, Hu, Liu, and Fu}]{ruan2025vlrmbench}
Jiacheng Ruan, Wenzhen Yuan, Xian Gao, Ye~Guo, Daoxin Zhang, Zhe Xu, Yao Hu, Ting Liu, and Yuzhuo Fu. 2025.
\newblock \href {{https://arxiv.org/pdf/2503.07478}} {{VLRMBench}: A comprehensive and challenging benchmark for vision-language reward models}.
\newblock \emph{arXiv preprint arXiv:2503.07478}.

\bibitem[{Russell and Santos(2019)}]{russell2019explaining-reward-functions}
Jacob Russell and Eugene Santos. 2019.
\newblock \href {https://aaai.org/ocs/index.php/FLAIRS/FLAIRS19/paper/view/18275} {Explaining reward functions in markov decision processes}.
\newblock In \emph{Proceedings of the Thirty-Second International Florida Artificial Intelligence Research Society Conference, Sarasota, Florida, USA, May 19-22 2019}, pages 56--61. {AAAI} Press.

\bibitem[{Saunders et~al.(2022)Saunders, Yeh, Wu, Bills, Ouyang, Ward, and Leike}]{saunders2022self-critiquing-assisting}
William Saunders, Catherine Yeh, Jeff Wu, Steven Bills, Long Ouyang, Jonathan Ward, and Jan Leike. 2022.
\newblock \href {{https://arxiv.org/pdf/2206.05802}} {Self-critiquing models for assisting human evaluators}.
\newblock \emph{arXiv preprint arXiv:2206.05802}.

\bibitem[{Schulman et~al.(2017)Schulman, Wolski, Dhariwal, Radford, and Klimov}]{schulman2017proximal}
John Schulman, Filip Wolski, Prafulla Dhariwal, Alec Radford, and Oleg Klimov. 2017.
\newblock \href {https://arxiv.org/pdf/1707.06347} {Proximal policy optimization algorithms}.
\newblock \emph{arXiv preprint arXiv:1707.06347}.

\bibitem[{Shao et~al.(2025)Shao, Li, Xin, Geng, Wang, Oh, Du, Lambert, Min, Krishna, Tsvetkov, Hajishirzi, Koh, and Zettlemoyer}]{shao2025spurious-rewards}
Rulin Shao, Shuyue~Stella Li, Rui Xin, Scott Geng, Yiping Wang, Sewoong Oh, Simon~Shaolei Du, Nathan Lambert, Sewon Min, Ranjay Krishna, Yulia Tsvetkov, Hannaneh Hajishirzi, Pang~Wei Koh, and Luke Zettlemoyer. 2025.
\newblock \href {https://rethink-rlvr.notion.site/Spurious-Rewards-Rethinking-Training-Signals-in-RLVR-1f4df34dac1880948858f95aeb88872f} {Spurious rewards: Rethinking training signals in rlvr}.
\newblock https://rethink-rlvr.notion.site/Spurious-Rewards-Rethinking-Training-Signals-in-RLVR-1f4df34dac1880948858f95aeb88872f.
\newblock Notion Blog.

\bibitem[{Shao et~al.(2024)Shao, Wang, Zhu, Xu, Song, Bi, Zhang, Zhang, Li, Wu et~al.}]{shao2024deepseekmath}
Zhihong Shao, Peiyi Wang, Qihao Zhu, Runxin Xu, Junxiao Song, Xiao Bi, Haowei Zhang, Mingchuan Zhang, YK~Li, Y~Wu, et~al. 2024.
\newblock \href {{https://arxiv.org/pdf/2402.03300}} {Deepseekmath: Pushing the limits of mathematical reasoning in open language models}.
\newblock \emph{arXiv preprint arXiv:2402.03300}.

\bibitem[{She et~al.(2025)She, Liu, Liu, Chen, Huang, and Huang}]{she2025reasoning-prm}
Shuaijie She, Junxiao Liu, Yifeng Liu, Jiajun Chen, Xin Huang, and Shujian Huang. 2025.
\newblock \href {{https://arxiv.org/pdf/2503.21295}} {R-prm: Reasoning-driven process reward modeling}.
\newblock \emph{arXiv preprint arXiv:2503.21295}.

\bibitem[{Shen et~al.(2025{\natexlab{a}})Shen, Liu, Li, Fang, Ma, Liao, Shen, Zhang, Zhao, Zhang et~al.}]{shen2025vlm-r1}
Haozhan Shen, Peng Liu, Jingcheng Li, Chunxin Fang, Yibo Ma, Jiajia Liao, Qiaoli Shen, Zilun Zhang, Kangjia Zhao, Qianqian Zhang, et~al. 2025{\natexlab{a}}.
\newblock \href {{https://arxiv.org/pdf/2504.07615}} {Vlm-r1: A stable and generalizable r1-style large vision-language model}.
\newblock \emph{arXiv preprint arXiv:2504.07615}.

\bibitem[{Shen et~al.(2025{\natexlab{b}})Shen, Liu, Wu, Zhu, Yang, Xin, Yue, and Yan}]{shen2025data-scaling-rlhf}
Wei Shen, Guanlin Liu, Zheng Wu, Ruofei Zhu, Qingping Yang, Chao Xin, Yu~Yue, and Lin Yan. 2025{\natexlab{b}}.
\newblock \href {{https://arxiv.org/pdf/2503.22230}} {Exploring data scaling trends and effects in reinforcement learning from human feedback}.
\newblock \emph{arXiv preprint arXiv:2503.22230}.

\bibitem[{Shinn et~al.(2023)Shinn, Cassano, Labash, Gopinath, Narasimhan, and Yao}]{shinn2023reflexion}
Noah Shinn, Federico Cassano, Beck Labash, Ashwin Gopinath, Karthik Narasimhan, and Shunyu Yao. 2023.
\newblock \href {https://arxiv.org/pdf/2303.11366} {Reflexion: Language agents with verbal reinforcement learning}.
\newblock \emph{arxiv preprint arXiv:2303.11366}.

\bibitem[{Silver and Sutton(2025)}]{silver2025welcome-era-experience}
David Silver and Richard~S Sutton. 2025.
\newblock \href {http://incompleteideas.net/papers/TheEraOfExperience.pdf} {Welcome to the era of experience}.
\newblock \emph{Google AI}.

\bibitem[{Snell et~al.(2025)Snell, Lee, Xu, and Kumar}]{snell2025scaling}
Charlie~Victor Snell, Jaehoon Lee, Kelvin Xu, and Aviral Kumar. 2025.
\newblock \href {https://arxiv.org/pdf/2408.03314} {Scaling llm test-time compute optimally can be more effective than scaling parameters for reasoning}.
\newblock In \emph{The Thirteenth International Conference on Learning Representations}, volume~2, page~7.

\bibitem[{Song et~al.(2025{\natexlab{a}})Song, Jiang, Min, Chen, Chen, Zhao, Fang, and Wen}]{song2025r1-searcher}
Huatong Song, Jinhao Jiang, Yingqian Min, Jie Chen, Zhipeng Chen, Wayne~Xin Zhao, Lei Fang, and Ji-Rong Wen. 2025{\natexlab{a}}.
\newblock \href {{https://arxiv.org/pdf/2503.05592}} {R1-searcher: Incentivizing the search capability in llms via reinforcement learning}.
\newblock \emph{arXiv preprint arXiv:2503.05592}.

\bibitem[{Song et~al.(2025{\natexlab{b}})Song, Su, Qu, Zhou, and Cheng}]{song2025prmbench}
Mingyang Song, Zhaochen Su, Xiaoye Qu, Jiawei Zhou, and Yu~Cheng. 2025{\natexlab{b}}.
\newblock \href {{https://arxiv.org/pdf/2501.03124}} {Prmbench: A fine-grained and challenging benchmark for process-level reward models}.
\newblock \emph{arXiv preprint arXiv:2501.03124}.

\bibitem[{Sun et~al.(2024{\natexlab{a}})Sun, Haider, Zhang, Yang, Qiu, Yin, Wang, Bartlett, and Zanette}]{sun2024fast-best-of-n}
Hanshi Sun, Momin Haider, Ruiqi Zhang, Huitao Yang, Jiahao Qiu, Ming Yin, Mengdi Wang, Peter Bartlett, and Andrea Zanette. 2024{\natexlab{a}}.
\newblock \href {{https://arxiv.org/pdf/2410.20290}} {Fast best-of-n decoding via speculative rejection}.
\newblock \emph{arXiv preprint arXiv:2410.20290}.

\bibitem[{Sun et~al.(2024{\natexlab{b}})Sun, Li, Yuan, Yuan, Li, and Liu}]{sun2024metacritique}
Shichao Sun, Junlong Li, Weizhe Yuan, Ruifeng Yuan, Wenjie Li, and Pengfei Liu. 2024{\natexlab{b}}.
\newblock \href {{https://arxiv.org/pdf/2401.04518}} {The critique of critique}.
\newblock \emph{arXiv preprint arXiv:2401.04518}.

\bibitem[{Sun et~al.(2023)Sun, Shen, Cao, Liu, Li, Shen, Gan, Gui, Wang, Yang et~al.}]{sun2023fact-rlhf}
Zhiqing Sun, Sheng Shen, Shengcao Cao, Haotian Liu, Chunyuan Li, Yikang Shen, Chuang Gan, Liang-Yan Gui, Yu-Xiong Wang, Yiming Yang, et~al. 2023.
\newblock \href {{https://arxiv.org/pdf/2309.14525}} {Aligning large multimodal models with factually augmented rlhf}.
\newblock \emph{arXiv preprint arXiv:2309.14525}.

\bibitem[{Sutton et~al.(1999)Sutton, McAllester, Singh, and Mansour}]{sutton1999policy}
Richard~S Sutton, David McAllester, Satinder Singh, and Yishay Mansour. 1999.
\newblock \href {http://papers.neurips.cc/paper/1713-policy-gradient-methods-for-reinforcement-learning-with-function-approximation.pdf} {Policy gradient methods for reinforcement learning with function approximation}.
\newblock \emph{Advances in neural information processing systems}, 12.

\bibitem[{Tan et~al.(2025)Tan, Ji, Hao, Lin, Wang, Wang, and Zhang}]{tan2025reason-rft}
Huajie Tan, Yuheng Ji, Xiaoshuai Hao, Minglan Lin, Pengwei Wang, Zhongyuan Wang, and Shanghang Zhang. 2025.
\newblock \href {{https://arxiv.org/pdf/2503.20752}} {Reason-rft: Reinforcement fine-tuning for visual reasoning}.
\newblock \emph{arXiv preprint arXiv:2503.20752}.

\bibitem[{Tao et~al.(2024)Tao, Chen, Yu, Mai, Rossi, Li, and Mitra}]{tao2024codelutra}
Leitian Tao, Xiang Chen, Tong Yu, Tung Mai, Ryan~A. Rossi, Yixuan Li, and Saayan Mitra. 2024.
\newblock \href {{https://arxiv.org/pdf/2411.05199}} {Codelutra: Boosting {LLM} code generation via preference-guided refinement}.
\newblock \emph{arXiv preprint arXiv:2411.05199}.

\bibitem[{Taori et~al.(2023)Taori, Gulrajani, Zhang, Dubois, Li, Guestrin, Liang, and Hashimoto}]{taori2023stanford-alpaca}
Rohan Taori, Ishaan Gulrajani, Tianyi Zhang, Yann Dubois, Xuechen Li, Carlos Guestrin, Percy Liang, and Tatsunori~B Hashimoto. 2023.
\newblock \href {https://crfm.stanford.edu/2023/03/13/alpaca.html} {Stanford alpaca: An instruction-following llama model}.

\bibitem[{Tian et~al.(2023)Tian, Mitchell, Yao, Manning, and Finn}]{tian2023factuality}
Katherine Tian, Eric Mitchell, Huaxiu Yao, Christopher~D Manning, and Chelsea Finn. 2023.
\newblock \href {https://arxiv.org/pdf/2311.08401} {Fine-tuning language models for factuality}.
\newblock In \emph{The Twelfth International Conference on Learning Representations}.

\bibitem[{Troshin et~al.(2024)Troshin, Niculae, and Fokkens}]{troshin2024lowrankrad}
Sergey Troshin, Vlad Niculae, and Antske Fokkens. 2024.
\newblock \href {{https://arxiv.org/pdf/2407.04615}} {Efficient controlled language generation with low-rank autoregressive reward models}.
\newblock \emph{arXiv preprint arXiv:2407.04615}.

\bibitem[{Tu et~al.(2025)Tu, Feng, Chen, Liu, Tang, and Xie}]{tu2025vilbench}
Haoqin Tu, Weitao Feng, Hardy Chen, Hui Liu, Xianfeng Tang, and Cihang Xie. 2025.
\newblock \href {{https://arxiv.org/pdf/2503.20271}} {Vilbench: A suite for vision-language process reward modeling}.
\newblock \emph{arXiv preprint arXiv:2503.20271}.

\bibitem[{Tyen et~al.(2023)Tyen, Mansoor, C{\u{a}}rbune, Chen, and Mak}]{tyen2023bigbench-mistake}
Gladys Tyen, Hassan Mansoor, Victor C{\u{a}}rbune, Peter Chen, and Tony Mak. 2023.
\newblock \href {{https://arxiv.org/pdf/2311.08516}} {Llms cannot find reasoning errors, but can correct them given the error location}.
\newblock \emph{arXiv preprint arXiv:2311.08516}.

\bibitem[{Uesato et~al.(2020)Uesato, Kumar, Krakovna, Everitt, Ngo, and Legg}]{uesato2020avoiding-decoupled-approval}
Jonathan Uesato, Ramana Kumar, Victoria Krakovna, Tom Everitt, Richard Ngo, and Shane Legg. 2020.
\newblock \href {{https://arxiv.org/pdf/2011.08827}} {Avoiding tampering incentives in deep rl via decoupled approval}.
\newblock \emph{arXiv preprint arXiv:2011.08827}.

\bibitem[{Uesato et~al.(2022)Uesato, Kushman, Kumar, Song, Siegel, Wang, Creswell, Irving, and Higgins}]{uesato2022solving}
Jonathan Uesato, Nate Kushman, Ramana Kumar, Francis Song, Noah Siegel, Lisa Wang, Antonia Creswell, Geoffrey Irving, and Irina Higgins. 2022.
\newblock \href {{https://arxiv.org/pdf/2211.14275}} {Solving math word problems with process-and outcome-based feedback}.
\newblock \emph{arXiv preprint arXiv:2211.14275}.

\bibitem[{Varshney et~al.(2023)Varshney, Yao, Zhang, Chen, and Yu}]{varshney2023stitch}
Neeraj Varshney, Wenlin Yao, Hongming Zhang, Jianshu Chen, and Dong Yu. 2023.
\newblock \href {{https://arxiv.org/pdf/2307.03987}} {A stitch in time saves nine: Detecting and mitigating hallucinations of llms by validating low-confidence generation}.
\newblock \emph{arXiv preprint arXiv:2307.03987}.

\bibitem[{Wallace et~al.(2024)Wallace, Dang, Rafailov, Zhou, Lou, Purushwalkam, Ermon, Xiong, Joty, and Naik}]{wallace2024diffusion}
Bram Wallace, Meihua Dang, Rafael Rafailov, Linqi Zhou, Aaron Lou, Senthil Purushwalkam, Stefano Ermon, Caiming Xiong, Shafiq Joty, and Nikhil Naik. 2024.
\newblock \href {https://arxiv.org/pdf/2311.12908} {Diffusion model alignment using direct preference optimization}.
\newblock In \emph{Proceedings of the IEEE/CVF Conference on Computer Vision and Pattern Recognition}, pages 8228--8238.

\bibitem[{Wang et~al.(2023{\natexlab{a}})Wang, Chen, Pei, Xie, Kang, Zhang, Xu, Xiong, Dutta, Schaeffer et~al.}]{wang2023decodingtrust}
Boxin Wang, Weixin Chen, Hengzhi Pei, Chulin Xie, Mintong Kang, Chenhui Zhang, Chejian Xu, Zidi Xiong, Ritik Dutta, Rylan Schaeffer, et~al. 2023{\natexlab{a}}.
\newblock \href {https://arxiv.org/pdf/2306.11698} {Decodingtrust: A comprehensive assessment of trustworthiness in gpt models}.
\newblock In \emph{NeurIPS}.

\bibitem[{Wang et~al.(2024{\natexlab{a}})Wang, Zhou, Huang, Xu, Zhang, Poon, and Chen}]{wang2024mdpo}
Fei Wang, Wenxuan Zhou, James~Y Huang, Nan Xu, Sheng Zhang, Hoifung Poon, and Muhao Chen. 2024{\natexlab{a}}.
\newblock \href {{https://arxiv.org/pdf/2406.11839}} {mdpo: Conditional preference optimization for multimodal large language models}.
\newblock \emph{arXiv preprint arXiv:2406.11839}.

\bibitem[{Wang et~al.(2024{\natexlab{b}})Wang, Xiong, Xie, Zhao, and Zhang}]{wang2024armorm}
Haoxiang Wang, Wei Xiong, Tengyang Xie, Han Zhao, and Tong Zhang. 2024{\natexlab{b}}.
\newblock \href {{https://arxiv.org/pdf/2406.12845}} {Interpretable preferences via multi-objective reward modeling and mixture-of-experts}.
\newblock \emph{arXiv preprint arXiv:2406.12845}.

\bibitem[{Wang et~al.(2025{\natexlab{a}})Wang, Qian, Zhong, Chen, Qiu, Huang, Jin, Wang, Wong, and Ji}]{wang2025tool-calls}
Hongru Wang, Cheng Qian, Wanjun Zhong, Xiusi Chen, Jiahao Qiu, Shijue Huang, Bowen Jin, Mengdi Wang, Kam-Fai Wong, and Heng Ji. 2025{\natexlab{a}}.
\newblock \href {https://arxiv.org/pdf/2504.14870} {Otc: Optimal tool calls via reinforcement learning}.
\newblock \emph{arXiv preprint arXiv:2504.14870}.

\bibitem[{Wang et~al.(2024{\natexlab{c}})Wang, Ma, Feng, Zhang, Yang, Zhang, Chen, Tang, Chen, Lin et~al.}]{wang2024survey-agents}
Lei Wang, Chen Ma, Xueyang Feng, Zeyu Zhang, Hao Yang, Jingsen Zhang, Zhiyuan Chen, Jiakai Tang, Xu~Chen, Yankai Lin, et~al. 2024{\natexlab{c}}.
\newblock \href {https://arxiv.org/pdf/2308.11432} {A survey on large language model based autonomous agents}.
\newblock \emph{Frontiers of Computer Science}, 18(6):186345.

\bibitem[{Wang et~al.(2024{\natexlab{d}})Wang, Xu, Zhou, Xiong, and Joty}]{wang2024directjudgement}
Peifeng Wang, Austin Xu, Yilun Zhou, Caiming Xiong, and Shafiq Joty. 2024{\natexlab{d}}.
\newblock \href {{https://arxiv.org/pdf/2409.14664}} {Direct judgement preference optimization}.
\newblock \emph{arXiv preprint arXiv:2409.14664}.

\bibitem[{Wang et~al.(2023{\natexlab{b}})Wang, Li, Shao, Xu, Dai, Li, Chen, Wu, and Sui}]{wang2023math-shepherd}
Peiyi Wang, Lei Li, Zhihong Shao, RX~Xu, Damai Dai, Yifei Li, Deli Chen, Yu~Wu, and Zhifang Sui. 2023{\natexlab{b}}.
\newblock \href {{https://arxiv.org/pdf/2312.08935}} {Math-shepherd: Verify and reinforce llms step-by-step without human annotations}.
\newblock \emph{arXiv preprint arXiv:2312.08935}.

\bibitem[{Wang et~al.(2025{\natexlab{b}})Wang, Jiang, He, Yang, Zheng, Li, He, Tong, and Gong}]{wang2025hierarchical-rm}
Teng Wang, Zhangyi Jiang, Zhenqi He, Wenhan Yang, Yanan Zheng, Zeyu Li, Zifan He, Shenyang Tong, and Hailei Gong. 2025{\natexlab{b}}.
\newblock \href {{https://arxiv.org/pdf/2503.13551}} {Towards hierarchical multi-step reward models for enhanced reasoning in large language models}.
\newblock \emph{arXiv preprint arXiv:2503.13551}.

\bibitem[{Wang et~al.(2024{\natexlab{e}})Wang, Kulikov, Golovneva, Yu, Yuan, Dwivedi-Yu, Pang, Fazel-Zarandi, Weston, and Li}]{wang2024self-taught}
Tianlu Wang, Ilia Kulikov, Olga Golovneva, Ping Yu, Weizhe Yuan, Jane Dwivedi-Yu, Richard~Yuanzhe Pang, Maryam Fazel-Zarandi, Jason Weston, and Xian Li. 2024{\natexlab{e}}.
\newblock \href {{https://arxiv.org/pdf/2408.02666}} {Self-taught evaluators}.
\newblock \emph{arXiv preprint arXiv:2408.02666}.

\bibitem[{Wang et~al.(2023{\natexlab{c}})Wang, Yu, Tan, O'Brien, Pasunuru, Dwivedi-Yu, Golovneva, Zettlemoyer, Fazel-Zarandi, and Celikyilmaz}]{wang2023shepherd}
Tianlu Wang, Ping Yu, Xiaoqing~Ellen Tan, Sean O'Brien, Ramakanth Pasunuru, Jane Dwivedi-Yu, Olga Golovneva, Luke Zettlemoyer, Maryam Fazel-Zarandi, and Asli Celikyilmaz. 2023{\natexlab{c}}.
\newblock \href {{https://arxiv.org/pdf/2308.04592}} {Shepherd: A critic for language model generation}.
\newblock \emph{arXiv preprint arXiv:2308.04592}.

\bibitem[{Wang et~al.(2025{\natexlab{c}})Wang, Gao, Chen, Chen, Zhu, Zhao, Liu, Cao, Ye, Zhu et~al.}]{wang2025visualprm}
Weiyun Wang, Zhangwei Gao, Lianjie Chen, Zhe Chen, Jinguo Zhu, Xiangyu Zhao, Yangzhou Liu, Yue Cao, Shenglong Ye, Xizhou Zhu, et~al. 2025{\natexlab{c}}.
\newblock \href {{https://arxiv.org/pdf/2503.10291}} {Visualprm: An effective process reward model for multimodal reasoning}.
\newblock \emph{arXiv preprint arXiv:2503.10291}.

\bibitem[{Wang et~al.(2022)Wang, Wei, Schuurmans, Le, Chi, Narang, Chowdhery, and Zhou}]{wang2022self-consistency}
Xuezhi Wang, Jason Wei, Dale Schuurmans, Quoc Le, Ed~Chi, Sharan Narang, Aakanksha Chowdhery, and Denny Zhou. 2022.
\newblock \href {{https://arxiv.org/pdf/2203.11171}} {Self-consistency improves chain of thought reasoning in language models}.
\newblock \emph{arXiv preprint arXiv:2203.11171}.

\bibitem[{Wang et~al.(2025{\natexlab{d}})Wang, Li, Zang, Wang, Lu, Jin, and Wang}]{wang2025unifiedreward-think}
Yibin Wang, Zhimin Li, Yuhang Zang, Chunyu Wang, Qinglin Lu, Cheng Jin, and Jiaqi Wang. 2025{\natexlab{d}}.
\newblock \href {{https://arxiv.org/pdf/2505.03318}} {Unified multimodal chain-of-thought reward model through reinforcement fine-tuning}.
\newblock \emph{arXiv preprint arXiv:2505.03318}.

\bibitem[{Wang et~al.(2024{\natexlab{f}})Wang, Tan, Wang, Yang, Jin, and Li}]{wang2024lift}
Yibin Wang, Zhiyu Tan, Junyan Wang, Xiaomeng Yang, Cheng Jin, and Hao Li. 2024{\natexlab{f}}.
\newblock \href {{https://arxiv.org/pdf/2412.04814}} {Lift: Leveraging human feedback for text-to-video model alignment}.
\newblock \emph{arXiv preprint arXiv:2412.04814}.

\bibitem[{Wang et~al.(2025{\natexlab{e}})Wang, Zang, Li, Jin, and Wang}]{wang2025unified}
Yibin Wang, Yuhang Zang, Hao Li, Cheng Jin, and Jiaqi Wang. 2025{\natexlab{e}}.
\newblock \href {{https://arxiv.org/pdf/2503.05236}} {Unified reward model for multimodal understanding and generation}.
\newblock \emph{arXiv preprint arXiv:2503.05236}.

\bibitem[{Wang et~al.(2025{\natexlab{f}})Wang, Feng, Lin, Cai, Bian, Yan, and Zhu}]{wang2025crowdvlm-r1}
Zhiqiang Wang, Pengbin Feng, Yanbin Lin, Shuzhang Cai, Zongao Bian, Jinghua Yan, and Xingquan Zhu. 2025{\natexlab{f}}.
\newblock \href {{https://arxiv.org/pdf/2504.03724}} {Crowdvlm-r1: Expanding r1 ability to vision language model for crowd counting using fuzzy group relative policy reward}.
\newblock \emph{arXiv preprint arXiv:2504.03724}.

\bibitem[{Wang et~al.(2025{\natexlab{g}})Wang, Wang, Wang, Zhang, Li, Yang, Yu, Nguyen, Liu, Gottlieb, Lam, Lu, Cho, Wu, Fei-Fei, Wang, Choi, and Li}]{wang2025ragen-self-evolution-agents}
Zihan Wang, Kangrui Wang, Qineng Wang, Pingyue Zhang, Linjie Li, Zhengyuan Yang, Kefan Yu, Minh~Nhat Nguyen, Licheng Liu, Eli Gottlieb, Monica Lam, Yiping Lu, Kyunghyun Cho, Jiajun Wu, Li~Fei-Fei, Lijuan Wang, Yejin Choi, and Manling Li. 2025{\natexlab{g}}.
\newblock \href {https://arxiv.org/pdf/2504.20073} {Ragen: Understanding self-evolution in llm agents via multi-turn reinforcement learning}.

\bibitem[{Wei et~al.(2025)Wei, Duchenne, Copet, Carbonneaux, Zhang, Fried, Synnaeve, Singh, and Wang}]{wei2025swe-rl}
Yuxiang Wei, Olivier Duchenne, Jade Copet, Quentin Carbonneaux, Lingming Zhang, Daniel Fried, Gabriel Synnaeve, Rishabh Singh, and Sida~I Wang. 2025.
\newblock \href {{https://arxiv.org/pdf/2502.18449}} {Swe-rl: Advancing llm reasoning via reinforcement learning on open software evolution}.
\newblock \emph{arXiv preprint arXiv:2502.18449}.

\bibitem[{Welleck et~al.(2022)Welleck, Lu, West, Brahman, Shen, Khashabi, and Choi}]{welleck2022generating-self-correct}
Sean Welleck, Ximing Lu, Peter West, Faeze Brahman, Tianxiao Shen, Daniel Khashabi, and Yejin Choi. 2022.
\newblock \href {{https://arxiv.org/pdf/2211.00053}} {Generating sequences by learning to self-correct}.
\newblock \emph{arXiv preprint arXiv:2211.00053}.

\bibitem[{Wen et~al.(2024)Wen, Lu, Guan, Lu, Lin, He, Han, and Sun}]{wen2024policy-rlfh}
Xueru Wen, Xinyu Lu, Xinyan Guan, Yaojie Lu, Hongyu Lin, Ben He, Xianpei Han, and Le~Sun. 2024.
\newblock \href {{https://arxiv.org/pdf/2406.12221}} {On-policy fine-grained knowledge feedback for hallucination mitigation}.
\newblock \emph{arXiv preprint arXiv:2406.12221}.

\bibitem[{Weng(2024)}]{weng2024rewardhack}
Lilian Weng. 2024.
\newblock \href {https://lilianweng.github.io/posts/2024-11-28-reward-hacking/} {Reward hacking in reinforcement learning.}
\newblock \emph{lilianweng.github.io}.

\bibitem[{Wu et~al.(2024{\natexlab{a}})Wu, Yuan, Golovneva, Xu, Tian, Jiao, Weston, and Sukhbaatar}]{wu2024metarewarding}
Tianhao Wu, Weizhe Yuan, Olga Golovneva, Jing Xu, Yuandong Tian, Jiantao Jiao, Jason Weston, and Sainbayar Sukhbaatar. 2024{\natexlab{a}}.
\newblock \href {{https://arxiv.org/pdf/2407.19594}} {Meta-rewarding language models: Self-improving alignment with llm-as-a-meta-judge}.
\newblock \emph{arXiv preprint arXiv:2407.19594}.

\bibitem[{Wu et~al.(2023{\natexlab{a}})Wu, Hao, Sun, Chen, Zhu, Zhao, and Li}]{wu2023hpsv2}
Xiaoshi Wu, Yiming Hao, Keqiang Sun, Yixiong Chen, Feng Zhu, Rui Zhao, and Hongsheng Li. 2023{\natexlab{a}}.
\newblock \href {{https://arxiv.org/pdf/2306.09341}} {Human preference score v2: A solid benchmark for evaluating human preferences of text-to-image synthesis}.
\newblock \emph{arXiv preprint arXiv:2306.09341}.

\bibitem[{Wu et~al.(2023{\natexlab{b}})Wu, Sun, Zhu, Zhao, and Li}]{wu2023hps}
Xiaoshi Wu, Keqiang Sun, Feng Zhu, Rui Zhao, and Hongsheng Li. 2023{\natexlab{b}}.
\newblock \href {https://arxiv.org/pdf/2303.14420} {Human preference score: Better aligning text-to-image models with human preference}.
\newblock In \emph{Proceedings of the IEEE/CVF International Conference on Computer Vision}, pages 2096--2105.

\bibitem[{Wu et~al.(2023{\natexlab{c}})Wu, Hu, Shi, Dziri, Suhr, Ammanabrolu, Smith, Ostendorf, and Hajishirzi}]{wu2023fine-grained-rlhf}
Zeqiu Wu, Yushi Hu, Weijia Shi, Nouha Dziri, Alane Suhr, Prithviraj Ammanabrolu, Noah~A Smith, Mari Ostendorf, and Hannaneh Hajishirzi. 2023{\natexlab{c}}.
\newblock \href {https://arxiv.org/pdf/2306.01693} {Fine-grained human feedback gives better rewards for language model training}.
\newblock \emph{Advances in Neural Information Processing Systems}, 36:59008--59033.

\bibitem[{Wu et~al.(2024{\natexlab{b}})Wu, Qiu, Ross, Aky{\"u}rek, Chen, Wang, Kim, Andreas, and Kim}]{wu2024reasoning-or-reciting}
Zhaofeng Wu, Linlu Qiu, Alexis Ross, Ekin Aky{\"u}rek, Boyuan Chen, Bailin Wang, Najoung Kim, Jacob Andreas, and Yoon Kim. 2024{\natexlab{b}}.
\newblock \href {https://aclanthology.org/2024.naacl-long.102/} {Reasoning or reciting? exploring the capabilities and limitations of language models through counterfactual tasks}.
\newblock In \emph{Proceedings of the 2024 Conference of the North American Chapter of the Association for Computational Linguistics: Human Language Technologies (Volume 1: Long Papers)}, pages 1819--1862.

\bibitem[{Xi et~al.(2024)Xi, Yang, Huang, Tang, Li, Ding, He, Hong, Do, Zhan et~al.}]{xi2024automathcritique}
Zhiheng Xi, Dingwen Yang, Jixuan Huang, Jiafu Tang, Guanyu Li, Yiwen Ding, Wei He, Boyang Hong, Shihan Do, Wenyu Zhan, et~al. 2024.
\newblock \href {{https://arxiv.org/pdf/2411.16579}} {Enhancing llm reasoning via critique models with test-time and training-time supervision}.
\newblock \emph{arXiv preprint arXiv:2411.16579}.

\bibitem[{Xia et~al.(2024)Xia, Li, Liu, Wu, and Liu}]{xia2024mr-math}
Shijie Xia, Xuefeng Li, Yixin Liu, Tongshuang Wu, and Pengfei Liu. 2024.
\newblock \href {{https://arxiv.org/pdf/2404.05692}} {Evaluating mathematical reasoning beyond accuracy}.
\newblock \emph{arXiv preprint arXiv:2404.05692}.

\bibitem[{Xia et~al.(2025)Xia, Fan, Chen, Yan, Cong, Zhang, Lu, Lin, Liu, and Sun}]{xia2025agentrm}
Yu~Xia, Jingru Fan, Weize Chen, Siyu Yan, Xin Cong, Zhong Zhang, Yaxi Lu, Yankai Lin, Zhiyuan Liu, and Maosong Sun. 2025.
\newblock \href {{https://arxiv.org/pdf/2502.18407}} {Agentrm: Enhancing agent generalization with reward modeling}.
\newblock \emph{arXiv preprint arXiv:2502.18407}.

\bibitem[{Xie et~al.(2025{\natexlab{a}})Xie, Gao, Ren, Luo, Hong, Dai, Zhou, Qiu, Wu, and Luo}]{xie2025logicrl}
Tian Xie, Zitian Gao, Qingnan Ren, Haoming Luo, Yuqian Hong, Bryan Dai, Joey Zhou, Kai Qiu, Zhirong Wu, and Chong Luo. 2025{\natexlab{a}}.
\newblock \href {{https://arxiv.org/pdf/2502.14768}} {Logic-rl: Unleashing llm reasoning with rule-based reinforcement learning}.
\newblock \emph{arXiv preprint arXiv:2502.14768}.

\bibitem[{Xie et~al.(2023)Xie, Kawaguchi, Zhao, Zhao, Kan, He, and Xie}]{xie2023self}
Yuxi Xie, Kenji Kawaguchi, Yiran Zhao, James~Xu Zhao, Min-Yen Kan, Junxian He, and Michael Xie. 2023.
\newblock \href {https://arxiv.org/pdf/2305.00633} {Self-evaluation guided beam search for reasoning}.
\newblock \emph{Advances in Neural Information Processing Systems}, 36:41618--41650.

\bibitem[{Xie et~al.(2025{\natexlab{b}})Xie, Chen, Mao, Xu, Kong et~al.}]{xie2025teaching-ctrl}
Zhihui Xie, Liyu Chen, Weichao Mao, Jingjing Xu, Lingpeng Kong, et~al. 2025{\natexlab{b}}.
\newblock \href {{https://arxiv.org/pdf/2502.03492}} {Teaching language models to critique via reinforcement learning}.
\newblock \emph{arXiv preprint arXiv:2502.03492}.

\bibitem[{Xiong et~al.(2024)Xiong, Wang, Guo, Ye, Fan, Gu, Huang, and Li}]{xiong2024llava-critic}
Tianyi Xiong, Xiyao Wang, Dong Guo, Qinghao Ye, Haoqi Fan, Quanquan Gu, Heng Huang, and Chunyuan Li. 2024.
\newblock \href {{https://arxiv.org/pdf/2410.02712}} {Llava-critic: Learning to evaluate multimodal models}.
\newblock \emph{arXiv preprint arXiv:2410.02712}.

\bibitem[{Xiong et~al.(2025)Xiong, Zhang, Ye, Chen, Jiang, and Zhang}]{xiong2025rewardingcorrection}
Wei Xiong, Hanning Zhang, Chenlu Ye, Lichang Chen, Nan Jiang, and Tong Zhang. 2025.
\newblock \href {{https://arxiv.org/pdf/2502.19613}} {Self-rewarding correction for mathematical reasoning}.
\newblock \emph{arXiv preprint arXiv:2502.19613}.

\bibitem[{Xu et~al.(2025{\natexlab{a}})Xu, Mao, Li, Wu, Chen, Zhang, and Luu}]{xu2025fullstepdpo}
Huimin Xu, Xin Mao, Feng-Lin Li, Xiaobao Wu, Wang Chen, Wei Zhang, and Anh~Tuan Luu. 2025{\natexlab{a}}.
\newblock \href {{https://arxiv.org/pdf/2502.14356}} {Full-step-dpo: Self-supervised preference optimization with step-wise rewards for mathematical reasoning}.
\newblock \emph{arXiv preprint arXiv:2502.14356}.

\bibitem[{Xu et~al.(2025{\natexlab{b}})Xu, Mao, Li, Wu, Chen, Zhang, and Luu}]{xu2025scope}
Huimin Xu, Xin Mao, Feng-Lin Li, Xiaobao Wu, Wang Chen, Wei Zhang, and Anh~Tuan Luu. 2025{\natexlab{b}}.
\newblock \href {https://arxiv.org/pdf/2505.14419} {Scope: Compress mathematical reasoning steps for efficient automated process annotation}.
\newblock \emph{arXiv preprint arXiv:2505.14419}.

\bibitem[{Xu et~al.(2023)Xu, Liu, Wu, Tong, Li, Ding, Tang, and Dong}]{xu2023imagereward}
Jiazheng Xu, Xiao Liu, Yuchen Wu, Yuxuan Tong, Qinkai Li, Ming Ding, Jie Tang, and Yuxiao Dong. 2023.
\newblock \href {https://arxiv.org/pdf/2304.05977} {Imagereward: Learning and evaluating human preferences for text-to-image generation}.
\newblock \emph{Advances in Neural Information Processing Systems}, 36:15903--15935.

\bibitem[{Xu et~al.(2025{\natexlab{c}})Xu, Savani, Fang, and Kolter}]{xu2025downsampling-rollouts-rl}
Yixuan~Even Xu, Yash Savani, Fei Fang, and Zico Kolter. 2025{\natexlab{c}}.
\newblock \href {{https://arxiv.org/pdf/2504.13818}} {Not all rollouts are useful: Down-sampling rollouts in llm reinforcement learning}.
\newblock \emph{arXiv preprint arXiv:2504.13818}.

\bibitem[{Yao et~al.(2023)Yao, Yu, Zhao, Shafran, Griffiths, Cao, and Narasimhan}]{yao2023treeofthoughts}
Shunyu Yao, Dian Yu, Jeffrey Zhao, Izhak Shafran, Tom Griffiths, Yuan Cao, and Karthik Narasimhan. 2023.
\newblock \href {https://arxiv.org/pdf/2305.10601} {Tree of thoughts: Deliberate problem solving with large language models}.
\newblock \emph{Advances in neural information processing systems}, 36:11809--11822.

\bibitem[{Yasunaga et~al.(2025)Yasunaga, Zettlemoyer, and Ghazvininejad}]{yasunaga2025multimodal-rewardbench}
Michihiro Yasunaga, Luke Zettlemoyer, and Marjan Ghazvininejad. 2025.
\newblock \href {{https://arxiv.org/pdf/2502.14191}} {Multimodal rewardbench: Holistic evaluation of reward models for vision language models}.
\newblock \emph{arXiv preprint arXiv:2502.14191}.

\bibitem[{Ye et~al.(2024{\natexlab{a}})Ye, Greenlee-Scott, Bartolo, Blunsom, Campos, and Gall{\'e}}]{ye2024improving-rm-synthetic-critiques}
Zihuiwen Ye, Fraser Greenlee-Scott, Max Bartolo, Phil Blunsom, Jon~Ander Campos, and Matthias Gall{\'e}. 2024{\natexlab{a}}.
\newblock \href {{https://arxiv.org/pdf/2405.20850}} {Improving reward models with synthetic critiques}.
\newblock \emph{arXiv preprint arXiv:2405.20850}.

\bibitem[{Ye et~al.(2024{\natexlab{b}})Ye, Li, Li, Ai, Zhou, Shen, Yan, and Liu}]{ye2024con-j}
Ziyi Ye, Xiangsheng Li, Qiuchi Li, Qingyao Ai, Yujia Zhou, Wei Shen, Dong Yan, and Yiqun Liu. 2024{\natexlab{b}}.
\newblock \href {{https://arxiv.org/pdf/2410.03742}} {Beyond scalar reward model: Learning generative judge from preference data}.
\newblock \emph{arXiv preprint arXiv:2410.03742}.

\bibitem[{Yu et~al.(2023{\natexlab{a}})Yu, Gao, and Wang}]{yu2023ovm}
Fei Yu, Anningzhe Gao, and Benyou Wang. 2023{\natexlab{a}}.
\newblock \href {{https://arxiv.org/pdf/2311.09724}} {Ovm, outcome-supervised value models for planning in mathematical reasoning}.
\newblock \emph{arXiv preprint arXiv:2311.09724}.

\bibitem[{Yu et~al.(2025{\natexlab{a}})Yu, Sun, Hu, Yan, Yu, and Li}]{yu2025improve-llm-as-a-judge}
Jiachen Yu, Shaoning Sun, Xiaohui Hu, Jiaxu Yan, Kaidong Yu, and Xuelong Li. 2025{\natexlab{a}}.
\newblock \href {{https://arxiv.org/pdf/2502.11689}} {Improve llm-as-a-judge ability as a general ability}.
\newblock \emph{arXiv preprint arXiv:2502.11689}.

\bibitem[{Yu et~al.(2025{\natexlab{b}})Yu, Zhang, Zhu, Yuan, Zuo, Yue, Fan, Liu, Liu, Liu et~al.}]{yu2025dapo}
Qiying Yu, Zheng Zhang, Ruofei Zhu, Yufeng Yuan, Xiaochen Zuo, Yu~Yue, Tiantian Fan, Gaohong Liu, Lingjun Liu, Xin Liu, et~al. 2025{\natexlab{b}}.
\newblock \href {{https://arxiv.org/pdf/2503.14476}} {Dapo: An open-source llm reinforcement learning system at scale}.
\newblock \emph{arXiv preprint arXiv:2503.14476}.

\bibitem[{Yu et~al.(2023{\natexlab{b}})Yu, Yao, Zhang, He, Han, Cui, Hu, Liu, Zheng, Sun, and Chua}]{yu2024rlhf-v}
Tianyu Yu, Yuan Yao, Haoye Zhang, Taiwen He, Yifeng Han, Ganqu Cui, Jinyi Hu, Zhiyuan Liu, Hai{-}Tao Zheng, Maosong Sun, and Tat{-}Seng Chua. 2023{\natexlab{b}}.
\newblock \href {https://arxiv.org/pdf/2312.00849} {{RLHF-V:} towards trustworthy mllms via behavior alignment from fine-grained correctional human feedback}.
\newblock \emph{arXiv preprint arXiv:2312.00849}.

\bibitem[{Yu et~al.(2024{\natexlab{a}})Yu, Zhang, Yao, Dang, Chen, Lu, Cui, He, Liu, Chua, and Sun}]{yu2024rlaif-v}
Tianyu Yu, Haoye Zhang, Yuan Yao, Yunkai Dang, Da~Chen, Xiaoman Lu, Ganqu Cui, Taiwen He, Zhiyuan Liu, Tat-Seng Chua, and Maosong Sun. 2024{\natexlab{a}}.
\newblock \href {https://doi.org/10.48550/arXiv.2405.17220} {Rlaif-v: Aligning mllms through open-source ai feedback for super gpt-4v trustworthiness}.
\newblock \emph{arXiv preprint arXiv:2405.17220}.

\bibitem[{Yu et~al.(2023{\natexlab{c}})Yu, Zhang, Liang, Jiang, and Sabharwal}]{yu2023refeed}
Wenhao Yu, Zhihan Zhang, Zhenwen Liang, Meng Jiang, and Ashish Sabharwal. 2023{\natexlab{c}}.
\newblock \href {{https://arxiv.org/pdf/2305.14002}} {Improving language models via plug-and-play retrieval feedback}.
\newblock \emph{arXiv preprint arXiv:2305.14002}.

\bibitem[{Yu et~al.(2024{\natexlab{b}})Yu, Chen, Zhang, Tan, Zhu, Pang, Qian, Wang, Gururangan, Zhang, Kambadur, Mahajan, and Hou}]{yu2024critic-rm}
Yue Yu, Zhengxing Chen, Aston Zhang, Liang Tan, Chenguang Zhu, Richard~Yuanzhe Pang, Yundi Qian, Xuewei Wang, Suchin Gururangan, Chao Zhang, Melanie Kambadur, Dhruv Mahajan, and Rui Hou. 2024{\natexlab{b}}.
\newblock \href {{https://arxiv.org/pdf/2411.16646}} {Self-generated critiques boost reward modeling for language models}.
\newblock \emph{arXiv preprint arXiv:2411.16646}.

\bibitem[{Yu et~al.(2024{\natexlab{c}})Yu, Gu, Wang, Zeng, Wang, Ye, and Zhang}]{yu2024orps}
Zhuohao Yu, Weizheng Gu, Yidong Wang, Zhengran Zeng, Jindong Wang, Wei Ye, and Shikun Zhang. 2024{\natexlab{c}}.
\newblock \href {{https://arxiv.org/pdf/2412.15118}} {Outcome-refining process supervision for code generation}.
\newblock \emph{arXiv preprint arXiv:2412.15118}.

\bibitem[{Yuan et~al.(2024{\natexlab{a}})Yuan, Li, Chen, Cui, Ding, Zhang, Zhou, Liu, and Peng}]{yuan2024free-process-rewards}
Lifan Yuan, Wendi Li, Huayu Chen, Ganqu Cui, Ning Ding, Kaiyan Zhang, Bowen Zhou, Zhiyuan Liu, and Hao Peng. 2024{\natexlab{a}}.
\newblock \href {{https://arxiv.org/pdf/2412.01981}} {Free process rewards without process labels}.
\newblock \emph{arXiv preprint arXiv:2412.01981}.

\bibitem[{Yuan et~al.(2024{\natexlab{b}})Yuan, Pang, Cho, Li, Sukhbaatar, Xu, and Weston}]{yuan2024self-rewarding}
Weizhe Yuan, Richard~Yuanzhe Pang, Kyunghyun Cho, Xian Li, Sainbayar Sukhbaatar, Jing Xu, and Jason Weston. 2024{\natexlab{b}}.
\newblock \href {{https://arxiv.org/pdf/2401.10020}} {Self-rewarding language models}.
\newblock \emph{arXiv preprint arXiv:2401.10020}.

\bibitem[{Yuan et~al.(2023{\natexlab{a}})Yuan, Yuan, Li, Dong, Lu, Tan, Zhou, and Zhou}]{yuan2023scaling-math-rsft}
Zheng Yuan, Hongyi Yuan, Chengpeng Li, Guanting Dong, Keming Lu, Chuanqi Tan, Chang Zhou, and Jingren Zhou. 2023{\natexlab{a}}.
\newblock \href {{https://arxiv.org/pdf/2308.01825}} {Scaling relationship on learning mathematical reasoning with large language models}.
\newblock \emph{arXiv preprint arXiv:2308.01825}.

\bibitem[{Yuan et~al.(2023{\natexlab{b}})Yuan, Yuan, Tan, Wang, Huang, and Huang}]{yuan2023rrhf-rank-responses-align}
Zheng Yuan, Hongyi Yuan, Chuanqi Tan, Wei Wang, Songfang Huang, and Fei Huang. 2023{\natexlab{b}}.
\newblock \href {{https://arxiv.org/pdf/2304.05302}} {Rrhf: Rank responses to align language models with human feedback without tears}.
\newblock \emph{arXiv preprint arXiv:2304.05302}.

\bibitem[{Zeng et~al.(2023)Zeng, Chen, Liu, Jiang, and Jia}]{zeng2023mr-gsm8k}
Zhongshen Zeng, Pengguang Chen, Shu Liu, Haiyun Jiang, and Jiaya Jia. 2023.
\newblock \href {{https://arxiv.org/pdf/2312.17080}} {Mr-gsm8k: A meta-reasoning benchmark for large language model evaluation}.
\newblock \emph{arXiv preprint arXiv:2312.17080}.

\bibitem[{Zeng et~al.(2024)Zeng, Liu, Wan, Li, Chen, Dai, Yao, Xu, Qi, Zhao et~al.}]{zeng2024mr-ben}
Zhongshen Zeng, Yinhong Liu, Yingjia Wan, Jingyao Li, Pengguang Chen, Jianbo Dai, Yuxuan Yao, Rongwu Xu, Zehan Qi, Wanru Zhao, et~al. 2024.
\newblock \href {{https://arxiv.org/pdf/2406.13975}} {Mr-ben: A meta-reasoning benchmark for evaluating system-2 thinking in llms}.
\newblock \emph{arXiv preprint arXiv:2406.13975}.

\bibitem[{Zhan et~al.(2025)Zhan, Zhu, Zheng, Zhao, Yang, Tang, and Wang}]{zhan2025vision-r1-object-localization}
Yufei Zhan, Yousong Zhu, Shurong Zheng, Hongyin Zhao, Fan Yang, Ming Tang, and Jinqiao Wang. 2025.
\newblock \href {{https://arxiv.org/pdf/2503.18013}} {Vision-r1: Evolving human-free alignment in large vision-language models via vision-guided reinforcement learning}.
\newblock \emph{arXiv preprint arXiv:2503.18013}.

\bibitem[{Zhang et~al.(2024{\natexlab{a}})Zhang, Zhoubian, Hu, Yue, Dong, and Tang}]{zhang2024rest-mcts*}
Dan Zhang, Sining Zhoubian, Ziniu Hu, Yisong Yue, Yuxiao Dong, and Jie Tang. 2024{\natexlab{a}}.
\newblock \href {https://arxiv.org/pdf/2406.03816} {Rest-mcts*: Llm self-training via process reward guided tree search}.
\newblock \emph{Advances in Neural Information Processing Systems}, 37:64735--64772.

\bibitem[{Zhang et~al.(2024{\natexlab{b}})Zhang, Lei, Gui, Yang, He, Wang, and Xu}]{zhang2024continual-cppo}
Han Zhang, Yu~Lei, Lin Gui, Min Yang, Yulan He, Hui Wang, and Ruifeng Xu. 2024{\natexlab{b}}.
\newblock \href {https://arxiv.org/pdf/2402.14228} {Cppo: Continual learning for reinforcement learning with human feedback}.
\newblock In \emph{The Twelfth International Conference on Learning Representations}.

\bibitem[{Zhang et~al.(2025{\natexlab{a}})Zhang, Huang, Yao, Liu, Zhang, Lu, and Tao}]{zhang2025r1-vl}
Jingyi Zhang, Jiaxing Huang, Huanjin Yao, Shunyu Liu, Xikun Zhang, Shijian Lu, and Dacheng Tao. 2025{\natexlab{a}}.
\newblock \href {{https://arxiv.org/pdf/2503.12937}} {R1-vl: Learning to reason with multimodal large language models via step-wise group relative policy optimization}.
\newblock \emph{arXiv preprint arXiv:2503.12937}.

\bibitem[{Zhang et~al.(2023{\natexlab{a}})Zhang, Li, Li, Li, and Jin}]{zhang2023self-edit}
Kechi Zhang, Zhuo Li, Jia Li, Ge~Li, and Zhi Jin. 2023{\natexlab{a}}.
\newblock \href {{https://arxiv.org/pdf/2305.04087}} {Self-edit: Fault-aware code editor for code generation}.
\newblock \emph{arXiv preprint arXiv:2305.04087}.

\bibitem[{Zhang et~al.(2024{\natexlab{c}})Zhang, Hosseini, Bansal, Kazemi, Kumar, and Agarwal}]{zhang2024genrm}
Lunjun Zhang, Arian Hosseini, Hritik Bansal, Mehran Kazemi, Aviral Kumar, and Rishabh Agarwal. 2024{\natexlab{c}}.
\newblock \href {{https://arxiv.org/pdf/2408.15240}} {Generative verifiers: Reward modeling as next-token prediction}.
\newblock \emph{arXiv preprint arXiv:2408.15240}.

\bibitem[{Zhang et~al.(2024{\natexlab{d}})Zhang, Wu, Lu, Song, Rong, Yao, Zhao, Liu, Feng, Wang, and Sun}]{zhang2024amp}
Mengxi Zhang, Wenhao Wu, Yu~Lu, Yuxin Song, Kang Rong, Huanjin Yao, Jianbo Zhao, Fanglong Liu, Haocheng Feng, Jingdong Wang, and Yifan Sun. 2024{\natexlab{d}}.
\newblock \href {http://papers.nips.cc/paper\_files/paper/2024/hash/2e3073cb65608aa887bb945c382e687f-Abstract-Conference.html} {Automated multi-level preference for mllms}.
\newblock In \emph{Advances in Neural Information Processing Systems 38: Annual Conference on Neural Information Processing Systems 2024, NeurIPS 2024, Vancouver, BC, Canada, December 10 - 15, 2024}.

\bibitem[{Zhang et~al.(2023{\natexlab{b}})Zhang, Press, Merrill, Liu, and Smith}]{zhang2023hallucination-snowball}
Muru Zhang, Ofir Press, William Merrill, Alisa Liu, and Noah~A Smith. 2023{\natexlab{b}}.
\newblock \href {{https://arxiv.org/pdf/2305.13534}} {How language model hallucinations can snowball}.
\newblock \emph{arXiv preprint arXiv:2305.13534}.

\bibitem[{Zhang et~al.(2025{\natexlab{b}})Zhang, Wu, Zhang, Zhao, and Bian}]{zhang2025right-question-half-answer}
Qingyang Zhang, Haitao Wu, Changqing Zhang, Peilin Zhao, and Yatao Bian. 2025{\natexlab{b}}.
\newblock \href {{https://arxiv.org/pdf/2504.05812}} {Right question is already half the answer: Fully unsupervised llm reasoning incentivization}.
\newblock \emph{arXiv preprint arXiv:2504.05812}.

\bibitem[{Zhang et~al.(2025{\natexlab{c}})Zhang, Liu, Zhang, Liu, Luo, Huang, and Gong}]{zhang2025process-self-rewarding}
Shimao Zhang, Xiao Liu, Xin Zhang, Junxiao Liu, Zheheng Luo, Shujian Huang, and Yeyun Gong. 2025{\natexlab{c}}.
\newblock \href {{https://arxiv.org/pdf/2503.03746}} {Process-based self-rewarding language models}.
\newblock \emph{arXiv preprint arXiv:2503.03746}.

\bibitem[{Zhang et~al.(2023{\natexlab{c}})Zhang, Chen, Shen, Ding, Tenenbaum, and Gan}]{zhang2023pg-td-planning-code-geneartion}
Shun Zhang, Zhenfang Chen, Yikang Shen, Mingyu Ding, Joshua~B Tenenbaum, and Chuang Gan. 2023{\natexlab{c}}.
\newblock \href {{https://arxiv.org/pdf/2303.05510}} {Planning with large language models for code generation}.
\newblock \emph{arXiv preprint arXiv:2303.05510}.

\bibitem[{Zhang et~al.(2025{\natexlab{d}})Zhang, Wang, Liu, Huixin, Jiang, Shen, Hou, Zheng, Zhang, Li et~al.}]{zhang2025embodied-reasoner}
Wenqi Zhang, Mengna Wang, Gangao Liu, Xu~Huixin, Yiwei Jiang, Yongliang Shen, Guiyang Hou, Zhe Zheng, Hang Zhang, Xin Li, et~al. 2025{\natexlab{d}}.
\newblock \href {{https://arxiv.org/pdf/2503.21696}} {Embodied-reasoner: Synergizing visual search, reasoning, and action for embodied interactive tasks}.
\newblock \emph{arXiv preprint arXiv:2503.21696}.

\bibitem[{Zhang et~al.(2025{\natexlab{e}})Zhang, Sun, Zhang, Feng, Yang, and Meng}]{zhang2025critique-grpo}
Xiaoying Zhang, Hao Sun, Yipeng Zhang, Kaituo Feng, Chao Yang, and Helen Meng. 2025{\natexlab{e}}.
\newblock \href {{https://arxiv.org/pdf/2506.03106}} {Critique-grpo: Advancing llm reasoning with natural language and numerical feedback}.
\newblock \emph{arXiv preprint arXiv:2506.03106}.

\bibitem[{Zhang et~al.(2025{\natexlab{f}})Zhang, Wen, Wu, and Huang}]{zhang2025tinyllava-video-r1}
Xingjian Zhang, Siwei Wen, Wenjun Wu, and Lei Huang. 2025{\natexlab{f}}.
\newblock \href {{https://arxiv.org/pdf/2504.09641}} {Tinyllava-video-r1: Towards smaller lmms for video reasoning}.
\newblock \emph{arXiv preprint arXiv:2504.09641}.

\bibitem[{Zhang et~al.(2025{\natexlab{g}})Zhang, Yu, Tian, Fu, Li, Zeng, Xie, Shi, Zhang, Wu et~al.}]{zhang2025mm-rlhf}
Yi-Fan Zhang, Tao Yu, Haochen Tian, Chaoyou Fu, Peiyan Li, Jianshu Zeng, Wulin Xie, Yang Shi, Huanyu Zhang, Junkang Wu, et~al. 2025{\natexlab{g}}.
\newblock \href {{https://arxiv.org/pdf/2502.10391}} {Mm-rlhf: The next step forward in multimodal llm alignment}.
\newblock \emph{arXiv preprint arXiv:2502.10391}.

\bibitem[{Zhang et~al.(2023{\natexlab{d}})Zhang, Du, Huang, Wang, Wang, Fang, and Pechenizkiy}]{zhang2023interpretable-reward-distribution}
Yudi Zhang, Yali Du, Biwei Huang, Ziyan Wang, Jun Wang, Meng Fang, and Mykola Pechenizkiy. 2023{\natexlab{d}}.
\newblock \href {https://neurips.cc/virtual/2023/poster/70073} {Interpretable reward redistribution in reinforcement learning: A causal approach}.
\newblock \emph{Advances in Neural Information Processing Systems}, 36:20208--20229.

\bibitem[{Zhang et~al.(2025{\natexlab{h}})Zhang, Zheng, Wu, Zhang, Lin, Yu, Liu, Zhou, and Lin}]{zhang2025lessons-prm}
Zhenru Zhang, Chujie Zheng, Yangzhen Wu, Beichen Zhang, Runji Lin, Bowen Yu, Dayiheng Liu, Jingren Zhou, and Junyang Lin. 2025{\natexlab{h}}.
\newblock \href {{https://arxiv.org/pdf/2501.07301}} {The lessons of developing process reward models in mathematical reasoning}.
\newblock \emph{arXiv preprint arXiv:2501.07301}.

\bibitem[{Zhao et~al.(2025{\natexlab{a}})Zhao, Wang, Fang, Gao, Man, Cui, Wang, Chen, Li, and Zhu}]{zhao2025embodied-r}
Baining Zhao, Ziyou Wang, Jianjie Fang, Chen Gao, Fanhang Man, Jinqiang Cui, Xin Wang, Xinlei Chen, Yong Li, and Wenwu Zhu. 2025{\natexlab{a}}.
\newblock \href {{https://arxiv.org/pdf/2504.12680}} {Embodied-r: Collaborative framework for activating embodied spatial reasoning in foundation models via reinforcement learning}.
\newblock \emph{arXiv preprint arXiv:2504.12680}.

\bibitem[{Zhao et~al.(2025{\natexlab{b}})Zhao, Liu, Zhang, Zhou, Gao, Li, Lyu, Qian, Qi, Li et~al.}]{zhao2025genprm}
Jian Zhao, Runze Liu, Kaiyan Zhang, Zhimu Zhou, Junqi Gao, Dong Li, Jiafei Lyu, Zhouyi Qian, Biqing Qi, Xiu Li, et~al. 2025{\natexlab{b}}.
\newblock \href {{https://arxiv.org/pdf/2504.00891}} {Genprm: Scaling test-time compute of process reward models via generative reasoning}.
\newblock \emph{arXiv preprint arXiv:2504.00891}.

\bibitem[{Zhao et~al.(2025{\natexlab{c}})Zhao, Zhu, and Yang}]{zhao2025refalign}
Shuai Zhao, Linchao Zhu, and Yi~Yang. 2025{\natexlab{c}}.
\newblock \href {{https://arxiv.org/pdf/2504.09895}} {Learning from reference answers: Versatile language model alignment without binary human preference data}.
\newblock \emph{arXiv preprint arXiv:2504.09895}.

\bibitem[{Zhao et~al.(2023)Zhao, Wang, Ouyang, Dong, Wang, and He}]{zhao2023ha-dpo}
Zhiyuan Zhao, Bin Wang, Linke Ouyang, Xiaoyi Dong, Jiaqi Wang, and Conghui He. 2023.
\newblock \href {https://arxiv.org/pdf/2311.16839} {Beyond hallucinations: Enhancing lvlms through hallucination-aware direct preference optimization}.
\newblock \emph{arXiv preprint arXiv:2311.16839}.

\bibitem[{Zheng et~al.(2024)Zheng, Zhang, Zhang, Lin, Lu, Yu, Liu, Zhou, and Lin}]{zheng2024processbench}
Chujie Zheng, Zhenru Zhang, Beichen Zhang, Runji Lin, Keming Lu, Bowen Yu, Dayiheng Liu, Jingren Zhou, and Junyang Lin. 2024.
\newblock \href {{https://arxiv.org/pdf/2412.06559}} {Processbench: Identifying process errors in mathematical reasoning}.
\newblock \emph{arXiv preprint arXiv:2412.06559}.

\bibitem[{Zheng et~al.(2023)Zheng, Chiang, Sheng, Zhuang, Wu, Zhuang, Lin, Li, Li, Xing et~al.}]{zheng2023judging-llm-as-a-judge}
Lianmin Zheng, Wei-Lin Chiang, Ying Sheng, Siyuan Zhuang, Zhanghao Wu, Yonghao Zhuang, Zi~Lin, Zhuohan Li, Dacheng Li, Eric Xing, et~al. 2023.
\newblock \href {https://arxiv.org/pdf/2306.05685} {Judging llm-as-a-judge with mt-bench and chatbot arena}.
\newblock \emph{Advances in Neural Information Processing Systems}, 36:46595--46623.

\bibitem[{Zheng et~al.(2025)Zheng, Fu, Hu, Cai, Ye, Lu, and Liu}]{zheng2025deepresearcher}
Yuxiang Zheng, Dayuan Fu, Xiangkun Hu, Xiaojie Cai, Lyumanshan Ye, Pengrui Lu, and Pengfei Liu. 2025.
\newblock \href {{https://arxiv.org/pdf/2504.03160}} {Deepresearcher: Scaling deep research via reinforcement learning in real-world environments}.
\newblock \emph{arXiv preprint arXiv:2504.03160}.

\bibitem[{Zhou et~al.(2025{\natexlab{a}})Zhou, Zhang, Song, Chen, Gu, Ma, Tian, Zhang, and Hu}]{zhou2025refinecoder}
Changzhi Zhou, Xinyu Zhang, Dandan Song, Xiancai Chen, Wanli Gu, Huipeng Ma, Yuhang Tian, Mengdi Zhang, and Linmei Hu. 2025{\natexlab{a}}.
\newblock \href {{https://arxiv.org/pdf/2502.09183}} {Refinecoder: Iterative improving of large language models via adaptive critique refinement for code generation}.
\newblock \emph{arXiv preprint arXiv:2502.09183}.

\bibitem[{Zhou et~al.(2024{\natexlab{a}})Zhou, Zheng, Wang, Xi, Dou, Bao, Shen, Xiong, Fan, Mou, Zheng, Gui, Zhang, and Huang}]{zhou2024rmb}
Enyu Zhou, Guodong Zheng, Binghai Wang, Zhiheng Xi, Shihan Dou, Rong Bao, Wei Shen, Limao Xiong, Jessica Fan, Yurong Mou, Rui Zheng, Tao Gui, Qi~Zhang, and Xuanjing Huang. 2024{\natexlab{a}}.
\newblock \href {https://arxiv.org/pdf/2410.09893} {Rmb: Comprehensively benchmarking reward models in llm alignment}.
\newblock \emph{arXiv preprint arXiv:2410.09893}.

\bibitem[{Zhou et~al.(2025{\natexlab{b}})Zhou, Li, Wang, Cheng, Zhou, and Hsieh}]{zhou2025r1-zero-aha-moment}
Hengguang Zhou, Xirui Li, Ruochen Wang, Minhao Cheng, Tianyi Zhou, and Cho-Jui Hsieh. 2025{\natexlab{b}}.
\newblock \href {{https://arxiv.org/pdf/2503.05132}} {R1-zero's" aha moment" in visual reasoning on a 2b non-sft model}.
\newblock \emph{arXiv preprint arXiv:2503.05132}.

\bibitem[{Zhou et~al.(2025{\natexlab{c}})Zhou, Xu, Wang, Xiong, and Joty}]{zhou2025evaluating-judges-as-evaluators}
Yilun Zhou, Austin Xu, Peifeng Wang, Caiming Xiong, and Shafiq Joty. 2025{\natexlab{c}}.
\newblock \href {https://arxiv.org/pdf/2504.15253} {Evaluating judges as evaluators: The jetts benchmark of llm-as-judges as test-time scaling evaluators}.
\newblock \emph{arXiv preprint arXiv:2504.15253}.

\bibitem[{Zhou et~al.(2024{\natexlab{b}})Zhou, Cui, Rafailov, Finn, and Yao}]{zhou2024povid}
Yiyang Zhou, Chenhang Cui, Rafael Rafailov, Chelsea Finn, and Huaxiu Yao. 2024{\natexlab{b}}.
\newblock \href {{https://arxiv.org/pdf/2402.11411}} {Aligning modalities in vision large language models via preference fine-tuning}.
\newblock \emph{arXiv preprint arXiv:2402.11411}.

\bibitem[{Zhou et~al.(2024{\natexlab{c}})Zhou, Liu, Ning, Liu, Wang, Wong, Huang, Wang, and Huang}]{zhou2024mathcheck-gsm}
Zihao Zhou, Shudong Liu, Maizhen Ning, Wei Liu, Jindong Wang, Derek~F Wong, Xiaowei Huang, Qiufeng Wang, and Kaizhu Huang. 2024{\natexlab{c}}.
\newblock \href {{https://arxiv.org/pdf/2407.08733}} {Is your model really a good math reasoner? evaluating mathematical reasoning with checklist}.
\newblock \emph{arXiv preprint arXiv:2407.08733}.

\bibitem[{Zhu et~al.(2025)Zhu, Chen, Dou, Li, Guo, Chen, and Zhang}]{zhu2025dianjin-r1}
Jie Zhu, Qian Chen, Huaixia Dou, Junhui Li, Lifan Guo, Feng Chen, and Chi Zhang. 2025.
\newblock \href {https://arxiv.org/pdf/2504.15716} {Dianjin-r1: Evaluating and enhancing financial reasoning in large language models}.

\bibitem[{Zhu et~al.(2024)Zhu, Guo, Shao, Yang, Wang, Xu, Wu, Li, Gao, Ma et~al.}]{zhu2024deepseek-coder-v2}
Qihao Zhu, Daya Guo, Zhihong Shao, Dejian Yang, Peiyi Wang, Runxin Xu, Y~Wu, Yukun Li, Huazuo Gao, Shirong Ma, et~al. 2024.
\newblock \href {{https://arxiv.org/pdf/2406.11931}} {Deepseek-coder-v2: Breaking the barrier of closed-source models in code intelligence}.
\newblock \emph{arXiv preprint arXiv:2406.11931}.

\bibitem[{Zhu et~al.(2022)Zhu, Wang, Zhang, Zhang, Gan, Zhang, and Yang}]{zhu2022cooperative}
Xinyu Zhu, Junjie Wang, Lin Zhang, Yuxiang Zhang, Ruyi Gan, Jiaxing Zhang, and Yujiu Yang. 2022.
\newblock \href {{https://arxiv.org/pdf/2210.16257}} {Solving math word problems via cooperative reasoning induced language models}.
\newblock \emph{arXiv preprint arXiv:2210.16257}.

\bibitem[{Ziegler et~al.(2019)Ziegler, Stiennon, Wu, Brown, Radford, Amodei, Christiano, and Irving}]{ziegler2019finetuning}
Daniel~M. Ziegler, Nisan Stiennon, Jeffrey Wu, Tom~B. Brown, Alec Radford, Dario Amodei, Paul Christiano, and Geoffrey Irving. 2019.
\newblock \href {https://arxiv.org/pdf/1909.08593} {Fine-tuning language models from human preferences}.
\newblock \emph{arXiv preprint arXiv:1909.08593}.

\bibitem[{Zuo et~al.(2025)Zuo, Zhang, Qu, Sheng, Zhu, Qi, Sun, Cui, Ding, and Zhou}]{zuo2025test-time-rl}
Yuxin Zuo, Kaiyan Zhang, Shang Qu, Li~Sheng, Xuekai Zhu, Biqing Qi, Youbang Sun, Ganqu Cui, Ning Ding, and Bowen Zhou. 2025.
\newblock \href {{https://arxiv.org/pdf/2504.16084}} {Ttrl: Test-time reinforcement learning}.
\newblock \emph{arXiv preprint arXiv:2504.16084}.

\end{thebibliography}

\end{document}